\documentclass[11pt]{article}
\usepackage{fullpage}

\usepackage[utf8]{inputenc} % allow utf-8 input
\usepackage[T1]{fontenc}    % use 8-bit T1 fonts
\usepackage{hyperref}       % hyperlinks
\usepackage[tbtags]{amsmath}
\usepackage{amssymb}
\usepackage{mathtools}
\usepackage{amsthm}
\usepackage{dsfont}
\usepackage{tikz}

\usepackage[capitalize,noabbrev]{cleveref}
\usepackage{physics}

\hypersetup{
    pdfnewwindow=true,     % links in new window
    colorlinks=true,     % false: boxed links; true: colored links
    linkcolor=cyan,     % color of internal links (change box color with linkbordercolor)
    citecolor=cyan,     % color of links to bibliography
    filecolor=magenta,     % color of file links
    urlcolor=cyan     % color of external links
}
\usepackage{url} 
\usepackage{booktabs}       % professional-quality tables
\usepackage{amsfonts}       % blackboard math symbols
\usepackage{nicefrac}       % compact symbols for 1/2, etc.
\usepackage{enumerate}   

\usepackage{microtype}      % microtypography
\usepackage{xcolor}         % colors
\usepackage{bbm}
\usepackage{xfrac}
\usepackage{tocbibind}
\usepackage[title]{appendix}
\usepackage{tikz}
\usepackage{listofitems} % for \readlist to create arrays

\usepackage{subfiles}

 \makeatletter
\newtheorem*{rep@theorem}{\rep@title}
\newcommand{\newreptheorem}[2]{%
\newenvironment{rep#1}[1]{%
 \def\rep@title{#2 \ref{##1}}%
 \begin{rep@theorem}}%
 {\end{rep@theorem}}}
\makeatother

\makeatletter
\newcommand{\maketitlepage}{%
    \let\thanks\@gobble
    \let\footnote\@gobble
    \@maketitle
}
\makeatother

\newtheorem{innercustomgeneric}{\customgenericname}
\providecommand{\customgenericname}{}
\newcommand{\newcustomtheorem}[2]{%
  \newenvironment{#1}[1]
  {%
   \renewcommand\customgenericname{#2}%
   \renewcommand\theinnercustomgeneric{##1}%
   \innercustomgeneric
  }
  {\endinnercustomgeneric}
}
\theoremstyle{plain}
\newtheorem{theorem}{Theorem}[section]
\newtheorem{proposition}[theorem]{Proposition}
\newtheorem{lemma}[theorem]{Lemma}

\theoremstyle{definition}
\newtheorem{definition}[theorem]{Definition}
\newtheorem{assumption}{Assumption}
\newcustomtheorem{customassumption}{Assumption}

\theoremstyle{remark}
\newtheorem{remark}[theorem]{Remark}
\newtheorem{example}[theorem]{Example}

\newreptheorem{theorem}{Theorem}
\newreptheorem{lemma}{Lemma}
\newreptheorem{proposition}{Proposition}
\newreptheorem{corollary}{Corollary}
\newreptheorem{assumption}{Assumption}
\newreptheorem{example}{Example}

\newcommand{\ols}[1]{\,\widetilde{\!{#1}}} % ols short
\allowdisplaybreaks
\usepackage{xspace}

\newcounter{aliCounter}

\newcounter{lukaszCounter}

\newcounter{samCounter}

\newcommand{\mfm}{\mathfrak{m}}
\newcommand{\mrd }{\mathrm{d}}
\newcommand{\mdens}{\mfm_{\theta}}

\newcommand{\KLr}{\mathrm{KL}}
\newcommand{\mbR}{\mathbb{R}}
\newcommand{\nupop}{\nu_{\mathrm{pop}}}

\usepackage[textsize=tiny]{todonotes}

\usepackage{amsmath} % for aligned
\usepackage{tikz}
\usepackage{listofitems} % for \readlist to create arrays
\usetikzlibrary{arrows.meta} % for arrow size
\usepackage[outline]{contour} % glow around text
\contourlength{1.4pt}

\tikzset{>=latex} % for LaTeX arrow head
\usepackage{xcolor}
\colorlet{myred}{red!80!black}
\colorlet{myblue}{blue!80!black}
\colorlet{mygreen}{green!60!black}
\colorlet{myorange}{orange!70!red!60!black}
\colorlet{mydarkred}{red!30!black}
\colorlet{mydarkblue}{blue!40!black}
\colorlet{mydarkgreen}{green!30!black}
\tikzstyle{node}=[thick,circle,draw=myblue,minimum size=22,inner sep=0.5,outer sep=0.6]
\tikzstyle{node in}=[node,green!20!black,draw=mygreen!30!black,fill=mygreen!25]
\tikzstyle{node hidden}=[node,blue!20!black,draw=myblue!30!black,fill=myblue!20]
\tikzstyle{node convol}=[node,orange!20!black,draw=myorange!30!black,fill=myorange!20]
\tikzstyle{node out}=[node,red!20!black,draw=myred!30!black,fill=myred!20]
\tikzstyle{connect}=[thick,mydarkblue] %,line cap=round
\tikzstyle{connect arrow}=[-{Latex[length=4,width=3.5]},thick,mydarkblue,shorten <=0.5,shorten >=1]
\tikzset{ % node styles, numbered for easy mapping with \nstyle
  node 1/.style={node in},
  node 2/.style={node hidden},
  node 3/.style={node out},
}
\usepackage{parskip}

\usepackage{natbib}
\setcitestyle{citesep={;}}
\usepackage{cite}
\title{ Mean-field Analysis of Generalization Errors}

\author{
  Gholamali Aminian$^{*\, 1}$
  \and
  Samuel N. Cohen$^{*\,1\,2}$
  \and
  {\L}ukasz Szpruch$^{*\, 1\,3}$
}

\begin{document}

\maketitle
\renewcommand\thefootnote{$*$}\footnotetext{Equal contribution.}
\renewcommand\thefootnote{$^1$}\footnotetext{The Alan Turing Institute.}
\renewcommand\thefootnote{$^2$}\footnotetext{Mathematical Institute, University of Oxford.}
\renewcommand\thefootnote{$^3$}\footnotetext{School of Mathematics, University of Edinburgh.}

\begin{abstract}
 We propose a novel framework for exploring weak and $L_2$ generalization errors of algorithms through the lens of differential calculus on the space of probability measures. Specifically, we consider the KL-regularized empirical risk minimization problem and establish generic conditions under which the generalization error convergence rate, when training on a sample of size $n$, is $\mathcal{O}(1/n)$. In the context of supervised learning with a one-hidden layer neural network in the mean-field regime, these conditions are reflected in suitable integrability and regularity assumptions on the loss and activation functions.
\end{abstract}

\section{Introduction}

A crucial problem in learning theory is understanding the performance of a learning algorithm on unseen data. This is described by the generalization error, which quantifies the deviation between the algorithm's performance on the training data and its performance on unseen data sampled from the same underlying distribution. In the under-parameterized regime, where the number of model parameters is significantly less than the number of training data points, the theory of the generalization error is well established \citep{vapnik/chervonenkis:2015,bartlett/mendelson:2002}. However, in the overparameterized regime, where the number of parameters may greatly exceed the number of training data points, this theory becomes inadequate. For example, deep neural network (NN) models can exhibit nearly zero training loss, which implies that they interpolate the training data, and yet still perform well on out-of-sample data \citep{belkin/hsu/ma/mandal:2019,spigler/geigereta:2019,bartlett/montanari:2021}. Mathematical models, such as the Neural Tangent Kernel~\citep{jacot2018neural}, Mean-Field \citep{mei/montanari/nguyen:2018}, and Random Feature models \citep{rahimi2008uniform}, have been proposed to understand the behavior of overparameterized NNs. However, our current understanding of generalization in these contexts still needs to be completed.
Our approach is partially motivated by the recently developed mean-field perspective on training  NNs. In a series of works  \citep{mei/montanari/nguyen:2018,chizat/bach:2018,mei/misiakiewicz/montanari:2019,sirignano/spiliopoulos:2019,MFLD,tzen/raginsky:2020}, the task of learning the optimal weights in deep NNs is viewed as a sampling problem. The picture that emerges from the mean-field regime is that the aim of a learning algorithm is to find an optimal distribution over the parameter space (rather than optimal values of the parameters). This work investigates how the mean-field perspective can shed light on the generalization performance of overparameterized NNs. 

The contributions of this work are as follows:

\begin{itemize} \vspace{-1.5mm}
    \item We introduce and develop a general approach to study generalization error (cf. \eqref{def generalization error} for the definition) that utilizes calculus on the space of measures, encompassing a wide range of learning models and risk functions.\vspace{-1.5mm}
    \item  In the context of Kullback--Leibler (KL) regularized empirical risk minimization, our analysis provides if-and-only-if  conditions for achieving a convergence rate of $\mathcal{O}({1}/{n})$ for weak generalization errors and
     sharp conditions for achieving a convergence rate of $\mathcal{O}({1}/{n})$ for $L_2$.   These bounds are not asymptotic, i.e.~they are valid for all $n$.
    \vspace{-1.5mm}
    \item For one-hidden layer neural networks in the mean-field regime, our analysis reveals precise conditions on model inputs that guarantee a convergence rate of $\mathcal{O}(1/n)$
    of both weak and $L_2$ generalization errors. This improves the generalization error bounds which can be derived from the proof techniques in 
    \citet{nishikawatwo} and \citet{nitanda2021particle}. 
\end{itemize}

\section{Preliminaries}\label{Sec:Problem-formulation}

\paragraph{Notation:}We adopt the following convention for random variables and their distributions. 
A random variable is denoted by an upper-case letter (e.g., $Z$), its space of possible values is denoted with the corresponding calligraphic letter (e.g. $\mathcal{Z}$), and an arbitrary value of this variable is denoted with the lower-case letter (e.g., $z$). We will write $\mathbb{E}_{Z}[\cdot]$ for the expectation taken over $Z$, all other random variables being left constant\footnote{Formally, this corresponds to the conditional expectation over all variables in our setting excluding $Z$. As we will only have countably many variables in our problem, this does not cause any technical difficulties.}. We will further write $\mathbb{E}_{Z\sim m}[\cdot]$ for the expectation over a random variable $Z$ independent of all others, which is distributed according to $m$, and similarly $\mathbb{V}_{Z\sim m}$ for the variance. We write $\delta_z$ for a Dirac measure supported at $z$.

 If $A$ is a normed space, then $ \mathcal{P}(A)$ denotes the space of probability measures on $A$ and $\mathcal{P}_{p}(A)$ the probability measures with finite $p$-th moment. We endow spaces of probability measures with the weak topology and corresponding Borel $\sigma$-algebra.

We will take the metric over spaces of interest (and finite products of these spaces) to be defined a priori, and will assume that the metric over a product space is equivalent to the product (Euclidean) metric.

We now introduce the functional linear derivative~\citep{cardaliaguet2019master} for functionals on measure spaces. For simplicity\footnote{This can be relaxed, along with most integrability assumptions in this paper, at a cost of complexity.}, we will restrict our attention to the finite variance ($\mathcal{P}_2(\mbR^n)$ case).

\begin{definition}{Extending \citep[Definition 5.43]{carmona/delarue:2018}}
\label{def:flatDerivative}
Consider $U:\mathcal P_2(\mathbb R^n) \times \mathbb R^k \to \mathbb R$. We say $m\mapsto U(m,x)$ is of class $\mathcal C^1$ if there exists a map $\frac{\delta U}{\delta m} : \mathcal P_2(\mathbb R^n) \times \mathbb R^k \times \mathbb R^n \to \mathbb R$, such that 
\begin{enumerate}[(i)]
    \item $\frac{\delta U}{\delta m}$ is measurable with respect to $x,a$, and continuous with respect to $m$;
    \item \label{flatder_item1}for every bounded set $B\subset \mathcal P_2(\mathbb R^n) \times \mathbb R^k$, there exists a constant $C>0$ such that  $|\frac{\delta U}{\delta m}(m,x,a)|\leq C(1+|a|^2)$ for all $(m,x)\in B$;
    \item for all $m, m' \in\mathcal P_2(\mathbb R^n)$, 
\[
U(m',x) - U(m,x) = \int_0^1 \int_{\mbR^n} \frac{\delta U}{\delta m}(m + \lambda(m' - m),x,a) \, (m'
-m)(da)\,\mrd \lambda\,.
\]
\end{enumerate}
Since $\frac{\delta U}{\delta m}$ is only defined up to a constant we demand $\int \frac{\delta U}{\delta m}(m,x,a)\,m(da) = 0$. By extension, we say $U$ is of class $\mathcal{C}^2$ if both $U$ and $\frac{\delta U}{\delta m}$ are of class $\mathcal{C}^1$.
\end{definition}

Similarly, we define derivatives for measure-valued functionals. See Appendix \ref{app_calculus} for proofs.

\begin{definition}\label{Def: func derivative m2m}
A functional $\mfm:\mathcal{P}_2(\mbR^n) \to \mathcal{P}_2(\mbR^k)$ is of class $\mathcal{C}^1$ if for all $A\in \mathcal{B}(\mbR^k)$, the functional $\mfm|_A:=m(\cdots)(A)$ is $\mathcal{C}^1$ in the sense of Definition \ref{def:flatDerivative}, and its derivative $\frac{\delta \mfm}{\delta \nu}(A):=\frac{\delta \mfm|_A}{\delta \nu}$ is a (Borel) signed measure on $\mbR^k$, where the constant $C$ in Definition \ref{def:flatDerivative}(\ref{flatder_item1}) does not depend on the set $A$.
\end{definition}

\begin{lemma}\label{Lemma: sam}
    Let $\mfm$ be a functional as in Definition \ref{Def: func derivative m2m}, with the density representation
    $\mfm(\nu)(A) = \int_A U(\nu, x)\mrd x$
    for all Borel sets $A$. Suppose $\mfm$ and $U(\cdot,x)$, for each $x$, admit linear derivatives.
    Then the linear derivatives of $\mfm$ and $U$ satisfy
    $\frac{\delta\mfm}{\delta\nu}(\nu,a)(A) = \int_A \frac{\delta U}{\delta\nu}(\nu,a,x)\mrd x,$
or, more succinctly, $\frac{\delta\mfm}{\delta\nu}(\nu,a)(\mrd x) =  \frac{\delta U}{\delta\nu}(\nu,a,x)\mrd x$.
\end{lemma}

\subsection{Problem formulation} \label{sec: problem formulation}
Consider a supervised learning scenario, with input space $\mathcal{X}$ and target space $\mathcal{Y}$. Define $\mathcal{Z}:=\mathcal{X}\times\mathcal{Y}$. We are often interested in learning a function $f : \mathcal{X} \to \mathcal{Y}$  parameterized  using a (large) number of parameters from $\Theta\subseteq \mathbb{R}^k$, for some $k>0$. 

Let $\nupop \in \mathcal{P}_2(\mathcal{Z})$ be the (unknown) distribution, describing the joint values of $(x,y)$ in the population. Intuitively, we suppose that the mass of $\nupop$ is near the graph of $f$ that is, $y\approx f(x)$ when $z=(x,y)$ is sampled from $\nupop$. Since $\nupop$ is unknown, we approximate $f$ based  on a finite dataset, $\mathbf{Z}_n =  \{Z_{i}\}_{i=1}^n$. We make the following basic assumption on our data, which will be taken to hold throughout the paper.

\begin{assumption}
    Our observations are given by $\mathbf{Z}_n =  \{Z_{i}\}_{i=1}^n$, where $Z_i=(X_{i},Y_{i})\sim \nupop$ are i.i.d. We write $\nu_n:= \frac{1}{n}\sum_{i=1}^n \delta_{Z_i}$ for the corresponding empirical measure.
\end{assumption}

We are interested in quantifying the performance of our model on unseen data. To this end, let $\mathbf{\ols{Z}}_n=\{\ols{Z}_i\}_{i=1}^n$ be a second i.i.d sample set with law $\nupop$, independent of $\mathbf{Z}_n$. We define the perturbations of $\nu_n$ obtained by `resampling' one or two data points, by 
\begin{equation}\label{Eq: nu replace one}\nu_{n,(1)}=\nu_n+\frac{1}{n}(\delta_{\ols{Z}_1}-\delta_{Z_1}), \qquad 
\nu_{n,(1,2)}=\nu_n + \frac{1}{n}(\delta_{\ols{Z}_1}-\delta_{Z_1}+\delta_{\ols{Z}_2}-\delta_{Z_2})\,.\end{equation}

\paragraph{Learning algorithm:}
We consider learning algorithms which can be represented as (measurable) maps from the empirical measure $\nu_n$ to a measure over parameters, $\nu_n \mapsto \mfm(\nu_n) \in \mathcal{P}(\Theta)$. In classical statistics, learning algorithms are  (deterministic) maps from observed data to parameter estimates. By generalizing to measures over parameters, we allow for randomness in training algorithms (e.g.~stochastic gradient descent), for Bayesian interpretations (where $\mfm(\nu_n)$ is the  posterior distribution), and for a formalism for overparameterized models; furthermore, we gain access to the tools of calculus on probability spaces.

To investigate resampling the training data, we will make use of the convex perturbations, for $\lambda\in [0,1]$ and $\mathcal{J}\in \{(1), (1,2)\}$,
\begin{align}
&\begin{cases}
\mfm_{\mathcal{J}}(\lambda)=\mfm(\nu_{n,\mathcal{J}})+\lambda (\mfm(\nu_{n})-\mfm(\nu_{n,\mathcal{J}}))\\
\nu_{(1)}(\lambda)=\nu_{n,\mathcal{J}}+ \lambda (\nu_{n}-\nu_{n,\mathcal{J}}),\\
\end{cases}\label{mperturb1}
\end{align}

\subsubsection{Loss function}

We focus on training methods where $\mfm$ is the minimizer of a loss function. In order to motivate our generic formulation of this problem, we present the concrete example of overparameterized one-hidden layer neural networks. An application to expected parametric loss can be found in Appendix \ref{app: expected loss}.
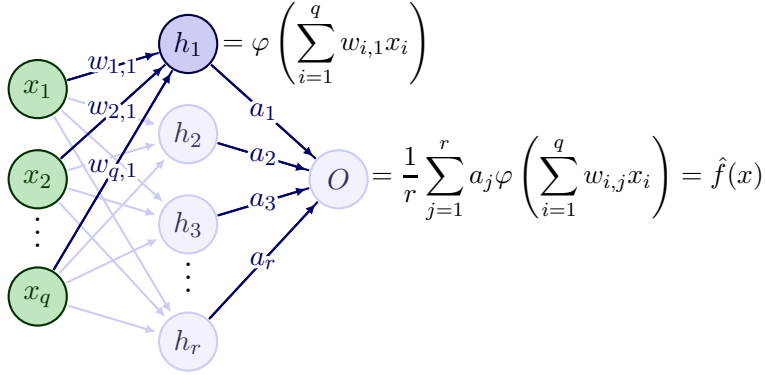
\begin{figure}[t!]
    \centering
% \begin{tikzpicture}[x=2cm,y=1.2cm]

% \centering
%   \readlist\Nnod{2,4,1} % number of nodes per layer
%   % \Nnodlen = length of \Nnod (i.e. total number of layers)
%   % \Nnod[1] = element (number of nodes) at index 1
%   \foreachitem \N \in \Nnod{ % loop over layers
%     % \N     = current element in this iteration (i.e. number of nodes for this layer)
%     % \Ncnt  = index of current layer in this iteration
%     \foreach \i [evaluate={\x=\Ncnt; \y=\N/2-\i+0.5; \prev=int(\Ncnt-1);}] in {1,...,\N}{ % loop over nodes
%       \node[mynode] (N\Ncnt-\i) at (\x,\y) {};
%       \ifnum\Ncnt>1 % connect to previous layer
%         \foreach \j in {1,...,\Nnod[\prev]}{ % loop over nodes in previous layer
%           \draw[thick] (N\prev-\j) -- (N\Ncnt-\i); % connect arrows directly
%         }
%       \fi % else: nothing to connect first layer
%     }
%   }
% %   \caption{One hidden neural network}
% \end{tikzpicture}

\begin{tikzpicture}[x=2cm,y=1.2cm]
  \message{^^Neural network activation}
  \def\NI{3} % number of nodes in input layers
  \def\NO{4} % number of nodes in output layers
  \def\NFO{2}
  \def\yshift{0.3} % shift last node for dots
  
  % INPUT LAYER
  \foreach \i [evaluate={\c=int(\i==\NI); \y=\NI/2-\i-\c*\yshift; \index=(\i<\NI?int(\i):"q");}]
              in {1,...,\NI}{ % loop over nodes
    \node[node in,outer sep=0.6] (NI-\i) at (0,\y) {$x_{\index}$};
  }
  
  % OUTPUT LAYER
  \foreach \i [evaluate={\c=int(\i==\NO); \y=\NO/2-\i-\c*\yshift; \index=(\i<\NO?int(\i):"r");}]
    in {\NO,...,1}{ % loop over nodes
    \ifnum\i=1 % high-lighted node
      \node[node hidden]
        (NO-\i) at (1,\y) {$h_{\index}$};
      \foreach \j [evaluate={\index=(\j<\NI?int(\j):"q");}] in {1,...,\NI}{ % loop over nodes in previous layer
        \draw[connect arrow] (NI-\j) -- (NO-\i);
        \draw[connect] (NI-\j) -- (NO-\i)
          node[pos=0.50] {\contour{white}{$w_{\index,1}$}};
      }
    \else % other light-colored nodes
      \node[node,blue!20!black!80,draw=myblue!20,fill=myblue!5]
        (NO-\i) at (1,\y) {$h_{\index}$};
      \foreach \j in {1,...,\NI}{ % loop over nodes in previous layer
        %\draw[connect,white,line width=1.2] (NI-\j) -- (NO-\i);
        \draw[connect arrow,myblue!20] (NI-\j) -- (NO-\i);
      }
    \fi
  }
  
  % OUTPUT LAYER
 % loop over nodes
      \node[node,blue!20!black!80,draw=myblue!20,fill=myblue!5]
        (NOF) at (2,-0.5) {$O$};
      \foreach \j [evaluate={\index=(\j<\NO?int(\j):"r");}] in {1,...,\NO}{ % loop over nodes in previous layer
        \draw[connect arrow] (NO-\j) -- (NOF);
        \draw[connect] (NO-\j) -- (NOF)
          node[pos=0.50] {\contour{white}{$a_{\index}$}};
      }

  % DOTS
  \path (NI-\NI) --++ (0,1+\yshift) node[midway,scale=1.2] {$\vdots$};
  \path (NO-\NO) --++ (0,1+\yshift) node[midway,scale=1.2] {$\vdots$};
  
  % EQUATIONS
  % \def\agr#1{{\color{mydarkgreen}a_{#1}^{(0)}}}

  \node[below=0.125,right=0.135,black,scale=0.95] at (NO-1)
{$\begin{aligned} \,=\varphi\left(\sum_{i=1}^{q}w_{i,1}x_i\right)\end{aligned}$};

%   \node[below=7,right=11,black,scale=0.95] at (NO-1)
% {$\begin{aligned} =\varphi\left(\sum_{i=1}^{r}w_{i,1}x_i\right)\end{aligned}$};
     
  % \def\agr#1{{\color{mydarkgreen}a_{#1}^{(0)}}}
  \node[below=0.125,right=0.125,black,scale=0.95] at (NOF)
    {$\begin{aligned} 
     \color{black}\,=\frac{1}{r}\sum_{j=1}^r a_j\varphi\left( 
           \sum_{i=1}^{q} w_{i,j}x_i 
           \right)=\hat f(x)
     \end{aligned}$};

\end{tikzpicture}
\caption{One Hidden-layer NN}
\label{Fig: NN}
\end{figure}

\textbf{Overparameterized one-hidden layer neural network (NN):} We consider a one-hidden layer NN, e.g.~as shown in Figure~\ref{Fig: NN}. Let $x\in \mathbb R^q$ be a feature vector. For each of $r$ hidden neurons, we denote the parameters of the hidden layer by $w\in\mbR^q$ and the outer layer by $a\in\mbR$. The parameter space is $\Theta=\mbR^{(q+1)}=\{a\in\mbR,w\in\mbR^q\}$.  To ease notation, we write $\phi(\theta_i,x)=a_i\varphi(w_i\cdot x)$, where $\varphi(\cdot)$ is the activation function.  The output of the NN is
\begin{equation}\label{Eq: phi def}
\hat f(x) = \frac{1}{r}\sum_{i=1}^r a_i\varphi( w_i\cdot x)=\frac{1}{r}\sum_{i=1}^r \phi(\theta_i,x)= \int \phi(\theta, x)\, m_r\! (\mrd \theta)= \mathbb{E}_{\theta\sim m_r}[\phi(\theta,x)],
\end{equation}
 where $m_r:=\frac{1}{r}\sum_{i=1}^r \delta_{(a_i,w_i)}.$ 
Observe that any hidden nodes can be switched (along with their parameters) without changing the output of the network; this symmetry implies that it is the (joint) distribution $m_r$ which is important. As $r$ increases, $m_r$ can converge (weakly) to a continuous distribution over the parameter space; the weights for individual neurons can be viewed as samples from this measure.  This is the mean-field model studied in~\citep{mei/montanari/nguyen:2018,mei/misiakiewicz/montanari:2019,MFLD}. 

Training neural networks chooses parameters $\theta$ to minimize a loss $\ell_o$ between $\hat f(x)$ and the observed value $y$. With $\hat y:=\hat f(x) = \mathbb{E}_{\theta\sim m_r}[\phi(\theta, x)]$, we can write our single-observation loss function as 
\begin{equation}
    \label{eq: nn loss function}
    \ell(m, z):=\ell_o\big(\mathbb{E}_{\theta\sim m}[\phi(\theta,x)],y\big).\end{equation} 
For example,  consider $\ell_o\big(\hat y,y\big)=c(y\hat y)$ for binary classification, where $c(\cdot)$ is a margin-based loss function~\citep{bartlett2006convexity}, or absolute loss $\ell_o(\hat y,y)=|y-\hat y|$ or quadratic loss $\ell_o(\hat y,y)=(y-\hat y)^2$ for classification and regression problems.

\textbf{Generic loss function based on parameter measure:} Inspired by the scenario above, we generically\footnote{As before, we assume that $\ell$ is measurable as a map from $\mathcal{P}(\Theta)\times \mathcal{Z}\to \mathbb{R}$.} represent the single-observation loss function as 
\begin{equation}\label{eq: general loss function}
(m,z)\mapsto \ell(m,z) \in \mathbb R^+\,.
\end{equation}
This represents the loss, for the parameter distribution $m$ and a data point $z$. The loss is typically non-linear in $m$ for the overparameterized one-hidden layer NN \eqref{eq: nn loss function}.
\subsubsection{Risks}
The risk function\footnote{The functional $\mathrm{R}$ is (weakly) continuous in $\nu$, and is measurable in $m$ (as a consequence of the measurability of $\ell$). As $\ell(m, z)$ is nonnegative, the integral \eqref{eq risk linear} can always be defined, without integrability assumptions on $\ell$, but may take the value $+\infty$.} takes distributions $m$ of parameters and $\nu$ of data points and evaluates the loss: 
\begin{align} \label{eq risk linear}
    \mathrm{R}(m,\nu):= \int_{\mathcal Z} \ell\big( m,z\big)\nu(\mrd z)\,.
   \end{align}
   The population risk is $\mathrm{R}(m, \nupop)$, and the empirical risk is  $\mathrm{R}(m,\nu_n)$, which simplifies as 
\begin{align} \mathrm{R}(m,\nu_n) =  \int_{\mathcal Z} \ell\big( m,z\big)\nu_n(\mrd z)=\frac{1}{n}\sum_{i=1}^n \ell\big( m,z_i\big).
\end{align}
\paragraph{Generalization error:} We aim to study the performance of the model trained with the empirical data measure $\nu_n$, and evaluated against the population measure $\nupop$, that is, $\mathrm{R}(\mfm(\nu_n),\nupop)$.
Following \citet{Bousquet-Elisseeff} and \citet{xu2017information}, the  risk can be decomposed:
\begin{align} \label{def generalization error} 
 \mathrm{R}(\mfm(\nu_n),\nupop) =    \underbrace{\Big(\mathrm{R}(\mfm(\nu_n),\nupop) - \mathrm{R}(\mfm(\nu_n),\nu_n)\Big)}_{\text{generalization error}}+ \underbrace{\mathrm{R}(\mfm(\nu_n),\nu_n)}_{\text{training error}}.
\end{align}
Due to the randomness of the training data $\mathbf{Z_n}$ (and hence $\nu_n$), the generalization error is not deterministic, so we seek to estimate its statistics.

\begin{assumption}\label{assn:training integrable}
    The training map $\mfm$, loss function $\ell$, and population measure $\nu$ are such that the training error satisfies $\mathbb{E}_{\mathbf{Z}_n}[\mathrm{R}(\mfm(\nu_n),\nu_n)] = \mathbb{E}_{\mathbf{Z}_n}[\int_\mathcal{Z}\ell(\mfm(\nu_n),z)\nu_n(\mrd z)]<\infty$.
\end{assumption}

For a training map $\mfm$, we will consider both the Weak Generalization Error (WGE),
\begin{align}
    \label{eq:WGE}
   \mathrm{gen}(\mfm,\nupop)\triangleq \mathbb{E}_{\mathbf{Z}_n}\big[ \mathrm{R}(\mfm(\nu_n),\nupop)-\mathrm{R}(\mfm(\nu_n),\nu_n) \big].
\end{align}
and the $L_2$-Generalization Error (LGE)
\begin{align}
    \label{eq:LGE}
\mathrm{gen}_{L_2}(\mfm,\nupop)\triangleq \mathbb{E}_{\mathbf{Z}_n}\big[ | \mathrm{R}(\mfm(\nu_n),\nupop)-\mathrm{R}(\mfm(\nu_n),\nu_n)|^2 \big].
\end{align} 
 Our main results will be conditions under which these errors are $\mathcal{O}(1/n)$. Markov's inequality shows that, for all $\epsilon>0$, 
 \[\mathbb{P}\big(|\mathrm{R}(\mfm(\nu_n),\nupop)-\mathrm{R}(\mfm(\nu_n),\nu_n)|>\epsilon\big)\leq \mathrm{gen}_{L_2}(\mfm,\nupop)\big/\epsilon^2,\] giving probabilistic bounds on the (random) generalization error.

\subsection{Related works}
\paragraph{Generalization error and overparameterization:}
There are three main approaches to analyzing learning problems in an overparameterized regime: the neural tangent kernel (NTK), random feature and mean-field approaches. Under some assumptions, the NTK approach (a.k.a. lazy training)  shows that an overparameterized one-hidden layer NN converges to an (infinite dimensional) linear model. The neural tangent random feature approach is similar to NTK, where the model is defined based on the network gradients at the initialization.
The random feature model is similar to NTK, with an extra assumption of constant weights in the single hidden layer of the NN. The mean-field approach uses the exchangeability of neurons to work with distributions of parameters. The study of these methods' generalization performance allows us to extend our understanding of the overparameterized regime. See, for example, for NTK ~\citep{li2018learning,cao2019generalization,arora2019exact,arora2019fine,allen2019learning,ji2019polylogarithmic}, neural tangent random feature \citep{cao2019generalization,chenmuch}, random feature~\citep{mei2022generalization,ma2019generalization} and mean-field \citep{nitanda2021particle,nishikawatwo} settings. The neural tangent kernel and random feature results do not precisely reflect practice due to their constraints, such as the solution's inability to deviate too far from the weights' initialization~\citep{fang2021mathematical}. A generalized NTK approach is considered in \citep{chen2020generalized}, inspired by the mean-field approach, they derive a high-probability bound on the generalization error of one-hidden layer NNs with  convergence rate $\mathcal{O}(1/\sqrt{n})$. A high probability generalization error bound with rate $\mathcal{O}(1/\sqrt{n})$ in one-hidden layer NN and with 0--1 and quadratic loss is studied in \citep{nitanda2021particle,nishikawatwo} assuming a mean-field regime. See Appendix \ref{appendix:related work} for further discussion of related work.

\section{Generalization errors}
\label{sec_main_results}
In this section, we apply functional calculus to study generalization error. Our goal is to give abstract conditions under which bounds can be established, which can be applied in a range of contexts, as will be explored in subsequent sections. All proofs are deferred to Appendix~\ref{App: Proofs of GE}.
\subsection{WGE via functional derivatives}
We begin this section by providing a general representation of generalization error, inspired by the approach of \citep[Lemma 7]{Bousquet-Elisseeff}.

\begin{lemma}\label{Lemma: rep gen}
   Consider a generic loss function $(m,z)\mapsto \ell(m,z)$, and $\nu_{n,(1)}$ as defined in \eqref{Eq: nu replace one}. The WGE \eqref{eq:WGE} is given by:
    \begin{align}\label{eq_intro_loss_gap2}
     &\mathrm{gen}(\mfm,\nupop) =\mathbb{E}_{\mathbf{Z}_n,\ols{Z}_1} \Big[\ell\big(\mfm(\nu_n),\ols{Z}_1\big) -   \ell\big(\mfm(\nu_{n,(1)}),\ols{Z}_1\big) \Big]\,.
\end{align} 
\end{lemma}
The right hand side of \eqref{eq_intro_loss_gap2} measures the expected change in the loss function subject to resampling one training data point, which connects generalization error to  stability of $\mfm$ with respect to the training data. We will next quantify this stability precisely, in terms of functional derivatives. 
\begin{assumption}\label{main_ass_square_bounded}
In addition to Assumption \ref{assn:training integrable}, 
\begin{enumerate}[(i)]  
\item\label{main_ass_square_bounded_item1} The loss function $\ell$ is $\mathcal{C}^1$ (cf. Definition \ref{def:flatDerivative}).
    \item\label{main_ass_square_bounded_item4} When restricted to $\nu \in \mathcal{P}_2(\mathcal{Z})$, the training map $\nu\mapsto \mfm(\nu)$ is $\mathcal{C}^1$, in the sense of Definition \ref{Def: func derivative m2m}. (In particular, this requires $\mfm(\nu)\in \mathcal{P}_2(\Theta)$ for all $\nu\in \mathcal{P}_2(\mathcal{Z})$.)
\end{enumerate}
\end{assumption}
\begin{theorem}\label{thm: weak_gen_func} 
Given Assumption \ref{main_ass_square_bounded}, the weak generalization error has the representation
\[
    \mathrm{gen}(\mfm,\nupop)=\mathbb{E}_{\mathbf{Z}_n,\ols{Z}_1}\Big[\int_{0}^1 \int_{\Theta}\Big( \frac{\delta \ell}{\delta m}\big(\mfm_{(1)}(\lambda),\ols{Z}_1,\theta\big)\Big) \big(\mfm(\nu_n)-\mfm(\nu_{n,(1)})\big)(\mrd \theta)\,\mrd \lambda\Big].
\]

\end{theorem}

Note that Theorem \ref{thm: weak_gen_func} holds for all differentiable loss functions, including non-convex cases. Assuming convexity of the loss function with respect to the parameter measure, we can also apply the functional derivative to provide a lower bound on the WGE.

\begin{proposition}\label{WGEconvex}
    Given Assumption \ref{main_ass_square_bounded}, suppose the loss function is convex with respect to $m$, i.e.~for all $m_1,m_2\in\mathcal{P}_2(\Theta)$, and $\lambda\in [0,1]$,
    $\ell\big(\lambda m_1 + (1-\lambda)m_2,z\big)\leq \lambda\ell(m_1,z)+(1-\lambda) \ell(m_2,z)$.
    Then
    \begin{align*}
    &\mathrm{gen}(\mfm,\nupop)\geq\mathbb{E}_{\mathbf{Z}_n,\ols{Z}_1}\Big[\int_{\Theta} \Big(\frac{\delta \ell}{\delta m}(\mfm(\nu_{n,(1)}),\ols{Z}_1,\theta) \Big)\mfm(\nu_n)(\mrd \theta)\Big].
\end{align*}
\end{proposition}
By applying the functional linear derivative to the term $\mfm(\nu_n)-\mfm(\nu_{n,(1)})$ in Theorem \ref{thm: weak_gen_func}, we can provide yet another representation of the WGE.

\begin{theorem}\label{thm: another rep WGE} The WGE can be written 
\begin{equation}\label{Eq:Finite-term}
        \mathrm{gen}(\mfm,\nupop)  =\frac{1}{n}\mathbb{E}_{\mathbf{Z}_n,\ols{Z}_1}\big[ h(\mathbf{Z}_n,\ols{Z}_1)  \big],
       \end{equation} 
       where
\begin{equation*}
    \begin{split}
    h(\mathbf{Z}_n,\ols{Z}_1)  =\int_{0}^1\int_{0}^1\int_{\mathcal{Z}}\Big(\int_{\Theta}  \frac{\delta \ell}{\delta m}\big(\mfm_{(1)}(\lambda),\ols{Z}_1,\theta\big)\frac{\delta \mfm}{\delta \nu}\big(\nu_{(1)}(\tilde{\lambda}),z\big)(\mrd \theta)\Big)\,\big(\delta_{Z_1}-\delta_{\tilde Z_1}\big)(\mrd z) \mrd \tilde{\lambda}\,\mrd \lambda.
    \end{split}
       \end{equation*} 
   \end{theorem}
   
\begin{remark}The representation in \eqref{Eq:Finite-term} reveals that the convergence rate of the WGE is at worst $\mathcal{O}({1}/{n})$ provided that $ \mathbb{E}_{\mathbf{Z}_n,\ols{Z}_1}[ h(\mathbf{Z}_n,\ols{Z}_1)]\leq\mathcal{O}(1)$ with respect to $n\to\infty$.
\end{remark}

\subsection{LGE via functional derivatives}
We now  consider the $L_2$-generalization error $\mathrm{gen}_{L_2}(\mfm, \nu)$. Our first result shows that, similarly to the WGE, the LGE can be controlled in terms of the functional derivatives of $\ell$.

\begin{theorem}\label{thm: square bound FD}
    Given Assumption \ref{main_ass_square_bounded},  the $L_2$-generalization error satisfies
\[\mathrm{gen}_{L_2}(\mfm,\nupop)< \frac{4K}{n} + 2K^{1/2}\mathbb{E}_{\mathbf{Z}_n,\mathbf{\ols{Z}}_2}\big[h_2(\mathbf{Z}_n, \mathbf{\ols{Z}}_2)^2\big]^{1/2} + \mathbb{E}_{\mathbf{Z}_n,\mathbf{\ols{Z}}_2}\big[h_2(\mathbf{Z}_n, \mathbf{\ols{Z}}_2)^2\big],\]
 where
 \[\begin{split}
    h_2(\mathbf{Z}_n, \mathbf{\ols{Z}}_2)&=\int_{0}^1\int_\Theta\int_\mathcal{Z} \Big(\frac{\delta \ell}{\delta m }(\mfm_{(1,2)}(\lambda),z,\theta)\Big)\big(\delta_{Z_1} - \nupop\big)(\mrd z)\big(\mfm(\nu_n)-\mfm( \nu_{n,(1,2)})\big)(\mrd \theta)\, \mrd \lambda,\\
    K&=\max_{Z\in\{Z_1, \tilde Z_1\}}\mathbb{E}\big[\big(\ell(\mfm(\nu_n),Z)\big)^2\big].
\end{split}\]

\end{theorem}
Similarly to Theorem~\ref{thm: another rep WGE}, by applying the functional derivative to $\mfm(\nu_n)-\mfm( \nu_{n,(1,2)})$, we can derive a further upper bound on LGE.

\begin{theorem}\label{thm: another rep LGE}
Given Assumption~\ref{main_ass_square_bounded}, with $K$ as in Theorem \ref{thm: square bound FD}, we have the upper bound on LGE
   \[
    \mathrm{gen}_{L_2}(\mfm,\nupop)\leq \frac{1}{n}\bigg(4K + 2K^{1/2}\mathbb{E}_{\mathbf{Z}_2,\mathbf{\ols{Z}}_2}\Big[\tilde{h}_2(\mathbf{Z}_2, \mathbf{\ols{Z}}_2)^2\Big]^{1/2} + \frac{1}{n}\mathbb{E}_{\mathbf{Z}_2,\mathbf{\ols{Z}}_2}\Big[\tilde{h}_2(\mathbf{Z}_2, \mathbf{\ols{Z}}_2)^2\Big]\bigg),
\]
where, with a slight abuse of notation,
\begin{align*}
    \tilde{h}_2(\mathbf{Z}_n, \mathbf{\ols{Z}}_2)&= \int_{0}^1\int_0^1\int_\Theta \int_\mathcal{Z} \Big(\frac{\delta \ell}{\delta m }(\mfm_{(1,2)}(\lambda),z,\theta)\Big)\big(\delta_{Z_1} - \nupop\big)(\mrd z)\Delta_\mfm(\mathbf{Z}_2, \mathbf{\ols{Z}}_2; \mrd\theta)\, \mrd \lambda\mrd\tilde\lambda,\\
    \Delta_\mfm(\mathbf{Z}_n, \mathbf{\ols{Z}}_2; \mrd\theta)&= \int_{\mathcal{Z}}\Big(\frac{\delta \mfm}{\delta\nu}(\nu_{(1,2)}(\tilde\lambda), z')(\mrd\theta)\Big)\big(\delta_{Z_1} - \delta_{\tilde Z_1} + \delta_{Z_2} - \delta_{\tilde Z_2}\big)(\mrd z').
\end{align*}
\end{theorem}
\begin{remark}
 The term $4K/n$ in the upper bound on LGE in Theorem~\ref{thm: square bound FD} and Proposition~\ref{thm_main_L2_gap_nu_direct}, corresponds loosely to the estimator's bias and is also observed in \citep{Bousquet-Elisseeff,aminian2021information,harutyunyan2021information}.    
\end{remark}
\begin{remark}
Theorem~\ref{thm: another rep LGE} yields an $\mathcal{O}(1/n)$ bound on LGE, provided  $\mathbb{E}_{\mathbf{Z}_n,\mathbf{\ols{Z}_2}}[\tilde{h}^2(\mathbf{Z}_n,\mathbf{\ols{Z}}_2)]\leq\mathcal{O}(1)$ as $n\to \infty$.
\end{remark}

\section{KL-regularized risk minimization}\label{Sec: KL-reg}
In the last section, we gave general conditions under which we can establish convergence rates for generalization error, in terms of the functional derivatives of the loss function and training map. Practically, these conditions are only useful if one can verify that they hold in specific examples. In this section, we will provide an intermediate step in this direction, where we assume the training map $\mfm$ is chosen to minimize a regularized version of the empirical risk. This will provide us with criteria which can be easily verified in practical examples; we will see this fully in Section \ref{Sec: Application}.

We will frequently use $\mdens$ to represent the density of a measure or measure-valued functional, that is, $\mfm(\nu)(\mrd \theta) = \mdens(\nu;\theta)d\theta$. For probability distributions $m$ and $m'$ with positive densities $m_\theta(\theta)$ and $m'_\theta(\theta)$ we define the Kullback--Leibler divergence $\KLr(m'\|m) =\int \log\big(\frac{m'_\theta(\theta)}{m_\theta(\theta)}\big)m_\theta(\theta)\mrd \theta$, and $\KLr(m'\|m)=\infty$ otherwise\footnote{Extending to the case where $m$ and $m'$ are equivalent but do not have (Lebesgue) densities will not be needed here.}. 
We will also write the Kullback--Leibler divergence in terms of the densities, with the abuse of notation $\KLr(m'\|m) = \KLr(m_\theta'\|m_\theta) = \KLr(m'\|m_\theta)$ as convenient.

We will consider training maps which minimize the regularized objective $\mathcal{V}^{\beta,\sigma}$ given by the sum of a risk function and KL-divergence, that is,
\begin{align}\label{Eq: regularized risk}
    &\mathcal{V}^{\beta,\sigma}(m,\nu)= \mathrm{R}(m,\nu)+ \frac{\sigma^2}{2\beta^2} \KLr(m\|\gamma^{\sigma}),
\end{align}
where $R(m,\nu)=\mathbb{E}_{Z\sim \nu}[\ell(m,z)]$ and $\gamma^{\sigma}(\theta) = \frac{1}{F^{\sigma}} \exp\big\{-\frac{1}{\sigma^2} U(\theta)\big\}$
is a Gibbs measure which serves as prior to the parameter measure; here $F^\sigma$ is simply a normalizing constant to ensure $\int \gamma^\sigma(\theta)\mrd\theta =1$, and we call $U:\Theta\to \mathbb{R}$ the `regularizing potential'. 

For notational convenience, we similarly define $\tilde\gamma^\sigma_p = \frac{1}{\tilde F^{\sigma}}\exp\big\{-\frac{1}{\sigma^2}U(\theta) + \|\theta\|^p\big\}$, this density will play a particularly useful role in obtaining our bounds. We know that $F^\sigma, \tilde{F}^\sigma_p<\infty$ under the following assumption, which we will use for our results in this section.

\begin{assumption}\label{ass:KLreg_assn}
For a fixed $p\geq 2$, there exists $g:\mathcal{P}_2(\Theta)\to(0,\infty)$ and $g_1:\mathcal{P}_2(\Theta)\times\Theta\to(0,\infty)$ such that, 
    \begin{enumerate}[(i)]
        \item \label{ass:KLreg_assn_item1} The loss function is $\ell$ is $\mathcal{C}^2$ (Definition \ref{def:flatDerivative}), nonnegative, and convex with respect to $m$;   
        \item \label{ass:KLreg_assn_item3} For all $z\in \mathcal{Z}$ and $m\in \mathcal{P}_p(\Theta)$, the loss satisfies
        \[\big|\ell(m,z) \big| \leq g(m)\big(1+\|z\|^2\big)\]
        and the derivative of the loss satisfies
        \[\Big|\frac{\delta \ell}{\delta m}(m,z,\theta) \Big| \leq g_1(m,\theta)\big(1+\|z\|^2\big);\]
        \item \label{ass:KLreg_assn_item4}  For all $m\in \mathcal{P}_p(\Theta)$, the regularizing potential $U$ satisfies
        $\lim_{\|\theta\|\to \infty}\frac{U(\theta)}{\|\theta\|^p+g_1(m,\theta)} = \infty$;       
        \item \label{ass:KLreg_D2loss_growth_bound}
        
   There exists $C_\theta >0$ such that, for all $m,m'\in \mathcal{P}_p(\Theta)$,
        \begin{equation}
            \label{eq: bounded g ass}
            \mathbb{E}_{\theta\sim m'}\Big[\big(g_1(m, \theta)\big)^2\Big]^{1/2}\leq C_\theta\Big(1+\mathbb{E}_{\theta\sim m'}\big[\|\theta\|^p\big]+\mathbb{E}_{\theta\sim m}\big[\|\theta\|^p\big]\Big),
            \end{equation}
        \item\label{ass:KLreg_D2loss_integrability} We have the pointwise integrability conditions $g(\tilde\gamma_p^\sigma)<\infty$ and, for all $\nu\in \mathcal{P}_2(\mathcal{Z})$ and $m,m'\in \mathcal{P}_p(\Theta)$;
            \begin{equation}\label{eq: bounded second moment ass}
            \mathbb{E}_{\theta,\theta'\sim m}\mathbb{E}_{z\sim \nu}\Big[\Big(\frac{\delta^2 \mathrm{\ell}}{\delta m^2}(m, z, \theta, \theta') \Big)^2\Big]^{1/2}<\infty.
            \end{equation}
    \end{enumerate}
\end{assumption}

We begin by identifying the minimizer of $\mathcal{V}^{\beta,\sigma}$ in the following lemma; the proof, together with others in this section, is deferred to Appendix \ref{app: proofs KL regularized}.

\begin{lemma}\label{lem:GibbsIntegrable}
    Under Assumption \ref{ass:KLreg_assn}, the unique minimizer of $m\mapsto \mathcal{V}^{\beta,\sigma}(m,\nu)$, denoted $\mfm^{\beta,\sigma}(\nu)$ exists, and will be called the \emph{Gibbs measure} associated with $\mathcal{V}^{\beta,\sigma}$. It is the fixed point of $m\mapsto M(m,\nu)$, where $M:\mathcal{P}_p(\Theta)\times \mathcal{P}_2(\mathcal{Z})\to \mathcal{P}_p(\Theta)$ is defined in terms of the density of its output
    \begin{align}\label{Eq: Gibbs measure map}
        &M_\theta(m, \nu;\theta) = \frac{1}{F_{\beta,\sigma}}\exp\Bigg\{-\frac{2\beta^2}{\sigma^2}\Big[\frac{\delta \mathrm{R}}{\delta m}(m, \nu,\theta) + \frac{1}{2\beta^2} U(\theta)\Big] \Bigg\},
    \end{align}
for $F_{\beta,\sigma}$ is a normalization constant. Furthermore, the map $\nu\mapsto \mfm^{\beta,\sigma}(\nu)$ is $\mathcal{C}^1$.
\end{lemma}
We observe that, unless $\frac{\delta \mathrm{R}}{\delta m}$ does not depend on $m$ (i.e.~unless $\mathrm{R}$ is linear in $m$), \eqref{Eq: Gibbs measure map} does not provide an explicit representation of $\mfm^{\beta,\sigma}(\nu_n)$, but instead describes it implicitly in terms of a fixed point. 

Applying Theorems~\ref{thm: another rep WGE} and \ref{thm: another rep LGE}, we can drive upper bounds on the generalization errors.
 \begin{theorem}[WGE of the Gibbs measure]\label{thm: WGE Gibbs KL}
    Given Assumption \ref{ass:KLreg_assn} and $\mathbb{E}_{Z\sim\nupop}\big[\|Z\|^{8}\big]<\infty$, the weak generalization error satisfies 
\[  |\mathrm{gen}(\mfm(\nu_n),\nupop)|\leq \frac{c}{n}\frac{2\beta^2}{\sigma^2} \mathbb{E}_{Z_1}\Big[\Big(1+\|Z_1\|^2\Big)^4\Big]\]
where $c=\frac{9\sqrt{2}}{2}C_\theta^2  \Big[1+ 2 \int_{\Theta}\|\theta\|^p\tilde\gamma_p^\sigma(\mrd \theta) + 4\frac{g(\tilde\gamma^\sigma_p)}{n}\Big]^2 >0$.
 \end{theorem}

\begin{theorem}[LGE of the Gibbs measure]\label{thm: LGE upper KL}
    Given Assumption \ref{ass:KLreg_assn} and $\mathbb{E}_{Z\sim\nupop}\big[\|Z\|^{16}\big]<\infty$, the $L_2$ generalization error satisfies 
  \begin{equation}
\begin{split}
    &\mathrm{gen}_{L_2}(\mfm^{\beta,\sigma}(\nu_n),\nupop) \leq \frac{1}{n} \Big(4K + 2K^{1/2} A^{1/2}+\frac{A}{n} \Big),
    \end{split}
\end{equation}
where \[\begin{split}
&A= (2\times 5^4)\Big(\frac{2\beta^2}{\sigma^2}\Big)^2C_\theta^4\Big(1+2\int_{\Theta}\|\theta\|^p\tilde\gamma_p^\sigma(\mrd \theta)+ 8g(\tilde\gamma^\sigma_p)\Big)^4\mathbb{E}_{Z_1}[(1+\|Z_1\|^2)^{8}]<\infty,
\end{split}\] and $K=\max_{Z\in\{Z_1, \tilde Z_1\}}\mathbb{E}\big[\big(\ell(\mfm^{\beta,\sigma}(\nu_n),Z)\big)^2\big]$.
\end{theorem}

\textbf{Expected population risk upper bound:} Using Theorem~\ref{thm: WGE Gibbs KL}, we can also provide an upper bound on the expected population risk. Suppose that there exists $\bar m \in \mathcal{P}_2(\Theta)$ such that $ \mathrm{R}(\bar m,\nu_n)=0$, cf.~\citep[Appendix A]{nishikawatwo}. From the optimization defining $\mfm^{\beta,\sigma}$, we know that
\begin{equation} \label{nishikawatwo bound 1}
\mathrm{R}(\mfm^{\beta,\sigma}(\nu_n),\nu_n) \leq  \mathcal{V}^{\beta,\sigma}(\mfm^{\beta,\sigma}(\nu_n),\nu_n) \leq \mathcal{V}^{\beta,\sigma}(\bar m,\nu_n) = \frac{\sigma^2}{2\beta^2} \KLr(\bar m \|\gamma^{\sigma}) \,.
\end{equation}
Hence
\begin{equation}\label{eq:Rdecomposition}
\mathrm{R}(\mdens^{\beta,\sigma}(\nu_n),\nupop)  \leq \mathrm{R}(\mdens^{\beta,\sigma}(\nu_n),\nupop) - \mathrm{R}(\mdens^{\beta,\sigma}(\nu_n),\nu_n)   + \frac{\sigma^2}{2\beta^2} \KLr(\bar m\|\gamma^{\sigma}) \,.    
\end{equation}

Theorem~\ref{thm: WGE Gibbs KL} implies that
\[\begin{split}
\Big|\mathbb{E}_{\mathbf{Z}_n}\big[\mathrm{R}(\mdens^{\beta,\sigma}(\nu_n),\nupop) \big]\Big| &\leq \Big|\mathbb{E}_{\mathbf{Z}_n}\Big[\mathrm{R}(\mdens^{\beta,\sigma}(\nu_n),\nupop) - \mathrm{R}(\mdens^{\beta,\sigma}(\nu_n),\nu_n)\Big]\Big| + \frac{\sigma^2}{2\beta^2} \KLr(\bar m \|\gamma^{\sigma})\\
&\leq \frac{c}{n}\frac{2\beta^2}{\sigma^2}  \mathbb{E}_{Z_1}\Big[\big(1+\|Z_1\|^2\big)^4\Big]+ \frac{\sigma^2}{2\beta^2} \KLr(\bar m \|\gamma^{\sigma})\,.
\end{split}
\]
Consequently, setting $\beta$ to be of order $n^{1/4}$ and using Theorem \ref{thm: WGE Gibbs KL}, we deduce that expected population risk, $\big|\mathbb{E}_{\mathbf{Z}_n}\big[\mathrm{R}(\mdens^{\beta,\sigma}(\nu_n),\nupop) \big]\big|$, is of order $n^{-1/2}$. This agrees in principle with the order of error computed in \citep{nitanda2021particle} for specific cases.

\textbf{$L_2$ population risk upper bound:} Similarly, squaring and taking the expected value of \eqref{eq:Rdecomposition}, and using Theorem \ref{thm: LGE upper KL}, indicates that taking $\beta$ to be of order $n^{1/6}$ we have that  $\mathbb{E}_{\mathbf{Z}_n}\big[ \big|\mathrm{R}(\mdens^{\beta,\sigma}(\nu_n),\nupop) \big|^2\big]$, is of order $n^{-2/3}$.

\section{Application: Overparameterized one-hidden layer NN}\label{Sec: Application}

We now highlight the application of our model to an overparameterized one-hidden layer neural network in the mean-field regime~\citep{MFLD,mei/montanari/nguyen:2018}.
The loss function for an overparameterized one-hidden layer NN is represented in \eqref{eq: nn loss function}. For simplicity, we also write $(\hat y,y)\mapsto \ell_o(\hat y,y)$ for the loss function, where $\hat y = \mathbb{E}_{\theta\sim m}[\phi(\theta,x)]$ is the predicted output of the overparameterized one-hidden layer NN.
We assume that the risk function $\mathrm{R}$ is given by $ \mathrm{R}(\mfm(\nu_n),\nupop)= \mathbb{E}_{(X,Y) \sim \nupop}[\ell_o(\mathbb{E}_{\theta\sim \mfm^{\beta,\sigma}(\nu_n)}[\phi(\theta,X)],Y)]$, as is done in mean-field models of one-hidden layer neural networks \citep{MFLD,mei/misiakiewicz/montanari:2019,tzen/raginsky:2020}, and is KL-regularized as seen previously. We make the following assumptions on an overparameterized one-hidden layer NN to investigate generalization performance:
\begin{assumption}\label{ass:NN_KL}
For a constant $p\geq 2$, the regularizing potential $U$ satisfies $\lim_{\|\theta\|\to \infty}\frac{U(\theta)}{\|\theta\|^p} = \infty,$ the loss $\ell_o$ is convex and nonnegative, and  there exist finite constants  $L_\ell, L_{\ell,1}, L_{\ell,2}$, and $L_\phi$ such that, for all $(x,y)\in\mathcal{X}\times \mathcal{Y}$ and $\theta\in\Theta$, 
         \[\begin{split}
          &\big|\ell_o(\hat y, y)\big| \leq L_\ell(1+\|\hat y\|^2 + \|y\|^2),\qquad|\partial_{\hat{y}}\ell_o(\hat y,y)\big|\leq L_{\ell,1}(1+\|\hat{y}\|+\|y\|),\qquad
         \big|\partial_{\hat{y}\hat{y}}\ell_o(\hat y,y)\big|\leq L_{\ell,2},\\
           \end{split}\]
           and
           \[\big|\phi(\theta,x)\big|\leq L_{\phi}(1+\|x\|)(1+\|\theta\|^2).\]
\end{assumption}

\begin{remark}[Activation function examples]
     Recall that $\phi(\theta,x)=a\varphi(w\cdot x)$, where $a$ is the parameter of  the outer layer and $\varphi(\cdot )$ is the activation function. If the activation function is of linear growth, then Assumption \ref{ass:NN_KL} is satisfied. Note that the ReLU, Heaviside unit-step, tanh and sigmoid activation functions are of this type, and no smoothness assumption is needed.
 \end{remark}

\begin{remark}[Loss function examples]
    Twice differentiable loss functions which are of quadratic growth with derivatives of linear growth satisfy Assumption~\ref{ass:NN_KL}. For example, the quadratic, product-type (as in \citep{bartlett2006convexity}), and logcosh loss functions~\citep{wang2022comprehensive} are of this type. 
\end{remark}

We begin by verifying Assumption~\ref{ass:KLreg_assn}, for one-hidden layer neural network model under Assumption~\ref{ass:NN_KL}, in the following lemma; the proof, together with others in this section, is deferred to Appendix~\ref{App: proofs of Sec: one hidden example}. While the constants are not given precisely here, they can be explicitly determined with only moderate effort, if so desired.
\begin{lemma}\label{lem: equ with main ass NN}
Under Assumption~\ref{ass:NN_KL}, there exists a constant $c>0$ independent of $n$ such that, for all $z\in\mathcal{Z}$,  $\theta\in \Theta$, $m\in \mathcal{P}_p(\Theta)$, and $\nu\in \mathcal{P}_2(\mathcal{Z})$,
\begin{align*}
\big|\ell(m,z) \big| &\leq c\Big[\mathbb{E}_{\theta\sim m}[1+\|\theta\|^2]^2\Big]\big(1+\|z\|^2\big),\\
\Big|\frac{\delta \ell}{\delta m}(m,z,\theta) \Big| &\leq c\Big[1+\|\theta\|^4 + \mathbb{E}_{\theta'\sim m}[\|\theta'\|^4]\Big]\big(1+\|z\|^2\big),\\
\mathbb{E}_{\theta,\theta'\sim m}\mathbb{E}_{z\sim \nu}\Big[\Big(\frac{\delta^2 \mathrm{\ell}}{\delta m^2}(m, z, \theta, \theta') \Big)^2\Big]^{1/2}&\leq c\Big[\mathbb{E}_{\theta\sim m}[1+\|\theta\|^4]\Big]\mathbb{E}_{Z\sim \nu}[1+\|Z\|^2] ,\\
K=\max_{Z\in \{Z_1, \ols Z_1\}}\mathbb{E}\big[\big(\ell(\mfm(\nu_n), Z)\big)^2\big]&\leq c\,\mathbb{E}_{Z\sim \nupop}\Big[(1+\|Z\|^2)^{6}\Big].
\end{align*}
In particular, if $p\ge 8$ in Assumption~\ref{ass:NN_KL}, then Assumption~\ref{ass:KLreg_assn} is satisfied for some choice of $C_\theta<\infty$, and if $\mathbb{E}_{Z\sim \nupop}[\|Z\|^{12}]<\infty$ then $K<\infty.$
\end{lemma}

\begin{remark}
    With stronger assumptions on the loss and activation function, it is possible (using the same techniques as in Lemma \ref{lem: equ with main ass NN}) to reduce the order of the moment assumption which needs to be made on $Z$.
\end{remark}

We now derive upper bounds on WGE and LGE of the overparameterized one hidden-layer NN by directly combining Lemma~\ref{lem: equ with main ass NN} with Theorems~\ref{thm: WGE Gibbs KL} and \ref{thm: LGE upper KL}, respectively. 

\begin{proposition}[WGE of overparameterized one-hidden layer NN]\label{Prop: WGE upper NN}
Given Assumption~\ref{ass:NN_KL} with $p\ge 8$, and $\mathbb{E}_{Z\sim\nupop}\big[\|Z\|^{8}\big]<\infty$, there exists a constant $c>0$ such that
\[|\mathrm{gen}(\mfm^{\beta,\sigma}(\nu_n),\nupop)|\leq  \frac{c\beta^2}{n}.\]
\end{proposition}

\begin{proposition}[LGE  of overparameterized one-hidden layer NN] \label{thm_main_L2_gap_nu_direct}

Given Assumption~\ref{ass:NN_KL} with $p\ge 8$, and $\mathbb{E}_{Z\sim\nupop}\big[\|Z\|^{16}\big]<\infty$, there exists a constant $c>0$ such that
 \begin{equation*}
\begin{split}
    &\mathrm{gen}_{L_2}(\mfm^{\beta,\sigma}(\nu_n),\nupop) \leq \frac{c\beta^2}{n}.
    \end{split}
\end{equation*}
\end{proposition}

It is instructive to compare our results to those of \citet{nishikawatwo,nitanda2021particle}. There, building on \citep[Lemma~5.5]{chen2020generalized} and Rademacher complexity analysis, the authors study generalization error for one-hidden layer NN for binary classification and square loss regression problems. The authors make the key assumption that there exists $\bar m\in \mathcal{P}(\Theta)$ such that $ \mathrm{R}(\bar m,\nu_n)=0$. We see in \citep[Appendix A]{nishikawatwo},  that there is a constant $c>0$ such that, for $\delta \in (0,1)$,
 with probability  $1-\delta$ we have
 \begin{align}\label{Eq: bound nintanda}
     \Big(\mathrm{R}(\mdens^{\beta,\sigma}(\nu_n),\nupop) - \mathrm{R}(\mdens^{\beta,\sigma}(\nu_n),\nu_n)\Big)
     \leq
     \frac{c}{\sqrt{n}} 
     \Big(
  \sqrt{\KLr(\bar m\|\gamma^{\sigma})}
      + 
      \sqrt{ \log(1/\delta) }
      \Big)\,.
 \end{align} 
In contrast, our bounds are non-asymptotic and hold under verifiable assumptions on model inputs. 

\section{Conclusions and future works}
\label{sec_conclusions}
Our study introduces a novel framework for analyzing the generalization error of risk functions. The framework utilizes calculus on the space of probability measures, which allows us to gain a deeper understanding of the factors that influence the generalization of machine learning models. We demonstrate the efficacy of our framework by applying it to learning problems within the mean-field regime. This application results in the derivation of new bounds for overparameterized models that are particularly relevant to one-hidden layer neural networks.

 In future work, it is promising to extend our methods to establish upper bounds for weak and $L_2$ generalization errors of deep neural networks, via the mean-field analysis outlined in \citep{sirignano/spiliopoulos:2019}, \citet{chizat2022infinite}, \citet{jabir2019mean} and \citet{geshkovski2023emergence}. Additionally, we intend to apply our framework to a broader range of examples, beyond mean-field models, while relaxing the regularization requirements. Our goal is to develop comparable generalization and stability bounds for specific algorithms. 

    \section*{Acknowledgements} Authors acknowledge the support of the UKRI Prosperity Partnership Scheme (FAIR) under EPSRC Grant EP/V056883/1 and the Alan Turing Institute. Samuel N. Cohen  and {\L}ukasz Szpruch also acknowledge the support of the Oxford--Man Institute for Quantitative Finance.
    
    We also thank Joe Colvin (former PhD student at MAC-MIGS - EPSRC-funded doctoral school in Mathematical Modelling, Analysis and Computation based in Edinburgh) for his contributions to the initial draft of this manuscript.

 \bibliography{Refs}
 \newpage

\clearpage

\appendix

\section{Other related works}\label{appendix:related work}

\paragraph{Expected generalization error:} The pioneering work in establishing upper bounds on the expected generalization error through an information-theoretic approach was done by \citet{russo2019much} and \citet{xu2017information}. They proposed using the mutual information between the input training set and the output parameters. Building upon their work, \citet{bu2020tightening} introduced tighter upper bounds that consider the mutual information of individual samples. \citet{asadi2018chaining} suggested the use of chaining mutual information, while \citet{hafez2020conditioning} and \citet{steinke2020reasoning} advocated conditioning and processing techniques. In addition to mutual information, other researchers have explored alternative information measures for establishing information-theoretic bounds on generalization error. These measures include $f$-divergences, $\alpha$-Rényi divergence, and generalized Jensen-Shannon divergence, as demonstrated in the works of \citet{esposito2019generalization}, \citet{aminian2022learning}, and \citet{aminian2021information}. Wasserstein distance-based upper bounds, which consider the discrepancy between the training sample data and the output of a randomized learning algorithm, were proposed by \citet{lopez2018generalization}, \citet{wang2019information}, \citet{galvez2021tighter}, and \citet{aminian2022tighter}. Furthermore, some studies have explored the application of rate-distortion theory to analyze generalization error, as seen in the works of \citet{masiha2023f} and \citet{sefidgaran2022rate}. In our approach, we employ calculus on probability spaces, allowing us to apply it to both the overparameterized regime and the expected parametric loss scenarios, to derive an upper bound on the weak generalization error and $L_2$ generalization error.

\paragraph{Generalization error and the parametric Gibbs measure:} The mean-field perspective on training gives close connections with ideas from statistical mechanics. The parametric Gibbs measure (a.k.a. Gibbs Algorithm or Gibbs Posterior) first introduced by \citep{Gibbs1902elementary} in statistical mechanics and further studied by \citep{jaynes1957information} describes the stable behaviour of an ensemble of particles evolving subject to a potential, as is given by our (parametric) loss function. The weak generalization error of the parametric Gibbs measure with expected parametric loss has been studied from information-theoretic and PAC-Bayesian perspectives. An information-theoretic upper bound on weak (expected) generalization error, with a convergence rate of $\mathcal{O}(1/n)$ is provided in \citep{raginsky2016information} and \citep{aminian2021exact,kuzborskij2019distribution} for the parametric Gibbs measure, assuming a bounded or sub-Gaussian parametric loss function, respectively. PAC-Bayesian upper bounds for parametric Gibbs measures are also derived in \citep{alquier2016properties, aminian2021exact}. The sensitivity of the parametric Gibbs measure is studied in \citep{perlaza2022empirical}. Our work differs from this body of research by reparameterization of the loss function in terms of a parameter measure, which results in a general Gibbs measure and can be applied to overparameterized one-hidden layer NN under the mean-field regime. We also provide an upper bounds on the expected generalization error of the parametric Gibbs measure with weaker assumptions.

\paragraph{Other approaches to generalization error:} There are other approaches applied to study generalization error of general learning problems, including  VC dimension-based bounds, Rademacher complexity, PAC-Bayesian bounds, and algorithmic stability-based bounds. A group of studies \citep{neyshabur2015norm,neyshabur2017pac,golowich2017size,arora2018stronger} has investigated the generalization error of neural networks via a uniform convergence approach. However, these results are algorithm independent. Rademacher complexity of functions classes like neural networks have been studied by \citep{bartlett2017spectrally,yin2019rademacher} to provide upper bound on generalization error. These results based on VC dimension and Rademacher complexity are often vacuous in NN scenarios~\citep{dziugaite2017computing}. A data dependent Rademacher complexity bound is studied for overparameterized NNs by \citet{bubeck2023universal} which has a convergence rate of $\mathcal{O}(1/\sqrt{n})$. Another approach is based on PAC-Bayes learning \citep{mcallester2003pac,catoni2007pac}.  In this approach high probability bounds are derived on generalization error in terms of different information measures~\citep{rivasplata2020pac,amit2022integral,tolstikhin2013pac}. Some works also studied the stability-based bounds for generalization error~\citep{Bousquet-Elisseeff,hardt2016train,mou2017generalization,chen2018stability}. These approaches are not applied to derive the generalization error upper bound in overparameterized neural networks by considering the regularization.

\paragraph{Benign overfitting:} A body of literature has emerged to investigate the occurrence of benign overfitting in overparameterized models, wherein accurate predictions are made on unseen test data despite overfitting to the training data \citep{bartlett2020benign}. Although this phenomenon has been explored in various studies, most of them have been confined to linear models \citep{belkin2019two, bartlett2020benign, hastie2019surprises, wu2020optimal, cao2021risk} and kernel/random feature models \citep{belkin2018understand, liang2020just, montanari2020interpolation}. Recently, \citet{li2021towards} have examined the benign overfitting phenomenon in random feature models, which are defined as two-layer neural networks with fixed random initialization parameters in the first layer. Furthermore, the study by \citet{cao2022benign} demonstrates that benign overfitting can also occur in two-layer convolutional neural networks. In contrast, our research focuses on the generalization error in the mean-field regime, whereas these works assume that the input data dimension is large enough with respect to the number of parameters, maintaining a constant ratio to observe benign overfitting.

\paragraph{Mean-field:} Our study employs the mean-field framework utilized in a recent line of research \citep{chizat/bach:2018,mei/montanari/nguyen:2018,mei/misiakiewicz/montanari:2019,fang2019over,fang2021mathematical,sirignano/spiliopoulos:2019,MFLD}. \citet{chizat/bach:2018} establishes the convergence of gradient descent for training one-hidden layer NNs with infinite width under certain structural assumptions. The study of \citet{mei/montanari/nguyen:2018} proves the global convergence of noisy stochastic gradient descent and establishes approximation bounds between finite and infinite neural networks. Furthermore, \citet{mei/misiakiewicz/montanari:2019} demonstrates that this approximation error can be independent of the input dimension in certain cases, and establishes that the residual dynamics of noiseless gradient descent are close to the dynamics of NTK-based kernel regression under some conditions. The mean-field approach is mostly restricted to one-hidden layer NNs and the extension to Deep NNs is not trivial~\citep{sirignano/spiliopoulos:2019,chizat2022infinite}. \citet{fang2019over} proposes a new concept known as neural feature repopulation inspired by the mean field view. Lastly, the mean-field approach's performance with respect to other methods in feature learning~\citep{fang2021mathematical} suggests it is a viable option for analyzing one-hidden layer NNs. The focus of our study is based on the work conducted by \citet{MFLD} in the area of non-linear functional minimization with KL regularization. Their investigation demonstrated the linear convergence (in continuous time) of the resulting mean-field Langevin dynamics under the condition of sufficiently robust regularization.

\section{Preliminaries}
\label{app_calculus}

\paragraph{Properties of functional derivatives}

\begin{proof}[Proof of Lemma \ref{Lemma: sam}]
    We know $\frac{\delta\mfm}{\delta \nu}$ is a Borel signed measure. For any Lebesgue null set $A$, we know $\mfm(\nu)(A)=0$ for all $\nu$ and, by uniqueness of the functional derivative, conclude $\frac{\delta \mfm}{\delta \nu}(a,\nu)(A)=0$. Therefore, $\frac{\delta \mfm}{\delta \nu}(a,\nu)(A)$ is absolutely continuous with respect to Lebesgue measure, and by the Radon--Nikodym theorem we have
    \begin{align}
        \frac{\delta \mfm}{\delta \nu}(a,\nu)(A) = \int_A G(a,\nu,x)dx,
    \end{align}
for some function $G$ integrable in $x$ and measurable in $a,\nu$. We now need to show that $G(a,\nu,x)$ is a version of $\frac{\delta U}{\delta \nu }$ for almost all $x$. 
For all $A$, we know 
\begin{align}
    \frac{1}{\text{Leb}(A)}\int_A G(a,\nu,x)dx&=\frac{1}{\text{Leb}(A)} \frac{\delta \mfm}{\delta \nu}(A)=\frac{1}{\text{Leb}(A)}\frac{\delta}{\delta \nu}\Big[\int_A U(a,\nu,x) dx\Big]\\
    &=\frac{\delta}{\delta \nu}\Big[\frac{1}{\text{Leb}(A)}\int_A U(a,\nu,x) dx\Big], 
\end{align}
where $\text{Leb}(A)=\int_A 1 dx$. Taking $A$ small, we conclude (by linearity of the functional derivative) that $G$ is a version of $\frac{\delta U}{\delta \nu}$ for every $x$ which is a Lebesgue point of $G$ and $U$, that is, for almost all $x$. 
\end{proof}

 \section{Proofs and details from Section \ref{sec_main_results}}\label{App: Proofs of GE}

 \begin{proof}[Proof of Lemma \ref{Lemma: rep gen}]
  Recall that $\ols{Z}_1\sim \nupop$ is independent of $\{Z_i\}_{i=1}^n$. Since elements of $\{Z_i\}_{i=1}^n$ are i.i.d., the WGE \eqref{eq:WGE} can be written 
\begin{equation}\label{eq_linear_functionaU_lderivative_linear_intro}
 \begin{split}
    \mathrm{gen}(\mfm(\nu_n),\nupop)&=  \mathbb{E}_{\mathbf{Z_n}}\big[ \mathrm{R}(\mfm(\nu_n),\nupop)-\mathrm{R}(\mfm(\nu_n),\nu_n) \big] 
    \\
     &= \mathbb{E}_{\mathbf{Z_n}}\Big[ \int_{\mathcal Z} \ell(\mfm(\nu_n),z)\nupop(\mrd z) -  \int_{\mathcal Z} \ell(\mfm(\nu_n),z)\nu_n(\mrd z) \Big]
     \\
    &= \mathbb{E}_{\mathbf{Z_n}}\Big[\mathbb E_{\ols{Z}_1}[ \ell(\mfm(\nu_n),\ols{Z}_1)] -   \frac{1}{n}\sum_{i=1}^n\ell(\mfm(\nu_n),Z_i) \Big]
    \\
   &= \mathbb{E}_{\mathbf{Z_n}}\Big[\mathbb E_{\ols{Z}_1}[ \ell(\mfm(\nu_n),\ols{Z}_1) ]-   \ell(\mfm(\nu_n),Z_1) \Big].
     \end{split}
\end{equation}
Recall that in \eqref{Eq: nu replace one} we defined the perturbation 
$\nu_{n,(1)}:= \nu_n + \frac{1}{n} \big( \delta_{\ols{Z}_1} - \delta_{Z_1}\big),$ which corresponds to a data measure with one different data point, and that, as $Z_1$ and $\ols{Z}_1$ are i.i.d., we have
\begin{equation}\label{Eq: LOO}
    \mathbb{E}_{\mathbf{Z_n}}\big[  \ell(\mfm(\nu_n),Z_1) \big] = 
\mathbb{E}_{\mathbf{Z_n}}\mathbb{E}_{\ols{Z}_1}\big[ \ell(\mfm(\nu_{n,(1)}),\ols{Z}_1) \big]\,.
\end{equation}
Combining this observation with \eqref{eq_linear_functionaU_lderivative_linear_intro} yields the representation \eqref{eq_intro_loss_gap2}.
\end{proof}

\begin{proof}[Proof of Theorem \ref{thm: weak_gen_func}]
Recall that from \eqref{mperturb1},  $\nu_{n,(1)} = \nu_n + \frac{1}{n} (\delta_{\ols{Z}_1} - \delta_{Z_1})$ and $
    \mfm_{(1)}(\lambda) = \mfm(\nu_n) + \lambda\left( \mfm(\nu_{n,(1)}) - \mfm(\nu_n)\right).
$
Using the definition of the linear functional derivative (Definition~\ref{def:flatDerivative}) and Lemma \ref{Lemma: rep gen} we have
\[
\begin{split}
    \mathrm{gen}(\mfm,\nu)
     &=\mathbb{E}_{\mathbf{Z_n}}\mathbb E_{\ols{Z}_1} \Big[\ell(\mfm(\nu_n),\ols{Z}_1) -   \ell(\mfm(\nu_{n,(1)}),\ols{Z}_1) \Big]\,\\
     &=\mathbb{E}_{\mathbf{Z_n}}\mathbb{E}_{\ols{Z}_1}\Big[ \int_{0}^1   \int_{\Theta}  \Big(\frac{\delta\ell}{\delta m}(\mfm_{(1)}(\lambda),\ols{Z}_1,\theta)\Big)\big(\mfm(\nu_n)-\mfm(\nu_{n,(1)})\big)(\mrd \theta)\,\mrd \lambda\Big].
    \end{split}
\]  
\end{proof}

\begin{proof}[Proof of Proposition \ref{WGEconvex}]
    As $\ell$ is convex and $\mathcal{C}^1$, we know that , for any $m_1,m_2\in \mathcal{P}_2(\Theta)$,
    \begin{align*}
        &\ell\big(\lambda m_1 + (1-\lambda)m_2,z\big)\\
         & \leq \lambda\ell(m_1,z)+(1-\lambda) \ell(m_2,z)\\
        & = \ell(m_2,z)+ \lambda\Big(\ell(m_1,z)-\ell(m_2,z)\Big) \\
        &= \ell(m_2,z)+\lambda \int_{0}^1 \int_{\Theta}\Big( \frac{\delta \ell}{\delta m}\big(\lambda' m_1 + (1-\lambda')m_2,z,\theta\big)\Big) \big(m_1-m_2\big)(\mrd \theta)\,\mrd \lambda'.
    \end{align*}
We also know that 
\begin{align*}
    &\ell\big(\lambda m_1 + (1-\lambda)m_2,z\big)\\
    & = \ell(m_2,z) + \lambda \int_0^1 \int_{\Theta}\Big( \frac{\delta \ell}{\delta m}\big(\lambda' (\lambda m_1 + (1-\lambda)m_2) + (1-\lambda')m_2,z,\theta\big)\Big) \big(m_1-m_2\big)(\mrd \theta)\,\mrd \lambda',
\end{align*}
from which we conclude 
\begin{align*}
    &\int_{0}^1 \int_{\Theta}\Big( \frac{\delta \ell}{\delta m}\big(\lambda' m_1 + (1-\lambda')m_2,z,\theta\big)\Big) \big(m_1-m_2\big)(\mrd \theta)\,\mrd \lambda' \\
    &\geq  \int_{0}^1\int_{\Theta}\Big( \frac{\delta \ell}{\delta m}\big(\lambda' (\lambda m_1 + (1-\lambda)m_2) + (1-\lambda')m_2,z,\theta\big)\Big) \big(m_1-m_2\big)(\mrd \theta)\,\mrd \lambda'\\
    &=  \frac{1}{1-\lambda}\int_{0}^{1-\lambda}\int_{\Theta}\Big( \frac{\delta \ell}{\delta m}\big(\lambda' m_1 + (1-\lambda')m_2,z,\theta\big)\Big) \big(m_1-m_2\big)(\mrd \theta)\,\mrd \lambda'
\end{align*}
for all $\lambda\in[0,1)$. Observe that the RHS of this inequality is essentially the same as the left hand side, but averaging over a smaller set of $\lambda'$ values. Repeating this argument and using continuity of $\frac{\delta \ell}{\delta m}$  with respect to $m$, we conclude that, for every $k>0$,
\begin{align*}
    &\int_{0}^1 \int_{\Theta}\Big( \frac{\delta \ell}{\delta m}\big(\lambda' m_1 + (1-\lambda')m_2,z,\theta\big)\Big) \big(m_1-m_2\big)(\mrd \theta)\,\mrd \lambda' \\
    &\geq \frac{1}{(1-\lambda)^k}\int_{0}^{(1-\lambda)^k}\int_{\Theta}\Big( \frac{\delta \ell}{\delta m}\big(\lambda' m_1 + (1-\lambda')m_2,z,\theta\big)\Big) \big(m_1-m_2\big)(\mrd \theta)\,\mrd \lambda'\\
    &\to \int_{\Theta}\Big( \frac{\delta \ell}{\delta m}\big(m_2,z,\theta\big)\Big) \big(m_1-m_2\big)(\mrd \theta) \quad \text{as }k\to \infty.
\end{align*}
Substituting back, we obtain
\[\ell\big(\lambda m_1 + (1-\lambda)m_2,z\big) - \ell\big(m_2,z\big)\geq \lambda \int_{\Theta}\Big( \frac{\delta \ell}{\delta m}\big(m_2,z,\theta\big)\Big) \big(m_1-m_2\big)(\mrd \theta).\]
In particular, taking $\lambda\to 1$ and using continuity, 
\[\ell\big(m_1 ,z\big) - \ell\big(m_2,z\big)\geq \int_{\Theta}\Big( \frac{\delta \ell}{\delta m}\big(m_2,z,\theta\big)\Big) \big(m_1-m_2\big)(\mrd \theta).\]
By definition of the derivative, we know
\[\int_{\Theta}\Big( \frac{\delta \ell}{\delta m}\big(m_2,z,\theta\big)\Big) m_2(\mrd \theta) = 0,\]
so this term can be omitted.
The result now follows from Theorem \ref{thm: weak_gen_func}, with $m_1 = \mfm(\nu_n)$ and $m_2 = \mfm(\nu_{n,(1)})$.
\end{proof}

\begin{proof}[Proof of Theorem \ref{thm: another rep WGE}]
        
        We know from \eqref{Eq: nu replace one} that 
        \[\nu_n - \nu_{n,(1)} = \frac{1}{n}\big(\delta_{Z_1}-\delta_{\ols{Z}_1}\big).\]
        Applying the functional derivative to $\mfm$  and recalling the definition of $\nu_{(1)}(\tilde\lambda)$ in \eqref{mperturb1}, for any Borel set $B\subset \Theta$ we have 
        \[\begin{split}
            \big(\mfm(\nu_n)-\mfm( \nu_{n,(1)})\big)(B) &= \int_0^1\int_{\mathcal{Z}} \Big(\frac{\delta \mfm}{\delta\nu}(\nu_{(1)}(\tilde\lambda), z)(B)\Big)(\nu_n- \nu_{n,(1,2)})(\mrd z)\mrd\tilde\lambda\\
            &= \int_0^1\Big(\frac{\delta \mfm}{\delta\nu}(\nu_{(1)}(\tilde\lambda), Z_1)(B)-\frac{\delta \mfm}{\delta\nu}(\nu_{(1)}(\tilde\lambda), \ols{Z}_1)(B)\Big)\mrd\tilde\lambda.
        \end{split}\]
        We now subsitute this expression for $\mfm(\nu_n)-\mfm(\nu_{n,(1)})$ in Theorem \ref{thm: weak_gen_func}. The result follows immediately.
        \end{proof}
        
\begin{proof}[Proof of Theorem \ref{thm: square bound FD}]
We have
\begin{equation} \label{eq_L2_gen_bound_pf_1}
    \begin{split}
     \mathrm{R}(\mfm(\nu_n),\nupop)-\mathrm{R}(\mfm(\nu_n),\nu_n)
     =  \frac{1}{n} \sum_{i=1}^n \Big(\int_{\mathcal Z}  \ell(\mfm(\nu_n),z) \nupop(\mrd z) - \ell(\mfm(\nu_n),Z_i)\Big)\,.
    \end{split}
\end{equation}
Define
\begin{equation} \label{eq_L2_gen_bound_pf_2}
    \alpha_i:= \int_{\mathcal Z}  \ell(\mfm(\nu_n),z) \nupop(\mrd z)  - \ell(\mfm(\nu_n),Z_i),
\end{equation}
which gives
\begin{equation} \label{eq_L2_gen_bound_pf_3}
      \Big(\mathrm{R}(\mfm(\nu_n),\nupop)-\mathrm{R}(\mfm(\nu_n),\nu_n)\Big)^2
\, =\, \frac{1}{n^2} \Big(\sum_{i=1}^n (\alpha_i)^2  + 2\sum_{i_1<i_2}\alpha_{i_1} \alpha_{i_2} \Big)   \,.
\end{equation}
For indices $i_1, i_2\in [n]$, we define
\begin{equation}\label{eq_L2_gen_bound_pf_4}
\begin{split}
    \nu_{n,(i_1,i_2)} & :=\nu_n + \frac{1}{n}\Big( (\delta_{\ols{Z}_{i_1}} -\delta_{Z_{i_1}}) + (\delta_{\ols{Z}_{i_2}} -\delta_{Z_{i_2}}) \Big), \\
     \alpha_{i,(i_1,i_2)} & := \int_{\mathcal Z}  \ell(\mfm(\nu_{n,(i_1,i_2)} ),z) \nupop(\mrd z) -\ell(\mfm(\nu_{n,(i_1,i_2)}),Z_i)\,.
    \end{split}
    \end{equation}
One can then write 
\begin{equation}\label{eq_L2_gen_bound_pf_5}
\begin{split}
    \sum_{i_1<i_2}\alpha_{i_1} \alpha_{i_2} &=
    \sum_{i_1<i_2}\Big\{
   (\alpha_{i_1} - \alpha_{i_1,(i_1,i_2)})(\alpha_{i_2} - \alpha_{i_2,(i_1,i_2)}) 
   +  (\alpha_{i_1} - \alpha_{i_1,(i_1,i_2)})\alpha_{i_2,(i_1,i_2)} \\
   &\quad+  (\alpha_{i_2} - \alpha_{i_2,(i_1,i_2)})\alpha_{i_1,(i_1,i_2)}
   +  \alpha_{i_{1},(i_1,i_2)} \alpha_{i_{2},(i_1,i_2)}\Big\}\,.
\end{split}
\end{equation}
Note that $\alpha_{i_1
    ,(i_1,i_2)}$ and $\alpha_{i_2
    ,(i_1,i_2)}$ are conditionally independent given the $\sigma$-algebra $\mathcal F^{(i_1,i_2)}=\sigma\big(\{Z_i\}_{i=1}^n  / \{Z_{i_1},Z_{i_2} \}\big)$. Hence, by the tower property of conditional expectation,
\begin{equation}\label{eq_L2_gen_bound_pf_6}
    \mathbb E\big[ \alpha_{i_1,(i_1,i_2)} \alpha_{i_{2},(i_{1},i_{2})}    \big] = \mathbb E \Big[ \mathbb E\big[ \alpha_{i_1,(i_1,i_2)} | \mathcal F^{(i_1,i_2)} \big] \mathbb E\big[ \alpha_{i_2,(i_1,i_2)} | \mathcal F^{(i_1,i_2)} \big]\Big]= 0\,.
\end{equation}
Using the fact that $\{Z_i\}_{i=1}^n$ are i.i.d.~and applying the Cauchy--Schwarz inequality,
\begin{equation}\label{eq_L2_gen_bound_pf_7_new}
     \begin{split}
     &\mathbb E\Big[\Big( \mathrm{R}\big(\mfm(\nu_n),\nupop)-\mathrm{R}(\mfm(\nu_n),\nu_n)\Big)^2 \Big]\\
&\leq \frac{1}{n}  \mathbb{E}[  (\alpha_1)^2] + \frac{2}{n^2}\sum_{i_1<i_2}\bigg[
 \mathbb E\big[(\alpha_{i_1} - \alpha_{i_1,(i_1,i_2)})^2\big]^{1/2} \mathbb E\big[(\alpha_{i_2} - \alpha_{i_2,(i_1,i_2)})^2\big]^{1/2} \\
  & \qquad+  \mathbb E \big[(\alpha_{i_1} - \alpha_{i_1,(i_1,i_2)})^2\big]^{1/2} \mathbb E\big[ (\alpha_{i_2,(i_1,i_2)})^2\big]^{1/2} \\
& \qquad  + \mathbb E\big[  (\alpha_{i_2} - \alpha_{i_2,(i_1,i_2)})^2\big]^{1/2} \mathbb E\big[ (\alpha_{i_1,(i_1,i_2)} )^2\big]^{1/2} \bigg].
     \end{split}
\end{equation}
It remains to estimate all the terms on the right hand side. By Assumption~\ref{main_ass_square_bounded} (and the elementary inequality $(a-b)^{2} \leq 2(a^{2}+b^{2})$) 
\begin{equation}\label{eq_L2_gen_bound_pf_8_new}
     \begin{split}
    \mathbb{E}\big[  (\alpha_1)^2 \big] &= \mathbb{E}\Big[\Big( \int_{\mathcal Z}  \ell(\mfm(\nu_n),z) \nu(\mrd z)  - \ell(\mfm(\nu_n),Z_1)\Big)^2 \Big]\\
    &\leq 2\mathbb{E}\Big[  \Big(\int_{\mathcal Z}  \ell(\mfm(\nu_n),z)\nupop(\mrd z) \Big)^2 \Big] + 2\mathbb{E}\big[ \big(\ell(\mfm(\nu_n),Z_1) \big)^2\big]\\
    &\leq 2\mathbb{E}\Big[  \int_{\mathcal Z}  \ell(\mfm(\nu_n),z)^2\nupop(\mrd z) \Big] + 2\mathbb{E}\big[ \big(\ell(\mfm(\nu_n),Z_1) \big)^2\big]\\
    &\leq 2\mathbb{E}\Big[ \big( \ell(\mfm(\nu_n),\tilde Z_1)\big)^2\Big] + 2\mathbb{E}\big[ \big(\ell(\mfm(\nu_n),Z_1) \big)^2\big]\leq 4K\,.
     \end{split}
\end{equation}
Observe that $\nu_n$ and $\nu_{n,(i_1, i_2)}$ are both empirical measures from iid samples.

Therefore $\int_{\mathcal Z}  \ell(\mfm(\nu_n),z) \nupop(\mrd z)$ and $\int_{\mathcal Z}  \ell(\mfm(\nu_{n,(i_1, i_2)}),z) \nupop(\mrd z)$ are equal in law. Similarly, as $Z_i$ does not appear in the construction of $\nu_{n,(i_1, i_2)}$, we know $\ell(\mfm(\nu_{n,(i_1,i_2)}),Z_i)$ and $\ell(\mfm(\nu_{n}),\tilde Z_1)$ are equal in law. Therefore, by essentially the same calculation as above, $\mathbb E\big[ (\alpha_{i_1,(i_1,i_2)} )^2\big]\leq 4K$.

To study $\mathbb E\big[\big(\alpha_{i} - \alpha_{i,(i_1,i_2)}\big)^2\big]$ for $i\in\{i_1,i_2\}$, 
first notice that, given our i.i.d.~assumption and symmetry, we just need to study $\mathbb E\big[\big(\alpha_{1} - \alpha_{1,(1,2)}\big)^2\big]$. We first express the integrals in terms of expectations over the independent value $\ols{Z}_3$,
\begin{equation*}
    \begin{split}
    & \mathbb{E}[(\alpha_{1} - \alpha_{1,(1,2)})^2] 
   \\
   &= 
   \mathbb{E}_{\mathbf{Z_n},\mathbf{\ols{Z}_2}}\bigg[ \bigg(\int_{\mathcal{Z}}
    \ell(\mfm(\nu_n),z)- \ell(\mfm(\nu_{n,(1,2)} ),z)\nupop(\mrd z)  - \Big(  \ell(\mfm(\nu_n),Z_1)  - \ell(\mfm(\nu_{n,(1,2)} ),Z_1)  \Big) \bigg)^2 \bigg]
    \\
    &    = \mathbb{E}_{\mathbf{Z_n},\mathbf{\ols{Z}_2}}\bigg[ \bigg(\mathbb{E}_{\ols{Z}_3}\Big[
    \ell(\mfm(\nu_n),\ols{Z}_3)- \ell(\mfm(\nu_{n,(1,2)} ),\ols{Z}_3)\Big]  - \Big(  \ell(\mfm(\nu_n),Z_1)  - \ell(\mfm(\nu_{n,(1,2)} ),Z_1)  \Big)  \bigg)^2 \bigg].
\end{split}
\end{equation*}
We next expand the two central terms using the functional derivative of $\ell$, to get
\begin{equation}\label{eqL2boundpf8shifted}
\begin{split}
    & \mathbb{E}[(\alpha_{1} - \alpha_{1,(1,2)})^2] \\
    & =\mathbb{E}_{\mathbf{Z_n},\mathbf{\ols{Z}_3}}\bigg[ \bigg(
    \int_{0}^1 \int_{\Theta} \Big( \mathbb E_{\ols{Z}_3} \Big[  \frac{\delta\ell}{\delta m}(\mfm_{(1,2)}(\lambda_1),\ols{Z}_3,\theta) \Big]- \frac{\delta\ell}{\delta m}(\mfm_{(1,2)}(\lambda_1),Z_i,\theta) \Big)\\
    &\qquad\qquad\qquad  \times \big(\mfm(\nu_n)-\mfm( \nu_{n,(1,2)})\big)(\mrd \theta) \mrd  \lambda_1 \bigg)^2 \bigg]\\
    &=\mathbb{E}_{\mathbf{Z}_n,\mathbf{\ols{Z}}_2}\big[h_2(\mathbf{Z}_n, \mathbf{\ols{Z}}_2)^2\big],
    \end{split}
    \end{equation}
    where $h_2(\mathbf{Z}_n, \mathbf{\ols{Z}}_2)$ is as given in the theorem statement. Returning to the estimate \eqref{eq_L2_gen_bound_pf_7_new}, noting that the double sum consists of $n(n-1)/2<n^2/2$ terms, and combining with \eqref{eqL2boundpf8shifted} and \eqref{eq_L2_gen_bound_pf_8_new}, we have
\[   \mathrm{gen}_{L_2}(\mfm,\nupop)
    < \frac{4K}{n} + 2K^{1/2}\mathbb{E}_{\mathbf{Z}_n,\mathbf{\ols{Z}}_2}\big[h_2(\mathbf{Z}_n, \mathbf{\ols{Z}}_2)^2\big]^{1/2} + \mathbb{E}_{\mathbf{Z}_n,\mathbf{\ols{Z}}_2}\big[h_2(\mathbf{Z}_n, \mathbf{\ols{Z}}_2)^2\big].\]
\end{proof}

\begin{proof}[Proof of Theorem \ref{thm: another rep LGE}]

We know from \eqref{Eq: nu replace one} that 
\[\nu_n - \nu_{n,(1,2)} = \frac{1}{n}\big(\delta_{Z_1}-\delta_{\ols{Z}_1}+ \delta_{Z_2}-\delta_{\ols{Z}_2}\big).\]
Applying the functional derivative to $\mfm$  and recalling the definition of $\nu_{(1,2)}(\tilde\lambda)$ in \eqref{mperturb1}, for any Borel set $B\subset \Theta$ we have 
\[\begin{split}
    \big(\mfm(\nu_n)-\mfm( \nu_{n,(1,2)})\big)(B) &= \int_0^1\int_{\mathcal{Z}} \Big(\frac{\delta \mfm}{\delta\nu}(\nu_{(1,2)}(\tilde\lambda), z)(B)\Big)(\nu_n- \nu_{n,(1,2)})(\mrd z)\mrd\tilde\lambda\\
    &= \int_0^1\Big(\frac{\delta \mfm}{\delta\nu}(\nu_{(1,2)}(\tilde\lambda), Z_1)(B)-\frac{\delta \mfm}{\delta\nu}(\nu_{(1,2)}(\tilde\lambda), \ols{Z}_1)(B)\\
    &\qquad +\frac{\delta \mfm}{\delta\nu}(\nu_{(1,2)}(\tilde\lambda), Z_2)(B)-\frac{\delta \mfm}{\delta\nu}(\nu_{(1,2)}(\tilde\lambda), \ols{Z}_2)(B)\Big)\mrd\tilde\lambda\\
    &= \int_0^1\Big(\Delta_\mfm(\mathbf{Z}_2, \mathbf{\ols{Z}}_2; B)\Big)\mrd\tilde\lambda.
\end{split}\]

Substituting into the definition of $h_2$ in Theorem \ref{thm: square bound FD} and applying Fubini's theorem,
\[
\begin{split}
&h_2(\mathbf{Z}_n, \mathbf{\ols{Z}}_2)\\
    &=\frac{1}{n}\int_{0}^1\int_0^1\int_\Theta \Big(\mathbb E_{\ols{Z}_3} \Big[  \frac{\delta\ell}{\delta m}(\mfm_{(1,2)}(\lambda_1),\ols{Z}_3,\theta) \Big]-\frac{\delta \ell}{\delta m }(\mfm_{(1,2)}(\lambda),Z_1,\theta)\Big)\Delta_\mfm(\mathbf{Z}_2, \mathbf{\ols{Z}}_2; \mrd\theta)\, \mrd \lambda\mrd\tilde\lambda\\
    &= \frac{1}{n}\tilde{h}_2(\mathbf{Z}_n, \mathbf{\ols{Z}}_2).
\end{split}
\]
The result follows by rearrangement and Theorem \ref{thm: square bound FD}.
\end{proof}

 \section{Proofs from Section \ref{Sec: KL-reg}}\label{app: proofs KL regularized}

\begin{proof}[Proof of Lemma \ref{lem:GibbsIntegrable}]
    Our first task is to verify that $m\mapsto M_\theta(m, \nu;\cdot)$ is a valid map from $\mathcal{P}_p(\Theta)$ to (densities of measures in) $\mathcal\mathcal{P}_p(\Theta)$, for any $\nu\in \mathcal{P}_2(\mathcal{Z})$.

    Given the form of \eqref{Eq: Gibbs measure map}, it is enough to verify that for fixed $m,\nu$, for some $c>0$, and all $\theta\in \Theta$ sufficiently large,  we have the inequality
    \begin{equation}\label{eq:Gibbsgrowth}
        -\frac{2\beta^2}{\sigma^2}\Big[\frac{\delta \mathrm{R}}{\delta m}(m, \nu,\theta) + \frac{1}{2\beta^2} U(\theta)\Big] \leq -c\|\theta\|^p.
    \end{equation}
Assumption \ref{ass:KLreg_assn}(\ref{ass:KLreg_assn_item3}) guarantees that 
\[\begin{split}\frac{\delta\ell}{\delta m}(m,z,\theta) 
&\leq g_1(m,\theta)\big(1+\|z\|^2\big) ,\end{split}\]

and hence, as $\nu\in \mathcal{P}_2(\mathcal{Z})$, our definition \eqref{eq risk linear} guarantees that for some $C'>0$ we have 
\[\begin{split}
    \frac{\delta \mathrm{R}}{\delta m}(m, \nu,\theta) &=\int_\mathcal{Z} \frac{\delta \ell}{\delta m}(m,z, \theta)\nu(\mrd z) \leq (1+C')g_1(m,\theta).
\end{split}\]

On the other hand, with the same $C'$ as above, Assumption \ref{ass:KLreg_assn}(\ref{ass:KLreg_assn_item4}) guarantees that for all $\theta$ sufficiently large, we have 
\[\begin{split}
    \frac{1}{2\beta^2}U(\theta)&\geq  (1+C')\big(\|\theta\|^p+g_1(m,\theta)\big) \geq (1+C')\|\theta\|^p + \frac{\delta \mathrm{R}}{\delta m}(m, \nu,\theta).\end{split}\] 
Substituting in \eqref{eq:Gibbsgrowth} gives the desired bound. We therefore know that $M$ is a well defined map $\mathcal{P}_p(\Theta)\times \mathcal{P}_2(\mathcal{Z})\to \mathcal{P}_p(\Theta)$.

It is easy to check that, under Assumption \ref{ass:KLreg_assn}, $\mathcal{V}^{\beta,\sigma}(\gamma^\sigma, \nu)$ is finite, and that $V^{\beta, \sigma}(m,\nu)\geq 0$ for all $m\in \mathcal{P}_2(\Theta)$. Consequently,  no minimizer of $\mathcal{V}^{\beta,\sigma}$ can be outside the set $\{m:\frac{\sigma^2}{2\beta^2}\KLr(m\|\gamma^\sigma)\leq \mathrm{R}(\gamma^\sigma, \nu)\}\subset\mathcal{P}_p(\Theta)$, and this set is relatively compact (in the weak topology). By convexity, (as the Kullback--Leibler divergence is strictly convex, and $\mathrm{R}$ is convex), we conclude that $\mathcal{V}^{\beta,\sigma}(\cdot, \nu)$ has a unique minimizer, which will have a density (due to the Kullback--Leibler divergence). Taking the first order condition for optimality derived in \citep[Proposition 2.5]{MFLD} , the minimizer of $\mathcal{V}^{\beta,\sigma}(\cdot, \nu_n)$ is uniquely characterized, for some normalizing constant $F$ and almost all $\theta$, by
\[\begin{split}
    0 = \frac{\delta R}{\delta m}(m,\nu,\theta) + \frac{1}{2\beta^2}U(\theta) + \frac{\sigma^2}{2\beta^2}\log(m_\theta(\theta))+F
\end{split}\]
which rearranges to give $m_\theta(\theta)=M_\theta(m,\nu;\theta)$, as desired. We defer the proof of differentiability to Lemma \ref{lem_parameter_measure_derivative}
\end{proof}

\begin{lemma}\label{lem:optimizerinBnu}
    Defining the set valued map 
        \[\begin{split}B(\nu) &:= \Big\{m\in \mathcal{P}_p(\Theta): \frac{\sigma^2}{2\beta^2}\mathrm{KL}(m\|\gamma^\sigma)\leq \mathrm{R}(\gamma^\sigma, \nu)\\
        &\qquad \text{ and } \int_\Theta \|\theta\|^pm(\mrd \theta)\leq \mathrm{R}(\tilde\gamma^\sigma,\nu) + \int_\Theta \|\theta\|^p\tilde\gamma^\sigma(\theta)\mrd\theta\Big\}\subset \mathcal{P}_2(\Theta),\end{split}\]
        for any $\nu\in \mathcal{P}_2(\mathcal{Z})$, the KL-regularized risk minimizer satisfies $\mfm^{\beta,\sigma}(\nu) \in B(\nu)$.
\end{lemma}
\begin{proof}
We have already seen in Lemma \ref{lem:GibbsIntegrable} that $\mfm^{\beta,\sigma}(\nu)\in \{\frac{\sigma^2}{2\beta^2}\KLr(m\|\gamma^\sigma)\leq \mathrm{R}(\gamma^\sigma, \nu)\}$.
It remains to show that $\mfm^{\beta,\sigma}(\nu)$ satisifes the stated $p$-moment bound.

    Recall that we defined the measure $\tilde \gamma^\sigma_p$ through its density $(\tilde F^{\alpha,\beta})^{-1}\exp(-\frac{1}{2\beta^2}U(\theta) + \|\theta\|^p)$, for $\tilde F^{\alpha,\beta}$ a normalization constant. Consider the problem of minimizing $\tilde{\mathrm{R}}(m,\nu) + \frac{\sigma^2}{2\beta^2}\mathrm{KL}(m\|\tilde{\gamma}^\sigma)$, where $\tilde{\mathrm{R}}(m,\nu) = \mathrm{R}(m,\nu) + \int_{\Theta}\|\theta\|^pm(\mrd \theta)$. As this falls in the class considered in Lemma \ref{lem:GibbsIntegrable}, we know that its optimizer exists and its density is the fixed point of 
    \[\begin{split}
        m_\theta&\mapsto \frac{1}{\tilde{F}_{\beta,\sigma}}\exp\Bigg\{-\frac{2\beta^2}{\sigma^2}\Big[\frac{\delta \tilde{\mathrm{R}}}{\delta m}(m, \nu,\theta) + \frac{1}{2\beta^2} U(\theta) - \|\theta\|^p\Big] \Bigg\}\\
         &= \frac{1}{\tilde{F}_{\beta,\sigma}}\exp\Bigg\{-\frac{2\beta^2}{\sigma^2}\Big[\frac{\delta \mathrm{R}}{\delta m}(m, \nu,\theta) + \frac{1}{2\beta^2} U(\theta)\Big] \Bigg\}.
    \end{split}\]
    In other words, $m = \mfm^{\beta,\sigma}(\nu)$ as defined in Lemma \ref{lem:GibbsIntegrable}. It follows that 
    \[\mathrm{R}(\mfm^{\beta,\sigma}(\nu),\nu) + \int_{\Theta}\|\theta\|^p\mfm^{\beta,\sigma}(\nu;\mrd \theta) + \frac{\sigma^2}{2\beta^2}\mathrm{KL}(\mfm^{\beta,\sigma}(\nu)\|\tilde{\gamma}_p^\sigma)\leq \mathrm{R}(\tilde\gamma_p^\sigma,\nu) + \int_{\Theta}\|\theta\|^p\tilde\gamma_p^\sigma(\mrd \theta).\]
    As $\ell$, and hence $\mathrm{R}$, is nonnegative, we conclude that 
    \[\int_{\Theta}\|\theta\|^p\mfm^{\beta,\sigma}(\nu;\mrd \theta) \leq \mathrm{R}(\tilde\gamma_p^\sigma,\nu) + \int_{\Theta}\|\theta\|^p\tilde\gamma_p^\sigma(\mrd \theta).\]
\end{proof}

\begin{lemma}\label{lem:Cmonotone}
    For any choice of $m\in B(\nu)$, the  linear map $\mathcal{C}_m:L^2(m,\Theta) \to L^2(m,\Theta)$ defined by 
    \[\mathcal{C}_{m} f(\theta) := \mathbb{E}_{\theta'\sim m}\Big[\frac{\delta^2 \mathrm{R}}{\delta m^2}(m, \nu, \theta, \theta') f(\theta')\Big]\]
    is positive, in the sense that
    \[\langle f, \mathcal{C}_{m}f\rangle_{L^2(m,\Theta)} = \int_\Theta f(\theta) (\mathcal{C}_{m} f)(\theta) m(\mrd\theta)\geq 0.\]
    In particular, $\mathcal{C}_{m}$ is a Hilbert--Schmidt operator with discrete spectrum 
    \[\sigma(\mathcal{C}_{m}) = \{\lambda^\mathcal{C}_i\}_{i\ge 0} \subset [0,\infty).\]
\end{lemma}
\begin{proof}
    We first write 
    \[\begin{split}\int_\Theta f(\theta) (\mathcal{C}_{m} f)(\theta) m(\mrd\theta) &= \int_\Theta\int_\Theta f(\theta) f(\theta') \frac{\delta^2 \mathrm{R}}{\delta m^2}(m, \nu, \theta, \theta') m(\mrd \theta) m(\mrd \theta')\\
    &= \int_\mathcal{Z}\int_\Theta\int_\Theta f(\theta) f(\theta') \frac{\delta^2 \mathrm{\ell}}{\delta m^2}(m, z, \theta, \theta') m(\mrd \theta) m(\mrd \theta')\nu(\mrd z).
    \end{split}\]
    As $\ell$ is convex with respect to $m$, we know that $\frac{\delta^2 \ell}{\delta m^2}$ is positive semidefinite, which gives positivity of $\mathcal{C}_m$. We also have the symmetry property, \citep[Lemma 2.4]{cardaliaguet2019master}
    \[\frac{\delta^2 \mathrm{\ell}}{\delta m^2}(m, z, \theta, \theta')= \frac{\delta^2 \mathrm{\ell}}{\delta m^2}(m, z, \theta', \theta).\]
    We next recall that, by Assumption \ref{ass:KLreg_assn}(\ref{ass:KLreg_D2loss_integrability}),
    \[\int_\mathcal{Z}\int_\Theta\int_\Theta \Big(\frac{\delta^2 \mathrm{\ell}}{\delta m^2}(m, z, \theta, \theta') \Big)^2 m(\mrd \theta) m(\mrd \theta')\nu(\mrd z)<\infty,\]
    which implies $\mathcal{C}_{m}$ is Hilbert--Schmidt, in particular it is a self-adjoint compact operator. The spectral theorem then states that $\mathcal{C}_{m}$ has a discrete spectrum made up of its eigenvalues, which are nonnegative as $\mathcal{C}_{m}$ is positive.
\end{proof}

Our next challenge is to prove that the Gibbs measure is differentiable. We begin by considering an auxiliary function $S$.
\begin{lemma}\label{lem:Sdifferentiable}
Define 
\begin{equation}\label{eq:Sdefn}
    S(\nu, \theta) :=\frac{\delta\mathrm{R}}{\delta m}(\mfm^{\beta,\sigma}(\nu),\nu,\theta) = \int_\mathcal{Z} \frac{\delta \ell}{\delta m}(\mfm^{\beta,\sigma}(\nu), z, \theta)\nu(\mrd z).
\end{equation}
    Under Assumption \ref{ass:KLreg_assn}, we know $S$ is differentiable in $\nu$, and its derivative satisfies the bound
    \[\begin{split}&\int_\Theta\Big(\frac{\delta S}{\delta \nu}(\nu, \theta, z)\Big)^2 \mfm^{\beta,\sigma}(\nu; \mrd \theta)\\& \leq\int_\Theta\Big( \frac{\delta\ell}{\delta m}(\mfm^{\beta,\sigma}(\nu),\theta,z) - \int_{\mathcal{Z}}\frac{\delta\ell}{\delta m}(\mfm^{\beta,\sigma}(\nu),\theta,z')\nu(\mrd z')\Big)^2 \mfm^{\beta,\sigma}(\nu; \mrd \theta).\end{split}\]
     In particular, we have the representation:
    \[\begin{split}    \frac{\delta S}{\delta \nu}(\nu,\theta, z)&= \frac{\delta \ell}{\delta m}\big(\mfm^{\beta,\sigma}(\nu),z,\theta\big) - \int_{\mathcal{Z}}\frac{\delta \ell}{\delta m}\big(\mfm^{\beta,\sigma}(\nu),z',\theta\big)\nu(\mrd z') \\
    &\quad -\frac{2\beta^2}{\sigma^2}\mathrm{Cov}_{\theta'\sim \mfm^{\beta,\sigma}(\nu)}\Big[ \int_{\mathcal{Z}}\frac{\delta^2 \ell}{\delta m^2}(\mfm^{\beta,\sigma}(\nu), z', \theta, \theta')\nu(\mrd z') ,\, \frac{\delta S}{\delta\nu}(\nu,\theta',z)\Big].
\end{split}    \]
\end{lemma}
\begin{proof}
Fix $\nu, \nu_1 \in \mathcal{P}_2(\mathcal{Z})$ and write $\nu^{\epsilon}=\nu+\epsilon(\nu_1-\nu)$.
Recall that
\begin{equation}\label{eq:Sdefn1}
    S(\nu, \theta) :=\frac{\delta\mathrm{R}}{\delta m}(\mfm^{\beta,\sigma}(\nu),\nu,\theta) = \int_\mathcal{Z} \frac{\delta \ell}{\delta m}(\mfm^{\beta,\sigma}(\nu), z, \theta)\nu(\mrd z)
\end{equation}
and $\mfm^{\beta,\sigma}(\nu)$ satisfies
\begin{equation}\label{mQreplemma}
\mfm^{\beta,\sigma}_\theta(\nu;\theta) = \frac{Q(\nu, \theta)}{\int_\Theta Q(\nu, \theta')\mrd\theta'},
\end{equation}
where $Q(\nu,\theta) =  \exp\Big\{-\frac{2\beta^2}{\sigma^2}\Big(S(\nu,\theta) + \frac{1}{2\beta^2}U(\theta)\Big)\Big\}.$ 
We define an approximate directional derivative of $S$:
\[\frac{\delta^\epsilon S}{\delta\nu^\epsilon}(\nu^\epsilon,\theta):=\frac{S(\nu^\epsilon,\theta)-S(\nu,\theta)}{\epsilon},\]
and similarly
\[\frac{\delta^{\epsilon}\mfm^{\beta,\sigma}_\theta}{\delta \nu^{\epsilon}}(\nu^{\epsilon}):=\frac{\mdens^{\beta,\sigma}(\nu^\epsilon)-\mdens^{\beta,\sigma}(\nu)}{\epsilon}, \qquad \frac{\delta^\epsilon Q}{\delta \nu^\epsilon}:=\frac{Q(\nu^\epsilon;\theta)-Q(\nu;\theta)}{\epsilon}.\]
We note that $S(\nu^\epsilon, \theta) \to S(\nu, \theta)$ as $\epsilon \to 0$. Using the fact that $|S(\nu', \theta)|<C(1+\|\theta\|^2)$ for all $\nu'$ in a compact set around $\nu$, the dominated convergence theorem shows that,  as $\epsilon \to 0$,
\begin{equation}\label{Sboundcrude}
\epsilon\Big\|\frac{\delta^\epsilon S}{\delta \nu^\epsilon}\Big\|_{L^2(\mfm^{\sigma, \beta}(\nu))} = \|S(\nu^\epsilon,\cdot) - S(\nu, \cdot)\|_{L^2(\mfm^{\sigma, \beta}(\nu))} \to 0. 
\end{equation}

We now give a representation of $\delta^\epsilon Q/\delta \nu^\epsilon$. Expanding the definition of the approximate directional derivative,
\begin{equation}\label{Qderivlemma}
\begin{split}
&\frac{\delta^\epsilon Q}{\delta \nu^\epsilon}(\nu^\epsilon, \theta, z) \\
&=\frac{1}{\epsilon}  \left( \exp\Big\{-\frac{2\beta^2}{\sigma^2}\Big(S(\nu^\epsilon,\theta) + \frac{1}{2\beta^2}U(\theta)\Big)\Big\}-\exp\Big\{-\frac{2\beta^2}{\sigma^2}\Big(S(\nu,\theta) + \frac{1}{2\beta^2}U(\theta)\Big)\Big\}\right)\\
&=\frac{1}{\epsilon} Q(\nu, \theta)  \left( \exp\Big\{-\frac{2\beta^2}{\sigma^2}\Big(S(\nu^\epsilon,\theta) - S(\nu,\theta))\Big)\Big\}-1\right)\\
&=\frac{1}{\epsilon} Q(\nu, \theta)  \left( \exp\Big\{-\frac{2\beta^2}{\sigma^2} \,\epsilon \,\frac{\delta^\epsilon S}{\delta\nu^\epsilon}(\nu^\epsilon,\theta)\Big\}-1\right)\\
&= 
-\frac{2\beta^2}{\sigma^2}Q(\nu, \theta) \frac{\delta^\epsilon S}{\delta\nu^\epsilon}(\nu^\epsilon,\theta)\int_{0}^1 \exp\Big\{ \frac{-2\beta^2}{\sigma^2}(\epsilon\lambda \frac{\delta^\epsilon S}{\delta\nu^\epsilon}(\nu^\epsilon,\theta,z))\Big\}\mrd \lambda\\
&
=-\frac{2\beta^2}{\sigma^2}Q(\nu, \theta) \frac{\delta^\epsilon S}{\delta\nu^\epsilon}(\nu^\epsilon,\theta) G(\nu^\epsilon,\theta),
\end{split}
\end{equation}
where 
\[\begin{split}G(\nu^\epsilon,\theta)&:=\int_{0}^1 \exp\Big\{ \frac{-2\beta^2}{\sigma^2}(\epsilon\lambda \frac{\delta^\epsilon S}{\delta\nu^\epsilon}(\nu^\epsilon,\theta))\Big\}\mrd \lambda\\
&=\int_{0}^1 \exp\Big\{ \frac{-2\beta^2}{\sigma^2}\lambda(S(\nu^\epsilon, \theta)-S(\nu,\theta) )\Big\}\mrd \lambda.
\end{split}\]
Observe that \eqref{Sboundcrude} implies $\|G(\nu^\epsilon, \theta)- 1\|_{L^2(\mfm^{\beta, \sigma}(\nu))}\to 0$ as $\epsilon \to 0$.

Next, we reexpress $\delta^\epsilon S/\delta \nu^\epsilon$ in terms of derivatives of $\mathrm{R}(m,\nu) = \int_{\mathcal{Z}}\ell(m,z)\mrd z$. Note that
\[\frac{\delta\mathrm{R}}{\delta m}(\mfm^{\beta,\sigma}(\nu^\epsilon),\nu^\epsilon,\theta)-\frac{\delta\mathrm{R}}{\delta m}(\mfm^{\beta,\sigma}(\nu^\epsilon),\nu,\theta)=\epsilon\int_{0}^1 \int_{\mathcal{Z}}\frac{\delta\ell}{\delta m}(\mfm^{\beta,\sigma}(\nu^\epsilon),\theta,z)(\nu_1-\nu)(\mrd z)\mrd \lambda,\]
and 
\[\frac{\delta\mathrm{R}}{\delta m}(\mfm^{\beta,\sigma}(\nu^\epsilon),\nu,\theta)-\frac{\delta\mathrm{R}}{\delta m}(\mfm^{\beta,\sigma}(\nu),\nu,\theta)=\int_{\Theta} \frac{\delta^2\mathrm{R}}{ \delta m^2}(\mfm^{\beta,\sigma}_{\lambda}(\nu^\epsilon),\nu,\theta,\theta')(\mfm^{\beta,\sigma}(\nu^\epsilon)-\mfm^{\beta,\sigma}(\nu))(\mrd \theta'),\]
where 
$\mfm^{\beta,\sigma}_{\lambda}(\nu^\epsilon):=\mfm^{\beta,\sigma}(\nu)+\lambda(\mfm^{\beta,\sigma}(\nu^\epsilon)-\mfm^{\beta,\sigma}(\nu))$.
Hence
\begin{equation}\label{Sderivlemma}\begin{split}
    & \frac{\delta^\epsilon S}{\delta\nu^\epsilon}(\nu^\epsilon,\theta,z)\\
    &=\frac{1}{\epsilon} \left( \frac{\delta\mathrm{R}}{\delta m}(\mfm^{\beta,\sigma}(\nu^\epsilon),\nu^\epsilon,\theta)-\frac{\delta\mathrm{R}}{\delta m}(\mfm^{\beta,\sigma}(\nu),\nu,\theta) \right)\\
    &=\frac{1}{\epsilon} \left( \frac{\delta\mathrm{R}}{\delta m}(\mfm^{\beta,\sigma}(\nu^\epsilon),\nu^\epsilon,\theta)-\frac{\delta\mathrm{R}}{\delta m}(\mfm^{\beta,\sigma}(\nu^\epsilon),\nu,\theta)+\frac{\delta\mathrm{R}}{\delta m}(\mfm^{\beta,\sigma}(\nu^\epsilon),\nu,\theta)-\frac{\delta\mathrm{R}}{\delta m}(\mfm^{\beta,\sigma}(\nu),\nu,\theta)\right)\\
    &= \int_{\mathcal{Z}}\frac{\delta\ell}{\delta m}(\mfm^{\beta,\sigma}(\nu^\epsilon),\theta,z)(\nu_1-\nu)(\mrd z)
    +\int_{0}^1\int_{\Theta} \frac{\delta^2\mathrm{R}}{ \delta m^2}(\mfm^{\beta,\sigma}_{\lambda}(\nu^\epsilon),\nu,\theta,\theta')\frac{\delta^{\epsilon}\mfm^{\beta,\sigma}_\theta}{\delta \nu^{\epsilon}}(\nu^{\epsilon}, \theta', z)(\mrd \theta')\mrd \lambda.
\end{split}
\end{equation}

Given these calculations, we can now give a characterization of $\delta \mfm_\theta^{\beta,\sigma}/\delta \nu^\epsilon$, making use of \eqref{mQreplemma}:
\[\begin{split}
&\frac{\delta^{\epsilon}\mfm^{\beta,\sigma}_\theta}{\delta \nu^{\epsilon}}(\nu^{\epsilon}, \theta)\\
&=
\frac{1}{\epsilon} \left( \frac{Q(\nu^\epsilon, \theta)}{\int_\Theta Q(\nu^\epsilon, \theta')\mrd\theta'}-\frac{Q(\nu, \theta)}{\int_\Theta Q(\nu, \theta')\mrd\theta'}\right)\\
    &=\frac{Q(\nu^\epsilon, \theta)\int_\Theta Q(\nu, \theta')\mrd\theta'-Q(\nu, \theta)\int_\Theta Q(\nu^\epsilon, \theta')\mrd\theta'}{\epsilon \int_\Theta Q(\nu^\epsilon, \theta')\mrd\theta' \int_\Theta Q(\nu, \theta')\mrd\theta'}\\
    &=\frac{Q(\nu^\epsilon, \theta)\Big(\int_\Theta Q(\nu, \theta')\mrd\theta'-\int_\Theta Q(\nu^\epsilon, \theta')\mrd\theta'\Big)+\int_\Theta Q(\nu^\epsilon, \theta')\mrd\theta'\Big(Q(\nu^\epsilon, \theta)-Q(\nu, \theta)\Big)}{\epsilon \int_\Theta Q(\nu^\epsilon, \theta')\mrd\theta' \int_\Theta Q(\nu, \theta')\mrd\theta'}\\
     &=\frac{Q(\nu^\epsilon, \theta)\Big( \frac{1}{\epsilon}\int_\Theta \left(Q(\nu, \theta')- Q(\nu^\epsilon, \theta') \right)\mrd\theta'\Big)+\int_\Theta Q(\nu^\epsilon, \theta')\mrd\theta'\Big( \frac{1}{\epsilon}\left( Q(\nu^\epsilon, \theta)-Q(\nu, \theta)\right) \Big)}{\int_\Theta Q(\nu^\epsilon, \theta')\mrd\theta' \int_\Theta Q(\nu, \theta')\mrd\theta'}\\
    &= -\frac{\mdens^{\beta,\sigma}(\nu^\epsilon;\theta)\Big(\int_\Theta \frac{\delta^{\epsilon}Q}{\delta \nu^{\epsilon}}(\nu^{\epsilon},\theta')\mrd\theta'\Big)}{ \int_\Theta Q(\nu, \theta')\mrd\theta'} 
    + 
    \frac{ \frac{\delta^{\epsilon}Q}{\delta \nu^{\epsilon}}(\nu^{\epsilon},\theta)}{ \int_\Theta Q(\nu, \theta')\mrd\theta'}
   \\
    &=-\frac{2\beta^2}{\sigma^2}\Big(\mdens^{\beta,\sigma}(\nu;\theta) \frac{\delta^\epsilon S}{\delta\nu^\epsilon}(\nu^\epsilon,\theta)G(\nu^\epsilon,\theta)-\mdens^{\beta,\sigma}(\nu^\epsilon;\theta)\Big[\int_\Theta \mdens^{\beta,\sigma}(\nu;\theta')\frac{\delta^\epsilon S}{\delta\nu^\epsilon}(\nu^\epsilon,\theta')G(\nu^\epsilon,\theta')\mrd \theta'\Big]\Big),
\end{split}\]
where the last equality follows from \eqref{Qderivlemma}. For any sufficiently integrable function $f:\Theta\to \mathbb{R}$, we can therefore compute
\[\begin{split}
    &\int_\Theta f(\theta) \frac{\delta^{\epsilon}\mfm^{\beta,\sigma}_\theta}{\delta \nu^{\epsilon}}(\nu^{\epsilon}, \theta)\mrd \theta \\
    &=-\frac{2\beta^2}{\sigma^2}\int_\Theta \bigg(f(\theta) \mdens^{\beta,\sigma}(\nu;\theta) \frac{\delta^\epsilon S}{\delta\nu^\epsilon}(\nu^\epsilon,\theta)G(\nu^\epsilon,\theta)\\
    &\qquad-f(\theta)\mdens^{\beta,\sigma}(\nu^\epsilon;\theta)\Big[\int_\Theta \mdens^{\beta,\sigma}(\nu;\theta')\frac{\delta^\epsilon S}{\delta\nu^\epsilon}(\nu^\epsilon,\theta')G(\nu^\epsilon,\theta')\mrd \theta'\Big]\bigg)\mrd \theta \\
    &=-\frac{2\beta^2}{\sigma^2}\mathrm{Cov}_{\theta'\sim \mfm^{\beta,\sigma}(\nu)}\Big[f(\theta),\,  \frac{\delta^\epsilon S}{\delta\nu^\epsilon}(\nu^\epsilon,\theta)G(\nu^\epsilon,\theta)\Big]\\
    &\quad +\frac{2\beta^2\epsilon}{\sigma^2}\int_{\Theta}f(\theta)\frac{\delta^{\epsilon}\mfm^{\beta,\sigma}_\theta}{\delta \nu^{\epsilon}}(\nu^{\epsilon}, \theta)\mrd \theta\Big[\int_\Theta \mdens^{\beta,\sigma}(\nu;\theta')\frac{\delta^\epsilon S}{\delta\nu^\epsilon}(\nu^\epsilon,\theta')G(\nu^\epsilon,\theta')\mrd \theta'\Big].
\end{split}\]
By rearranging terms, we have
\begin{equation}\label{fprodlemma}
\int_\Theta f(\theta) \frac{\delta^{\epsilon}\mfm^{\beta,\sigma}_\theta}{\delta \nu^{\epsilon}}(\nu^{\epsilon}, \theta)\mrd \theta=-\frac{2\beta^2}{\sigma^2}\mathrm{Cov}_{\theta\sim \mfm^{\beta,\sigma}(\nu)}\Big[f(\theta),\,  \frac{\delta^\epsilon S}{\delta\nu^\epsilon}(\nu^\epsilon,\theta)\frac{G(\nu^\epsilon,\theta)}{H(\nu^\epsilon, \epsilon)}\Big],
\end{equation}
with
\[
H(\nu^\epsilon,\nu)=1-\frac{2\beta^2\epsilon}{\sigma^2}\mathbb{E}_{\theta'\sim \mfm^{\beta,\sigma}(\nu)} \Big[\frac{\delta^\epsilon S}{\delta\nu^\epsilon}(\nu^\epsilon,\theta')G(\nu^\epsilon,\theta')\Big]\,.
\]
Observe that, from \eqref{Sboundcrude}, as $\epsilon\to 0$, we know $H(\nu^\epsilon, \nu) \to 1$ (and this convergence is uniform in $\theta$).

Given the characterization in \eqref{fprodlemma}, we return to the result of \eqref{Sderivlemma}, to derive a new representation of $\delta^\epsilon S/\delta \nu^\epsilon$:
\begin{equation}\label{Sderivlemma2}\begin{split}
    \frac{\delta^\epsilon S}{\delta\nu^\epsilon}(\nu^\epsilon,\theta)&= \int_{\mathcal{Z}}\frac{\delta\ell}{\delta m}(\mfm^{\beta,\sigma}(\nu^\epsilon),\theta,z)(\nu_1-\nu)(\mrd z)\\
   &\quad-\frac{2\beta^2}{\sigma^2}\int_{0}^1 \mathrm{Cov}_{\theta'\sim \mfm^{\beta,\sigma}(\nu)}\Big[\frac{\delta^2\mathrm{R}}{ \delta m^2}(\mfm^{\beta,\sigma}_{\lambda}(\nu^\epsilon),\nu,\theta,\theta')\frac{G(\nu^\epsilon,\theta')}{H(\nu^\epsilon, \nu)},\,  \frac{\delta^\epsilon S}{\delta\nu^\epsilon}(\nu^\epsilon,\theta')\Big]\mrd \lambda.
\end{split}
\end{equation}
Assumption \ref{ass:KLreg_assn}(\ref{ass:KLreg_D2loss_integrability}) guarantees that $\delta^2\mathrm{R}/ \delta m^2$ is sufficiently integrable for this result to hold (as we can approximate it with bounded functions $f$ in \eqref{fprodlemma}).

In order to show $S$ is differentiable, we will now construct a candidate for the directional derivative. For this purpose we temporarily fix $z$. Recalling that we expect $G(\nu^\epsilon, \theta)\to 1$ and $H(\nu^\epsilon, \nu)\to 1$ as $\epsilon\to0$,  consider the map $\mathcal{G}:L^2(\Theta;\mfm^{\beta, \sigma}(\nu))\to L^2(\Theta;\mfm^{\beta, \sigma}(\nu))$ defined by
\begin{equation}\label{vcontraction}\begin{split}
    \big(\mathcal{G}v\big)(\nu,\theta,z)&= \frac{\delta\ell}{\delta m}(\mfm^{\beta,\sigma}(\nu),\theta,z) - \int_{\mathcal{Z}}\frac{\delta\ell}{\delta m}(\mfm^{\beta,\sigma}(\nu),\theta,z')\nu(\mrd z')\\
   &\quad-\frac{2\beta^2}{\sigma^2} \mathbb{E}_{\theta'\sim \mfm^{\beta,\sigma}(\nu)}\Big[\frac{\delta^2\mathrm{R}}{ \delta m^2}(\mfm^{\beta,\sigma}(\nu),\nu,\theta,\theta')  v(\nu,\theta',z)\Big]\\
   \end{split}
\end{equation}
or equivalently, at the level of operators, for $z$ fixed,
\[\mathcal{G}v = \mathcal{L} - \frac{2\beta^2}{\sigma^2} \mathcal{C}_{\mfm^{\beta,\sigma}(\nu)}v\]
with $\mathcal{L} = \frac{\delta\ell}{\delta m}(\mfm^{\beta,\sigma}(\nu),\theta,z) - \int_{\mathcal{Z}}\frac{\delta\ell}{\delta m}(\mfm^{\beta,\sigma}(\nu),\theta,z')\nu(\mrd z')$ and $\mathcal{C}_{\mfm^{\beta,\sigma}(\nu)}$ as in Lemma \ref{lem:Cmonotone}.
We wish to define $v$ to be the fixed point of $\mathcal{G}v$, that is
\begin{equation}\label{eq:SderivFP}
v = \mathcal{G}v = \mathcal{L} - \frac{2\beta^2}{\sigma^2} \mathcal{C}_{\mfm^{\beta,\sigma}(\nu)}v
\end{equation}
which rearranges to give the formal expression 
\begin{equation}\label{eq:SderivFPSolved}
    v = \Big(\mathrm{id}+ \frac{2\beta^2}{\sigma^2} \mathcal{C}_{\mfm^{\beta,\sigma}(\nu)}\Big)^{-1} \mathcal{L}.
\end{equation}
We know from Lemma \ref{lem:Cmonotone} that $\mathcal{C}_{\mfm^{\beta,\sigma}(\nu)}$ has a nonnegative spectrum $\{\lambda^i\}_{i\ge 0}$, and write $\{e_i\}_{i\ge 0}$ for a corresponding orthonormal set of eigenvectors. Observe that the spectrum of $\mathrm{id}+ \frac{2\beta^2}{\sigma^2} \mathcal{C}_{\mfm^{\beta,\sigma}(\nu)}$ is given by $\{1+\frac{2\beta^2 }{\sigma^2}\lambda_i\}_{i\ge 0}$, with the same set of eigenvectors. By decomposing $\mathcal{L} = \sum_ic_ie_i$, as $\lambda_i\geq 0$ for all $i$ we compute 
\[\begin{split}
    v &= \sum_i\Big(\mathrm{id}+ \frac{2\beta^2}{\sigma^2} \mathcal{C}_{\mfm^{\beta,\sigma}(\nu)}\Big)^{-1} c_i e_i=\sum_i\frac{c_i}{1+\frac{2\beta^2 }{\sigma^2}\lambda_i} e_i.
\end{split}
\]
In particular, for every fixed value of $z$,
\begin{equation}\label{eqvbound}
    \|v\|^2_{L^2(\mfm^{\beta,\sigma}(\nu),\Theta)} = \sum_i\frac{c_i^2}{(1+\frac{2\beta^2 }{\sigma^2}\lambda_i)^2}\leq \sum_ic_i^2 = \|\mathcal{L}\|^2_{L^2(\mfm^{\beta,\sigma}(\nu),\Theta)}.
\end{equation}
Solving this system for all $z$, we obtain a function $v(\nu,\theta,z)$, which we can integrate with respect to $\nu(\mrd z)$ to see (from integrating \eqref{vcontraction}) that $\int_\mathcal{Z}v(\nu,\theta, z)\nu(dz)=0$.

If we integrate with respect to $(\nu_1 -\nu)(\mrd z)$, we get our candidate directional derivative:
\[\begin{split}
    &\int_{\mathcal{Z}} v(\nu,\theta,z)(\nu_1 -\nu)(\mrd z)\\
    &= \int_{\mathcal{Z}}\frac{\delta\ell}{\delta m}(\mfm^{\beta,\sigma}(\nu),\theta,z)(\nu_1-\nu)(\mrd z)\\
   &\quad-\frac{2\beta^2}{\sigma^2} \mathrm{Cov}_{\theta'\sim \mfm^{\beta,\sigma}(\nu)}\Big[\frac{\delta^2\mathrm{R}}{ \delta m^2}(\mfm^{\beta,\sigma}(\nu),\nu,\theta,\theta'),\,  \int_{\mathcal{Z}} v(\nu,\theta',z)(\nu_1 -\nu)(\mrd z)\Big],\end{split}
\]
(recalling that $\mathbb{E}_{\theta'\sim \mfm^{\beta,\sigma}(\nu)}\big[\frac{\delta^2\mathrm{R}}{ \delta m^2}(\mfm^{\beta,\sigma}(\nu),\nu,\theta,\theta')\big]=0$, so the covariance and the expectation of the product agree).
We next show that this candidate is indeed the directional derivative of $S$. 

We take \eqref{Sderivlemma2}, and write 
\[\begin{split}
   \frac{\delta^\epsilon S}{\delta\nu^\epsilon}
   &=\int_{\mathcal{Z}}\mathcal{L}^\epsilon(z)(\nu_1-\nu)(\mrd z) -\frac{2\beta^2}{\sigma^2}\mathcal{C}^\epsilon \Big[\frac{1}{H(\nu^\epsilon, \nu)}\frac{\delta^\epsilon S}{\delta\nu^\epsilon}(\nu^\epsilon,\cdot)\Big]\\
   &=\int_{\mathcal{Z}}\mathcal{L}^\epsilon(z)(\nu_1-\nu)(\mrd z) -\frac{2\beta^2}{\sigma^2}\mathcal{C}^\epsilon \Big[\Big(\frac{1}{H(\nu^\epsilon, \nu)}-1\Big)\frac{\delta^\epsilon S}{\delta\nu^\epsilon}(\nu^\epsilon,\cdot)\Big]-\frac{2\beta^2}{\sigma^2}\mathcal{C}^\epsilon \Big[\frac{\delta^\epsilon S}{\delta\nu^\epsilon}(\nu^\epsilon,\cdot)\Big]\\
\end{split}
\]
where 
\[\begin{split}
    \mathcal{L}^\epsilon(z) &:= \frac{\delta\ell}{\delta m}(\mfm^{\beta,\sigma}(\nu^\epsilon),\theta,z) - \int_\mathcal{Z}\frac{\delta\ell}{\delta m}(\mfm^{\beta,\sigma}(\nu^\epsilon),\theta,z')\nu(\mrd z'),\\
    \mathcal{C}^\epsilon v &:= \mathrm{Cov}_{\theta'\sim \mfm^{\beta,\sigma}(\nu)}\Big[\int_{0}^1\frac{\delta^2\mathrm{R}}{ \delta m^2}(\mfm^{\beta,\sigma}_{\lambda}(\nu^\epsilon),\nu,\theta,\theta')G(\nu^\epsilon,\theta')\mrd \lambda,\, v(\theta')\Big].
\end{split}\]
We formally rearrange to get 
\begin{equation}\label{eqSapproxcontract}
\begin{split}
   \frac{\delta^\epsilon S}{\delta\nu^\epsilon}
   &=\bigg(\mathrm{id}+\frac{2\beta^2}{\sigma^2}\mathcal{C}^\epsilon\bigg)^{-1}\bigg(\int_{\mathcal{Z}}\mathcal{L}^\epsilon(z)(\nu_1-\nu)(\mrd z) -\frac{2\beta^2}{\sigma^2}\Big(\frac{1}{H(\nu^\epsilon, \nu)}-1\Big)\mathcal{C}^\epsilon \Big[\frac{\delta^\epsilon S}{\delta\nu^\epsilon}(\nu^\epsilon,\cdot)\Big]\bigg).
\end{split}
\end{equation}

Taking $\epsilon \to 0$, we know that $\mfm^{\beta,\sigma}(\nu^\epsilon) \to \mfm^{\beta,\sigma}(\nu)$ and $\mfm^{\beta,\sigma}_\lambda(\nu^\epsilon) \to \mfm^{\beta,\sigma}(\nu)$ for all $\lambda\in [0,1]$.
Therefore, $\mathcal{L}^\epsilon(z) \to \mathcal{L}(z)$ and $\frac{\delta^2\mathrm{R}}{ \delta m^2}(\mfm^{\beta,\sigma}_{\lambda}(\nu^\epsilon),\nu,\theta,\theta') \to \frac{\delta^2\mathrm{R}}{ \delta m^2}(\mfm^{\beta,\sigma}(\nu),\nu,\theta,\theta')$. We have already seen $H(\nu^\epsilon, \nu)\to 1$ and  $G(\nu^\epsilon,\theta) \to 1$, and therefore, $\mathcal{C}^\epsilon \to \mathcal{C}_{\mfm^{\beta,\sigma}(\nu)}$ in the sense of Hilbert--Schmidt operators. In particular, as $\mathcal{C}_{\mfm^{\beta,\sigma}(\nu)}$ has a positive spectrum, this implies $(\mathrm{id} + \frac{2\beta^2}{\sigma^2}\mathcal{C}^\epsilon)^{-1} \to (\mathrm{id} + \frac{2\beta^2}{\sigma^2}\mathcal{C}_{\mfm^{\beta,\sigma}(\nu)})^{-1}$.

Therefore, we determine that, as $\epsilon\to 0$,
\[\frac{\delta^\epsilon S}{\delta\nu^\epsilon}(\nu^\epsilon,\cdot) \to \bigg(\mathrm{id}+\frac{2\beta^2}{\sigma^2}\mathcal{C}_{\mfm^{\beta,\sigma}(\nu)}\bigg)^{-1}\bigg(\int_{\mathcal{Z}}\mathcal{L}(z)(\nu_1-\nu)(\mrd z)\bigg) = \int_{\mathcal{Z}}v(z)(\nu_1-\nu)(\mrd z),\]
where $v$ is the fixed point constructed in \eqref{eq:SderivFP}.

From this, we conclude that  $\int_{\mathcal{Z}} v(\nu,\theta,z)(\nu_1 -\nu)(\mrd z)$ is the directional derivative of $S$ in the direction $(\nu_1-\nu)$. As the directional derivative is given by integrating $v$, it follows that 
\[\frac{\delta S}{\delta \nu}(\nu, \theta, z) = v(\nu, \theta, z) ,\]
and hence $S$ is differentiable in $\nu$, and has the desired representation. The bound in \eqref{eqvbound} completes the argument.
\end{proof}

\begin{lemma}[Derivative of Gibbs Measure]
\label{lem_parameter_measure_derivative}
Under Assumption \ref{ass:KLreg_assn}, the density of the Gibbs measure, that is, the map $\nu\mapsto \mfm_{\theta}^{\beta,\sigma}(\nu)$, has derivative
\[\frac{\delta\mfm^{\beta,\sigma}_\theta}{\delta \nu}(\nu, \theta, z)= -\frac{2\beta^2}{\sigma^2}\mfm^{\beta,\sigma}_\theta(\nu;\theta)\Big( \frac{\delta S}{\delta\nu}(\nu,\theta,z) - \Big[\int_\Theta \mfm^{\beta,\sigma}_\theta(\nu;\theta') \frac{\delta S}{\delta\nu}(\nu,\theta',z)\mrd\theta \Big]\Big).\]
In particular, for any $f\in L^2(\mrd \theta)$,
\[\int_\Theta f(\theta)\frac{\delta\mfm^{\beta,\sigma}_\theta}{\delta \nu}(\nu, \theta, z)\mrd\theta= -\frac{2\beta^2}{\sigma^2}\mathrm{Cov}_{\theta\sim \mfm^{\beta,\sigma}(\nu)}\Big[f(\theta),\,  \frac{\delta S}{\delta\nu}(\nu,\theta,z) \Big].\]
As a consequence, Assumption \ref{ass:KLreg_assn} implies Assumption \ref{main_ass_square_bounded}.

\end{lemma}

\begin{proof}
From Lemma \ref{lem:Sdifferentiable}, we know  $S$ is differentiable.
Recall that $\mfm^{\beta,\sigma}(\nu)$ satisfies
\[\mfm^{\beta,\sigma}_\theta(\nu;\theta) = \frac{Q(\nu, \theta)}{\int_\Theta Q(\nu, \theta')\mrd\theta'},\]
where $Q(\nu,\theta) =  \exp\Big\{-\frac{2\beta^2}{\sigma^2}\Big(S(\nu,\theta) + \frac{1}{2\beta^2}U(\theta)\Big)\Big\}$. Differentiating  with respect to $\nu$,
\[\frac{\delta Q}{\delta \nu}(\nu, \theta, z) = -\frac{2\beta^2}{\sigma^2}Q(\nu, \theta) \frac{\delta S}{\delta\nu}(\nu,\theta,z),\]
and hence
\[\begin{split}
&\frac{\delta\mfm^{\beta,\sigma}_\theta}{\delta \nu}(\nu, \theta, z) \\
&= \frac{\frac{\delta Q}{\delta \nu}(\nu,\theta,z) \int_\Theta Q(\nu,\theta')\mrd\theta' - Q(\nu,\theta) \frac{\delta }{\delta \nu} \int_\Theta Q(\nu,\theta')\mrd\theta' }{\big(\int_\Theta Q(\nu,\theta')\mrd\theta' \big)^2}\\
&= -\frac{2\beta^2}{\sigma^2}\frac{Q(\nu,\theta) \frac{\delta S}{\delta\nu}(\nu,\theta,z) \int_\Theta Q(\nu,\theta')\mrd\theta' - Q(\nu,\theta) (\int_\Theta Q(\nu, \theta') \frac{\delta S}{\delta\nu}(\nu,\theta',z)\mrd\theta' )}{\big(\int_\Theta Q(\nu,\theta')\mrd\theta' \big)^2}\\
&= -\frac{2\beta^2}{\sigma^2} \mfm^{\beta,\sigma}_\theta(\nu;\theta) \Big( \frac{\delta S}{\delta\nu}(\nu,\theta,z) - \Big[\int_\Theta \mfm^{\beta,\sigma}_\theta(\nu;\theta') \frac{\delta S}{\delta\nu}(\nu,\theta',z)\mrd\theta' \Big]\Big),
    \end{split}
\]
yielding the desired representation. The integrability of $\delta S/\delta \nu$ in Lemma \ref{lem:Sdifferentiable} also shows that $\frac{\delta\mfm^{\beta,\sigma}_\theta}{\delta \nu}(\nu, \theta, z) $ is square integrable with respect to $\mrd \theta$. The covariance representation follows immediately, as 
\[\int_\Theta \mfm^{\beta,\sigma}_\theta(\nu;\theta') \frac{\delta S}{\delta\nu}(\nu,\theta',z)\mrd\theta' = \mathbb{E}_{\theta'\sim \mfm^{\beta,\sigma}(\nu)}\Big[ \frac{\delta S}{\delta\nu}(\nu,\theta',z)\Big]. \]

Finally, we observe that Assumption \ref{ass:KLreg_assn}(\ref{ass:KLreg_assn_item1}) implies Assumption \ref{main_ass_square_bounded}(\ref{main_ass_square_bounded_item1}) and Assumption \ref{main_ass_square_bounded}(\ref{main_ass_square_bounded_item4}) is the result we have just shown. 
\end{proof}

\begin{proposition}\label{Prop_reccurence_estimate} 
Suppose Assumption \ref{ass:KLreg_assn} holds. Then
\[\begin{split}&\int_{\Theta}\Big[\frac{\delta S}{\delta\nu}(\nu,\theta, z)-\frac{\delta S}{\delta\nu}(\nu,\theta, z')\Big]^2\mfm^{\beta,\sigma}(\nu; \mrd \theta) \leq 2C_\theta^2\Big(1+2\mathbb{E}_{\theta\sim \mfm^{\beta,\sigma}(\nu)}\big[\|\theta\|^p\big]\Big)^2\Big(2+\|z\|^2+\|z'\|^2\Big)^2.
\end{split}\]
\end{proposition}
\begin{proof}
We write, using the notation of \eqref{eq:SderivFPSolved} and linearity,
\begin{align*}
&\frac{\delta S}{\delta\nu}(\nu,\theta, z)-\frac{\delta S}{\delta\nu}(\nu,\theta, z')\\
&=\Big(\mathrm{id}+\frac{2\beta^2}{\sigma^2}\mathcal{C}_{\mfm^{\beta,\sigma}(\nu)}\Big)^{-1}\Big[\frac{\delta \ell}{\delta m}(\mfm^{\beta,\sigma}(\nu),\cdot, z) - \frac{\delta \ell}{\delta m}(\mfm^{\beta,\sigma}(\nu),\cdot, z')\Big].
\end{align*}
We know from \eqref{eqvbound} that this is bounded in $L^2(\mfm^{\beta,\sigma}(\nu))$, in particular
\[\begin{split}
        &\int_{\Theta}\Big[\frac{\delta S}{\delta\nu}(\nu,\theta, z)-\frac{\delta S}{\delta\nu}(\nu,\theta, z')\Big]^2\mfm^{\beta,\sigma}(\nu; \mrd \theta) \\
    &\leq \int_{\Theta}\Big[\frac{\delta \ell}{\delta m}(\mfm^{\beta,\sigma}(\nu),\theta, z) - \frac{\delta \ell}{\delta m}(\mfm^{\beta,\sigma}(\nu),\theta, z')\Big]^2\mfm^{\beta,\sigma}(\nu; \mrd \theta)\\
    &\leq 2\int_{\Theta}\Big[\frac{\delta \ell}{\delta m}(\mfm^{\beta,\sigma}(\nu),\theta, z)\Big]^2 + \Big[\frac{\delta \ell}{\delta m}(\mfm^{\beta,\sigma}(\nu),\theta, z')\Big]^2\mfm^{\beta,\sigma}(\nu; \mrd \theta).
\end{split}\]

Assumption \ref{ass:KLreg_assn}(\ref{ass:KLreg_assn_item3}) yields
\[\begin{split}&\int_{\Theta}\Big[\frac{\delta S}{\delta\nu}(\nu,\theta, z)-\frac{\delta S}{\delta\nu}(\nu,\theta, z')\Big]^2\mfm^{\beta,\sigma}(\nu; \mrd \theta) \\
    &\leq 2\Big(2+\|z\|^2+\|z'\|^2\Big)^2\int_{\Theta}\big[g_1(\mfm^{\beta,\sigma}(\nu), \theta)\big]^2\mfm^{\beta,\sigma}(\nu; \mrd \theta).
\end{split}\]

The final result follows from Assumption \ref{ass:KLreg_assn}(\ref{ass:KLreg_D2loss_growth_bound}).

\end{proof}

\begin{remark}
    In the following arguments, we will regularly make use of the basic inequality (based on Cauchy--Schwarz), for any square integrable random variables $X,Y$,
    \[|\mathrm{Cov}(X,Y)| \leq \sqrt{\mathbb{V}(X)\mathbb{V}(Y)}\leq \sqrt{\mathbb{E}(X^2)\mathbb{E}(Y^2)}.\]
    A slightly more involved, but fundamentally similar, inequality is given in the following lemma.
\end{remark}
\begin{lemma}\label{lem:polybound}
    Let $\pi:(\mathbb{R}^+)^{2n}\to \mathbb{R}$ be a polynomial with positive coefficients. Then 
    \[\mathbb{E}_{\mathbf{Z}_n, \ols{\mathbf{Z}}_n}\Big[\pi\Big(\|Z_1\|,\|Z_2\|,..., \|Z_n\|,\|\ols Z_1\|, \|\ols Z_2\|,..., \|\ols Z_n\|\Big )\Big]\leq \mathbb{E}_{Z_1}\Big[\pi\Big(\|Z_1\|,\|Z_1\|,...,\|Z_1\|\Big )\Big].\]
    The same result also holds when the polynomial involves expectation of $\mathbb{E}_{Z_i}[\|Z_i\|^\alpha]$ terms.
\end{lemma}
\begin{proof}
First note that, by Jensen's inequality, for any $1\leq\beta\leq  \alpha$ and any nonnegative random variable $X$,
\begin{equation}\label{eq:polynomialjensen}
\mathbb{E}[X^\alpha]\mathbb{E}[X^\beta]=\mathbb{E}[X^\alpha]\mathbb{E}[X^{\alpha(\beta/\alpha)}]\leq  \mathbb{E}[X^\alpha]\mathbb{E}[X^\alpha]^{\beta/\alpha}=\mathbb{E}[X^\alpha]^{1+\beta/\alpha}  \leq \mathbb{E}[X^{\alpha+\beta}] .\end{equation}
By symmetry, this holds also for $1\leq \alpha\leq \beta$.    As $\pi$ is a polynomial, we can expand it fully as
    \[\pi\Big(\|Z_1\|,\|Z_2\|,..., \|Z_j\|,\|\ols Z_1\|, \|\ols Z_2\|,..., \|\ols Z_j\|\Big ) = \sum_{k=1}^{\kappa}c_k\prod_{j=1}^n \|Z_j\|^{\alpha_{j,k}}\|\ols Z_j\|^{\beta_{j,k}}\]
    where $\kappa\in \mathbb{N}$ and $c_k>0$ for each $k$.
    As every $Z_i$, $\ols Z_i$ is iid, for each term in this expansion
\[\begin{split}
    \mathbb{E}_{\mathbf{Z}_n, \ols{\mathbf{Z}}_n}\Big[\prod_{j=1}^n \|Z_j\|^{\alpha_{j,k}}\|\ols Z_j\|^{\beta_{j,k}} \Big] & =    \prod_{j=1}^n \mathbb{E}_{Z_j}[\|Z_j\|^{\alpha_{j,k}}]\mathbb{E}_{\ols Z_j}[\|\ols Z_j\|^{\beta_{j,k}}] =  \prod_{j=1}^n \mathbb{E}_{Z_1}[\|Z_1\|^{\alpha_{j,k}}]\mathbb{E}_{Z_1}[ \|Z_1\|^{\beta_{j,k}}]\\
    & \leq \mathbb{E}_{Z_1}\Big[\|Z_1\|^{\sum_{j=1}^n (\alpha_{j,k}+\beta_{j,k})}\Big],
\end{split} \]
where we have used \eqref{eq:polynomialjensen} to simplify the expansion. Substituting into the expansion of $\pi$, we obtain the desired inequality. If the polynomial involves expectation terms, these can be resolved using Jensen's inequality.
\end{proof}

\begin{proof}[Proof of Theorem \ref{thm: WGE Gibbs KL}]

Recall from \eqref{mperturb1} that $\nu_{(1)}(\lambda)=\nu_{n,(1)} +\lambda(\nu_{n}-\nu_{n,(1)})$. From Theorem~\ref{thm: another rep WGE},
\begin{equation}
    \label{WGEKLbasic}
    \mathrm{gen}(\mfm(\nu_n),\nupop)=\frac{1}{n}\mathbb{E}_{\mathbf{Z}_n,\ols{Z}_1}\big[ h(\mathbf{Z}_n,\ols{Z}_1)  \big],\end{equation}
where
\begin{equation}
\label{eq:hrepeat}
    \begin{split}
        &h(\mathbf{Z}_n,\ols{Z}_1)   \\
        &=\int_{0}^1\int_{0}^1\int_{\Theta} \Big( \frac{\delta \ell}{\delta m}\big(\mfm_{(1)}(\lambda),Z_1,\theta\big)\Big)\Big(\frac{\delta \mfm}{\delta \nu}\big(\nu_{(1)}(\tilde{\lambda}),Z_1\big)-\frac{\delta \mfm}{\delta \nu}\big(\nu_{(1)}(\tilde{\lambda}),\ols{Z}_1\big)\Big)(\mrd \theta)\, \mrd \tilde{\lambda}\,\mrd \lambda.
    \end{split}
\end{equation}
As $\mathbb{E}_{\theta\sim \mfm^{\beta, \sigma}(\nu_{(1)}(\tilde\lambda))}\big[\frac{\delta S}{\delta \nu}\big(\nu_{(1)}(\tilde\lambda), z, \theta\big)\big]\equiv 0$ (by definition of $S= \delta \mathrm{R}/\delta m$ and the normalization condition and chain rule), from Lemma~\ref{lem_parameter_measure_derivative} we know 
\begin{equation}\label{WGEKLexp1}
    \begin{split}
    &\int_{\Theta} \Big( \frac{\delta \ell}{\delta m}\big(\mfm_{(1)}(\lambda),Z_1,\theta\big)\Big)\Big(\frac{\delta \mfm}{\delta \nu}\big(\nu_{(1)}(\tilde{\lambda}),Z_1\big)-\frac{\delta \mfm}{\delta \nu}\big(\nu_{(1)}(\tilde{\lambda}),\ols{Z}_1\big)\Big)(\mrd \theta)\\
 & = -\frac{2\beta^2}{\sigma^2}\mathrm{Cov}_{\theta\sim \mfm^{\beta, \sigma}(\nu_{(1)}(\tilde\lambda))}\Big[ \frac{\delta \ell}{\delta m}\big(\mfm_{(1)}(\lambda),Z_1,\theta\big),\, \frac{\delta S}{\delta \nu}\big(\nu_{(1)}(\tilde\lambda), Z_1, \theta\big)-\frac{\delta S}{\delta \nu}\big(\nu_{(1)}(\tilde\lambda), \tilde Z_1, \theta\big)\Big]\\
 &\leq \frac{2\beta^2}{\sigma^2}\mathbb{E}_{\theta\sim \mfm^{\beta, \sigma}(\nu_{(1)}(\tilde\lambda))}\Big[\Big(\frac{\delta \ell}{\delta m}\big(\mfm_{(1)}(\lambda),Z_1,\theta\big)\Big)^2\Big]^{1/2}\\
 &\qquad \times \mathbb{E}_{\theta\sim \mfm^{\beta, \sigma}(\nu_{(1)}(\tilde\lambda))}\Big[\Big(\frac{\delta S}{\delta \nu}\big(\nu_{(1)}(\tilde\lambda), Z_1, \theta\big)-\frac{\delta S}{\delta \nu}\big(\nu_{(1)}(\tilde\lambda), \tilde Z_1, \theta\big)\Big)^2\Big]^{1/2}.
\end{split}
\end{equation}
Using Proposition \ref{Prop_reccurence_estimate} and \eqref{eq: bounded g ass},
\begin{equation}
    \begin{split}
    &\Big[\mathbb{E}_{\theta\sim \mfm^{\beta, \sigma}(\nu_{(1)}(\tilde\lambda))}\Big(\Big|\frac{\delta S}{\delta \nu}\big(\nu_{(1)}(\tilde\lambda), Z_1, \theta\big)-\frac{\delta S}{\delta \nu}\big(\nu_{(1)}(\tilde\lambda), \tilde Z_1, \theta\big)\Big|^2\Big)\Big]^{1/2}\\
    &\leq  \sqrt{2}\Big(\int_{\Theta}\big[g_1(\mfm^{\beta, \sigma}(\nu_{(1)}(\tilde\lambda)), \theta)\big]^2\mfm^{\beta, \sigma}(\nu_{(1)}(\tilde\lambda); \mrd \theta)\Big)^{1/2}\Big(2+\|Z_1\|^2+\|\ols Z_1\|^2\Big)\\
    &\leq \sqrt{2} C_{\theta}\Big(1+ 2\mathbb{E}_{\theta \sim \mfm^{\beta, \sigma}(\nu_{(1)}(\tilde\lambda))}\big[\|\theta\|^p\big]\Big)\Big(2+\|Z_1\|^2+\|\ols Z_1\|^2\Big).
\end{split}\end{equation}

From Lemma \ref{lem:optimizerinBnu}, we know 
\[\mathbb{E}_{\theta \sim \mfm^{\beta, \sigma}(\nu_{(1)}(\tilde\lambda))}\big[\|\theta\|^p\big]\leq \mathrm{R}(\tilde\gamma_p^\sigma, \nu_{(1)}(\tilde\lambda)) + \int_{\Theta}\|\theta\|^p\tilde\gamma_p^\sigma(\mrd \theta).\]
By construction,
\[\nu_{(1)}(\tilde\lambda) = \nu_n +\frac{1-\tilde \lambda}{n}\big(\delta_{\ols Z_1} - \delta_{Z_1}\big),\]
so 
\begin{equation}\label{eq:Rboundperturb1}
    \begin{split}
    \mathrm{R}(\tilde\gamma_p^\sigma, \nu_{(1)}(\tilde\lambda)) &= \mathrm{R}(\tilde\gamma_p^\sigma, \nu_{n}) + \frac{1-\tilde \lambda}{n}\Big(\ell(\tilde\gamma^\sigma_p, \ols Z_1) - \ell(\tilde\gamma^\sigma_p, Z_1)\Big)\\
    &\leq \mathrm{R}(\tilde\gamma_p^\sigma, \nu_{n}) + g(\tilde\gamma^\sigma_p)\frac{1-\tilde \lambda}{n}\Big(2+\|\ols Z_1\|^2+ \|Z_1\|^2)\Big),
\end{split}
\end{equation}
and hence 
\begin{equation}
    \label{WGEKLexp2}
    \begin{split}
    &\mathbb{E}_{\theta\sim \mfm^{\beta, \sigma}(\nu_{(1)}(\tilde\lambda))}\Big[\Big|\frac{\delta S}{\delta \nu}\big(\nu_{(1)}(\tilde\lambda), Z_1, \theta\big)-\frac{\delta S}{\delta \nu}\big(\nu_{(1)}(\tilde\lambda), \tilde Z_1, \theta\big)\Big|^2\Big]^{1/2}\\
    &\leq \sqrt{2} C_{\theta}\Big(1+ 2\mathrm{R}(\tilde\gamma_p^\sigma, \nu_{n}) +2\int_{\Theta}\|\theta\|^p\tilde\gamma_p^\sigma(\mrd \theta)+2g(\tilde\gamma^\sigma_p)\frac{1-\tilde \lambda}{n}\Big(2+ \|Z_1\|^2+\|\ols Z_1\|^2\Big)\Big)\\&\quad \times\Big(2+\|Z_1\|^2+\|\ols Z_1\|^2\Big).
\end{split}\end{equation}

We also know 
\begin{equation}
    \label{WGEKLexp3}
    \begin{split}
    &\mathbb{E}_{\theta\sim \mfm^{\beta, \sigma}(\nu_{(1)}(\tilde\lambda))}\Big[\Big(\frac{\delta \ell}{\delta m}\big(\mfm_{(1)}(\lambda),Z_1,\theta\big)\Big)^2\Big]^{1/2}\\
    &\leq \mathbb{E}_{\theta\sim \mfm^{\beta, \sigma}(\nu_{(1)}(\tilde\lambda))}\Big[g_1(\mfm_{(1)}(\lambda),\theta)^2\Big]^{1/2} \big(1+\|Z_1\|^2\big)\\
    &\leq C_\theta \Big(1+ \mathbb{E}_{\theta\sim \mfm^{\beta, \sigma}(\nu_{(1)}(\tilde\lambda))}\big[\|\theta\|^p\big]+\mathbb{E}_{\theta\sim \mfm_{(1)}(\lambda)}\big[\|\theta\|^p\big]\Big)\big(1+\|Z_1\|^2\big).
    \end{split}
\end{equation}
By definition, $\mfm_{(1)}(\lambda) = \mfm(\nu_{n,(1)}) + \lambda(\mfm(\nu_n) - \mfm(\nu_{n,(1)}))$, and hence
\[\mathbb{E}_{\theta\sim \mfm_{(1)}(\lambda)}\big[\|\theta\|^p\big] = (1-\lambda)\mathbb{E}_{\theta\sim \mfm(\nu_{n,(1)})}\big[\|\theta\|^p\big]+ \lambda\mathbb{E}_{\theta\sim \mfm(\nu_{n})}\big[\|\theta\|^p\big].\]
In a similar approach to \eqref{eq:Rboundperturb1}, we have,
\begin{equation}
    \begin{split}
    \mathrm{R}(\tilde\gamma_p^\sigma, \nu_{n,(1)}) &= \mathrm{R}(\tilde\gamma_p^\sigma, \nu_{n}) + \frac{1}{n}\Big(\ell(\tilde\gamma^\sigma_p, \ols Z_1) - \ell(\tilde\gamma^\sigma_p, Z_1)\Big)\\
    &\leq \mathrm{R}(\tilde\gamma_p^\sigma, \nu_{n}) + g(\tilde\gamma^\sigma_p)\frac{1}{n}\Big(2+\|\ols Z_1\|^2+ \|Z_1\|^2)\Big).
\end{split}
\end{equation} 
As before, it follows from Lemma \ref{lem:optimizerinBnu} that
\begin{equation}
\label{momentbound2}\begin{split}
&\mathbb{E}_{\theta\sim \mfm^{\beta, \sigma}(\nu_{(1)}(\tilde\lambda))}\big[\|\theta\|^p\big]+\mathbb{E}_{\theta\sim \mfm_{(1)}(\lambda)}\big[\|\theta\|^p\big]\\
& = \mathbb{E}_{\theta\sim \mfm^{\beta, \sigma}(\nu_{(1)}(\tilde\lambda))}\big[\|\theta\|^p\big]+(1-\lambda)\mathbb{E}_{\theta\sim \mfm(\nu_{n,(1)})}\big[\|\theta\|^p\big]+ \lambda\mathbb{E}_{\theta\sim \mfm(\nu_{n})}\big[\|\theta\|^p\big]\\
&\leq \Big[\mathrm{R}(\tilde\gamma^\sigma_p, \nu_{(1)}(\tilde\lambda)) + \int_{\Theta}\|\theta\|^p\tilde\gamma_p^\sigma(\mrd \theta) \Big] + (1-\lambda)\Big[\mathrm{R}(\tilde\gamma^\sigma_p, \nu_{n,(1)}) + \int_{\Theta}\|\theta\|^p\tilde\gamma_p^\sigma(\mrd \theta) \Big] \\
& \qquad + \lambda\Big[\mathrm{R}(\tilde\gamma^\sigma_p, \nu_{n}) + \int_{\Theta}\|\theta\|^p\tilde\gamma_p^\sigma(\mrd \theta) \Big]\\
    &\leq 2\mathrm{R}(\tilde\gamma_p^\sigma, \nu_{n}) + 2\int_{\Theta}\|\theta\|^p\tilde\gamma_p^\sigma(\mrd \theta) + g(\tilde\gamma^\sigma_p)\frac{(2-\tilde \lambda-\lambda)}{n}\Big(2+\|\ols Z_1\|^2+ \|Z_1\|^2\Big).
\end{split} 
\end{equation}

Therefore, substituting \eqref{WGEKLexp1}, \eqref{WGEKLexp2}, \eqref{WGEKLexp3} and \eqref{momentbound2} into \eqref{eq:hrepeat} and simplifying, we have
 \begin{equation}\label{WGEKLexp4}
 \begin{split}
        &h(\mathbf{Z}_n,\ols{Z}_1)   \\
        &\leq \frac{2\beta^2}{\sigma^2} C_\theta  \Big[1+2\mathrm{R}(\tilde\gamma_p^\sigma, \nu_{n}) + 2 \int_{\Theta}\|\theta\|^p\tilde\gamma_p^\sigma(\mrd \theta) + 2\frac{g(\tilde\gamma^\sigma_p)}{n}\Big(2+\|\ols Z_1\|^2+ \|Z_1\|^2\Big)\Big]\big(1+\|Z_1\|^2\big)\\
        &\qquad \times \sqrt{2} C_{\theta}\Big[1+ 2\mathrm{R}(\tilde\gamma_p^\sigma, \nu_{n})+ 2\int_{\Theta}\|\theta\|^p\tilde\gamma_p^\sigma(\mrd \theta) + 2\frac{g(\tilde\gamma^\sigma_p)}{n}\Big(2+ \|Z_1\|^2+\|\ols Z_1\|^2\Big)\Big]\Big(2+\|Z_1\|^2+\|\ols Z_1\|^2\Big)\\
        &\leq \sqrt{2}\frac{2\beta^2}{\sigma^2} C_\theta^2  \Big[1+2\mathrm{R}(\tilde\gamma_p^\sigma, \nu_{n}) + 2 \int_{\Theta}\|\theta\|^p\tilde\gamma_p^\sigma(\mrd \theta) + 2\frac{g(\tilde\gamma^\sigma_p)}{n}\Big(2+\|Z_1\|^2+\|\ols Z_1\|^2\Big)\Big]^2\\
        &\qquad \times \big(1+\|Z_1\|^2\big)\Big(2+\|Z_1\|^2+\|\ols Z_1\|^2\Big).
    \end{split}
    \end{equation}
We have the bound
\begin{equation}\label{eq:Rboundsum}
\mathrm{R}(\tilde\gamma_p^\sigma, \nu_n) \leq \int_\mathcal{Z} g(\tilde\gamma_p^\sigma) (1+\|z\|^2)\nu_n(\mrd{z}) = g(\tilde\gamma_p^\sigma)\frac{1}{n}\sum_{j=1}^n(1+\|Z_j\|^2).
\end{equation}

Therefore, using the inequality $(a+b(c+d))\leq  (a+cb)(c+d)/c$ for $a,b,c,d\ge 0$,
\begin{equation}\label{hexpansionfinal1} \begin{split}
        &h(\mathbf{Z}_n,\ols{Z}_1)   \\
        &\leq \sqrt{2}\frac{2\beta^2}{\sigma^2} C_\theta^2  \Big[1+ 2 \int_{\Theta}\|\theta\|^p\tilde\gamma_p^\sigma(\mrd \theta) + 2\frac{g(\tilde\gamma^\sigma_p)}{n}\Big(2+\|Z_1\|^2+\|\ols Z_1\|^2+\frac{1}{n}\sum_{j=1}^n(1+\|Z_j\|^2)\Big)\Big]^2\\
        &\qquad \times \big(1+\|Z_1\|^2\big)\Big(2+\|Z_1\|^2+\|\ols Z_1\|^2\Big)\\
        &\leq \frac{\sqrt{2}}{2}\frac{\beta^2}{\sigma^2} C_\theta^2  \Big[1+ 2 \int_{\Theta}\|\theta\|^p\tilde\gamma_p^\sigma(\mrd \theta) + 4\frac{g(\tilde\gamma^\sigma_p)}{n}\Big]^2\\
        &\qquad \times \Big(2+\|Z_1\|^2+\|\ols Z_1\|^2+\frac{1}{n}\sum_{j=1}^n(1+\|Z_j\|^2)\Big)^2\big(1+\|Z_1\|^2\big)\Big(2+\|Z_1\|^2+\|\ols Z_1\|^2\Big).
        \end{split}
        \end{equation}

Applying Lemma \ref{lem:polybound} we have 
\begin{equation}\label{polyexpansion1}\begin{split}
&\mathbb{E}_{\mathbf{Z}_n, \ols{\mathbf{Z}}_n}\Big[\Big(2+\|Z_1\|^2+\|\ols Z_1\|^2+\frac{1}{n}\sum_{j=1}^n(1+\|Z_j\|^2)\Big)^2\big(1+\|Z_1\|^2\big)\Big(2+\|Z_1\|^2+\|\ols Z_1\|^2\Big)\Big]\\
&\leq \mathbb{E}_{Z_1}\Big[\Big(2+\|Z_1\|^2+\|Z_1\|^2+\frac{1}{n}\sum_{j=1}^n(1+\|Z_1\|^2)\Big)^2\big(1+\|Z_1\|^2\big)\Big(2+\|Z_1\|^2+\|Z_1\|^2\Big)\Big]\\
&= (2\times 3^2)\mathbb{E}_{Z_1}\Big[(1+\|Z_1\|^2)^4\Big]
\end{split}
\end{equation}
And therefore, substituting \eqref{polyexpansion1} and \eqref{hexpansionfinal1} in \eqref{WGEKLbasic},
\begin{equation}\label{WGEKLexp5}
\begin{split}
 |\mathrm{gen}(\mfm(\nu_n),\nupop)|
 &\leq 9\sqrt{2}\frac{\beta^2}{\sigma^2} C_\theta^2  \Big[1+ 2 \int_{\Theta}\|\theta\|^p\tilde\gamma_p^\sigma(\mrd \theta) + 4\frac{g(\tilde\gamma^\sigma_p)}{n}\Big]^2 \mathbb{E}_{Z_1}\Big[\Big(1+\|Z_1\|^2\Big)^4\Big].\\
\end{split}
\end{equation}
\end{proof}

\begin{proof}[Proof of Theorem \ref{thm: LGE upper KL}]
    We start from the result of Theorem~\ref{thm: another rep LGE}, where we have    \[
        \mathrm{gen}_{L_2}(\mfm,\nupop)\leq \frac{1}{n}\bigg(4K + 2K^{1/2}\mathbb{E}_{\mathbf{Z}_2,\mathbf{\ols{Z}}_2}\Big[\tilde{h}_2(\mathbf{Z}_2, \mathbf{\ols{Z}}_2)^2\Big]^{1/2} + \mathbb{E}_{\mathbf{Z}_2,\mathbf{\ols{Z}}_2}\Big[\tilde{h}_2(\mathbf{Z}_2, \mathbf{\ols{Z}}_2)^2\Big]\bigg),
    \]
    where 
    \begin{align*}
        \tilde{h}_2(\mathbf{Z}_n, \mathbf{\ols{Z}}_2)&= \int_{0}^1\int_0^1\int_\Theta \Big(\mathbb E_{\ols{Z}_3} \Big[  \frac{\delta\ell}{\delta m}(\mfm_{(1,2)}(\lambda_1),\ols{Z}_3,\theta) \Big]-\frac{\delta \ell}{\delta m }(\mfm_{(1,2)}(\lambda),Z_1,\theta)\Big)\\
        & \qquad \times \Delta_\mfm(\mathbf{Z}_2, \mathbf{\ols{Z}}_2; \mrd\theta)\, \mrd \lambda\mrd\tilde\lambda,\\
        \Delta_\mfm(\mathbf{Z}_n, \mathbf{\ols{Z}}_2; \mrd\theta)&= \Big(\frac{\delta \mfm}{\delta\nu}(\nu_{(1,2)}(\tilde\lambda), Z_1)(\mrd\theta)-\frac{\delta \mfm}{\delta\nu}(\nu_{(1,2)}(\tilde\lambda), \ols{Z}_1)(\mrd\theta)\\
         &\qquad +\frac{\delta \mfm}{\delta\nu}(\nu_{(1,2)}(\tilde\lambda), Z_2)(\mrd\theta)-\frac{\delta \mfm}{\delta\nu}(\nu_{(1,2)}(\tilde\lambda), \ols{Z}_2)(\mrd\theta)\Big)\\
         &= \int_{\mathcal{Z}}\frac{\delta \mfm}{\delta\nu}(\nu_{(1,2)}(\tilde\lambda), z')(\mrd\theta)\big(\delta_{Z_1}-\delta_{\tilde Z_1}+\delta_{Z_2}-\delta_{\tilde Z_2}\big)\big(\mrd z').
    \end{align*}
Applying Lemma~\ref{lem_parameter_measure_derivative}, we compute 
\[\begin{split}
&\Big|\int_\Theta \frac{\delta \ell}{\delta m }(\mfm_{(1,2)}(\lambda),z,\theta)\Delta_\mfm(\mathbf{Z}_2, \mathbf{\ols{Z}}_2; \mrd\theta)\Big|\\
&=\Big|-\frac{2\beta^2}{\sigma^2}\mathrm{Cov}_{\theta\sim \mfm(\nu_{(1,2)}(\tilde\lambda))}\Big[\frac{\delta \ell}{\delta m }(\mfm_{(1,2)}(\lambda),z,\theta),\\
&\qquad \, \int_{\mathcal{Z}}\frac{\delta S}{\delta\nu}(\nu_{(1,2)}(\tilde\lambda), \theta, z')(\delta_{Z_1}-\delta_{\ols{Z}_1}+\delta_{Z_2}-\delta_{\ols{Z}_2})(\mrd z')\Big]\Big|\\
 &\leq \frac{2\beta^2}{\sigma^2}\mathbb{E}_{\theta\sim \theta\sim \mfm(\nu_{(1,2)}(\tilde\lambda))}\Big[\Big(\frac{\delta \ell}{\delta m }(\mfm_{(1,2)}(\lambda),z,\theta)\Big)^2\Big]^{1/2}\\
 &\qquad\times \mathbb{E}_{\theta\sim \theta\sim \mfm(\nu_{(1,2)}(\tilde\lambda))}\Big[\Big(\int_{\mathcal{Z}}\frac{\delta S}{\delta\nu}(\nu_{(1,2)}(\tilde\lambda), \theta, z')\Big(\delta_{Z_1}-\delta_{\ols{Z}_1}+\delta_{Z_2}-\delta_{\ols{Z}_2}\Big)(\mrd z')\Big)^2\Big]^{1/2},
\end{split}\]
and similarly
\[\begin{split}
    &\Big|\int_\Theta \mathbb{E}_{\ols{Z}_3}\Big[\frac{\delta \ell}{\delta m }(\mfm_{(1,2)}(\lambda),z,\theta)\Big]\Delta_\mfm(\mathbf{Z}_2, \mathbf{\ols{Z}}_2; \mrd\theta)\Big|\\
     &\leq \frac{2\beta^2}{\sigma^2}\mathbb{E}_{\theta\sim \mfm(\nu_{(1,2)}(\tilde\lambda))}\Big[\Big(\mathbb{E}_{\ols{Z}_3}\Big(\frac{\delta \ell}{\delta m }(\mfm_{(1,2)}(\lambda),\ols{Z}_3,\theta)\Big)\Big)^2\Big]^{1/2}\\
     &\qquad\times \mathbb{E}_{\theta\sim \mfm(\nu_{(1,2)}(\tilde\lambda))}\Big[\Big(\int_{\mathcal{Z}}\frac{\delta S}{\delta\nu}(\nu_{(1,2)}(\tilde\lambda), \theta, z')\Big(\delta_{Z_1}-\delta_{\ols{Z}_1}+\delta_{Z_2}-\delta_{\ols{Z}_2}\Big)(\mrd z')\Big)^2\Big]^{1/2}.
    \end{split}\]
Therefore, using the inequality $(a+b)^2\leq 2(a^2+b^2)$ and Jensen's inequality,
\begin{equation}\label{LGEhBound1}
    \begin{split}
    &\tilde{h}_2(\mathbf{Z}_n, \mathbf{\ols{Z}}_2)^2\\
    &\leq \frac{4\beta^4}{\sigma^4} \int_0^1\int_0^1\mathbb{E}_{\theta\sim \mfm(\nu_{(1,2)}(\tilde\lambda))}\Big[\Big(\int_{\mathcal{Z}}\frac{\delta S}{\delta\nu}(\nu_{(1,2)}(\tilde\lambda), \theta, z')\Big(\delta_{Z_1}-\delta_{\ols{Z}_1}+\delta_{Z_2}-\delta_{\ols{Z}_2}\Big)(\mrd z')\Big)^2\Big]\\
    &\quad \times \Big(\mathbb{E}_{\theta\sim \mfm(\nu_{(1,2)}(\tilde\lambda))}\Big[\Big(\mathbb{E}_{\ols{Z}_3}\Big(\frac{\delta \ell}{\delta m }(\mfm_{(1,2)}(\lambda),\ols{Z}_3,\theta)\Big)\Big)^2\Big]^{1/2}\\
    &\qquad\qquad  + \mathbb{E}_{\theta\sim \mfm(\nu_{(1,2)}(\tilde\lambda))}\Big[\Big(\frac{\delta \ell}{\delta m }(\mfm_{(1,2)}(\lambda),Z_1,\theta)\Big)^2\Big]^{1/2}\Big)^2\bigg] \mrd \lambda\mrd\tilde\lambda.\\
    &\leq \frac{8\beta^4}{\sigma^4} \int_0^1\int_0^1\mathbb{E}_{\theta\sim \mfm(\nu_{(1,2)}(\tilde\lambda))}\Big[\Big(\int_{\mathcal{Z}}\frac{\delta S}{\delta\nu}(\nu_{(1,2)}(\tilde\lambda), \theta, z')\Big(\delta_{Z_1}-\delta_{\ols{Z}_1}+\delta_{Z_2}-\delta_{\ols{Z}_2}\Big)(\mrd z')\Big)^2\Big]\\
    &\quad \times \bigg[\mathbb{E}_{\theta\sim \mfm(\nu_{(1,2)}(\tilde\lambda))}\Big[\mathbb{E}_{\ols{Z}_3}\Big[\Big(\frac{\delta \ell}{\delta m }(\mfm_{(1,2)}(\lambda),\ols{Z}_3,\theta)\Big)^2\Big]\Big]\\
    &\qquad\qquad  + \mathbb{E}_{\theta\sim \mfm(\nu_{(1,2)}(\tilde\lambda))}\Big[\Big(\frac{\delta \ell}{\delta m }(\mfm_{(1,2)}(\lambda),Z_1,\theta)\Big)^2\Big]\bigg] \mrd \lambda\mrd\tilde\lambda.
\end{split}
\end{equation}

Similarly to \eqref{eq:Rboundperturb1}, we know that 
\[    \begin{split}
    \mathrm{R}(\tilde\gamma_p^\sigma, \nu_{(1,2)}(\tilde\lambda)) &\leq \mathrm{R}(\tilde\gamma_p^\sigma, \nu_{n}) + g(\tilde\gamma^\sigma_p)\frac{1-\tilde \lambda}{n}\Big(4+\|\ols Z_1\|^2+ \|Z_1\|^2+\|\ols Z_2\|^2+ \|Z_2\|^2\Big),
\end{split}\]
and hence, as in \eqref{momentbound2}, we have
\[\begin{split}
&\mathbb{E}_{\theta\sim \mfm^{\beta, \sigma}(\nu_{(1,2)}(\tilde\lambda))}\big[\|\theta\|^p\big]+\mathbb{E}_{\theta\sim \mfm_{(1,2)}(\lambda)}\big[\|\theta\|^p\big]\\
& = (2-\lambda)\mathbb{E}_{\theta\sim \mfm(\nu_{n,(1,2)})}\big[\|\theta\|^p\big]+ \lambda\mathbb{E}_{\theta\sim \mfm(\nu_{n})}\big[\|\theta\|^p\big]\\
    &\leq 2\mathrm{R}(\tilde\gamma_p^\sigma, \nu_{n}) + 2\int_{\Theta}\|\theta\|^p\tilde\gamma_p^\sigma(\mrd \theta) \\
    &\qquad + g(\tilde\gamma^\sigma_p)\frac{(2-\lambda-\tilde \lambda)}{n}\Big(4+\|\ols Z_1\|^2+ \|Z_1\|^2+\|\ols Z_2\|^2+ \|Z_2\|^2\Big).
\end{split} \]

Using Proposition \ref{Prop_reccurence_estimate}, as in \eqref{WGEKLexp2}, we know 
\[\begin{split}
    &\mathbb{E}_{\theta\sim \mfm(\nu_{(1,2)}(\tilde\lambda))}\Big[\Big(\int_{\mathcal{Z}}\frac{\delta S}{\delta\nu}(\nu_{(1,2)}(\tilde\lambda), \theta, z')\Big(\delta_{Z_1}-\delta_{\ols{Z}_1}+\delta_{Z_2}-\delta_{\ols{Z}_2}\Big)(\mrd z')\Big)^2\Big]\\
    &\leq2\mathbb{E}_{\theta\sim \mfm(\nu_{(1,2)}(\tilde\lambda))}\Big[\Big(\int_{\mathcal{Z}}\frac{\delta S}{\delta\nu}(\nu_{(1,2)}(\tilde\lambda), \theta, z')\Big(\delta_{Z_1}-\delta_{\ols{Z}_1}\Big)(\mrd z')\Big)^2\Big]\\
    &\qquad+2\mathbb{E}_{\theta\sim \mfm(\nu_{(1,2)}(\tilde\lambda))}\Big[\Big(\int_{\mathcal{Z}}\frac{\delta S}{\delta\nu}(\nu_{(1,2)}(\tilde\lambda), \theta, z')\Big(\delta_{Z_2}-\delta_{\ols{Z}_2}\Big)(\mrd z')\Big)^2\Big]\\
    &\leq 4\,C_\theta^2\Big(4+\|Z_1\|^2+\|\tilde{Z}_1\|^2+\|Z_2\|^2+\|\tilde{Z}_2\|^2\Big)^2 \\&\qquad\times\Big(1+2\mathrm{R}(\tilde\gamma_p^\sigma, \nu_{n}) + 2\int_{\Theta}\|\theta\|^p\tilde\gamma_p^\sigma(\mrd \theta)+2g(\tilde\gamma^\sigma_p)\frac{1-\tilde \lambda}{n}\Big(4+\|\ols Z_1\|^2+ \|Z_1\|^2+\|\ols Z_2\|^2+ \|Z_2\|^2\Big)\Big)^2.
\end{split}\]
Using Lemma \ref{lem:optimizerinBnu} and \eqref{eq: bounded g ass},
\[\begin{split}
    &\mathbb{E}_{\theta\sim \mfm(\nu_{(1,2)}(\tilde\lambda))}\Big[\mathbb{E}_{\ols{Z}_3}\Big[\Big(\frac{\delta \ell}{\delta m }(\mfm_{(1,2)}(\lambda),\ols{Z}_3,\theta)\Big)^2\Big]\Big]\\
    &\leq C_{\theta}^2\Big(1+\mathbb{E}_{\theta\sim \mfm^{\beta, \sigma}(\nu_{(1,2)}(\tilde\lambda))}\big[\|\theta\|^p\big]+\mathbb{E}_{\theta\sim \mfm_{(1,2)}(\lambda)}\big[\|\theta\|^p\big]\Big)^2\mathbb{E}_{\ols{Z}_3}[(1+\|\ols{Z}_3\|^2)^2]\\
    &\leq C_{\theta}^2\Big(1+2\mathrm{R}(\tilde\gamma_p^\sigma, \nu_{n}) + 2\int_{\Theta}\|\theta\|^p\tilde\gamma_p^\sigma(\mrd \theta) \\
    &\qquad + g(\tilde\gamma^\sigma_p)\frac{(2-\lambda-\tilde \lambda)}{n}\Big(4+\|\ols Z_1\|^2+ \|Z_1\|^2+\|\ols Z_2\|^2+ \|Z_2\|^2\Big)\Big)^2\mathbb{E}_{\ols{Z}_3}[(1+\|\ols{Z}_3\|^2)^2],
\end{split}
\]
and similarly
\[\begin{split} &\mathbb{E}_{\theta\sim \mfm(\nu_{(1,2)}(\tilde\lambda))}\Big[\Big(\frac{\delta \ell}{\delta m }(\mfm_{(1,2)}(\lambda),Z_1,\theta)\Big)^2\Big]\\
&\leq C_{\theta}^2\Big(1+2\mathrm{R}(\tilde\gamma_p^\sigma, \nu_{n}) + 2\int_{\Theta}\|\theta\|^p\tilde\gamma_p^\sigma(\mrd \theta) \\
    &\qquad + g(\tilde\gamma^\sigma_p)\frac{(2-\lambda-\tilde \lambda)}{n}\Big(4+\|\ols Z_1\|^2+ \|Z_1\|^2+\|\ols Z_2\|^2+ \|Z_2\|^2\Big)\Big)^2(1+\|\ols{Z}_1\|^2)^2.
\end{split}\]
As in \eqref{eq:Rboundsum}, we know $R(\tilde\gamma_p^\sigma, \nu_n) \leq g(\tilde\gamma_p^\sigma)\frac{1}{n}\sum_{j=1}^n(1+\|Z_j\|^2)$, and hence
\[\begin{split}
    &\mathrm{R}(\tilde\gamma_p^\sigma, \nu_{n}) + g(\tilde\gamma^\sigma_p)\frac{1}{n}\Big(4+\|\ols Z_1\|^2+ \|Z_1\|^2+\|\ols Z_2\|^2+ \|Z_2\|^2\Big)\\
    &\leq  g(\tilde\gamma_p^\sigma)\Big(4+\|\ols Z_1\|^2+ \|Z_1\|^2+\|\ols Z_2\|^2+ \|Z_2\|^2+\frac{1}{n}\sum_{j=1}^n(1+\|Z_j\|^2)\Big).
\end{split}\]

By substitution in \eqref{LGEhBound1} and simplification we get
\[\begin{split}
&\tilde{h}_2(\mathbf{Z}_n, \mathbf{\ols{Z}}_2)^2\\
&\leq \frac{8\beta^4}{\sigma^4}4\,C_\theta^2\Big(4+\|Z_1\|^2+\|\tilde{Z}_1\|^2+\|Z_2\|^2+\|\tilde{Z}_2\|^2\Big)^2 \\
&\quad\times\Big(1+2\mathrm{R}(\tilde\gamma_p^\sigma, \nu_{n}) + 2\int_{\Theta}\|\theta\|^p\tilde\gamma_p^\sigma(\mrd \theta)+2g(\tilde\gamma^\sigma_p)\frac{1}{n}\Big(4+\|\ols Z_1\|^2+ \|Z_1\|^2+\|\ols Z_2\|^2+ \|Z_2\|^2\Big)\Big)^2\\
&\quad \times \bigg[C_{\theta}^2\Big(1+2\mathrm{R}(\tilde\gamma_p^\sigma, \nu_{n}) + 2\int_{\Theta}\|\theta\|^p\tilde\gamma_p^\sigma(\mrd \theta) \\
&\qquad + g(\tilde\gamma^\sigma_p)\frac{2}{n}\Big(4+\|\ols Z_1\|^2+ \|Z_1\|^2+\|\ols Z_2\|^2+ \|Z_2\|^2\Big)\Big)^2\Big[(1+\|\ols{Z}_1\|^2)^2+\mathbb{E}_{\ols{Z}_3}[(1+\|\ols{Z}_3\|^2)^2]\Big]\bigg]\\
&\leq 2^3\Big(\frac{2\beta^2}{\sigma^2}\Big)^2\,C_\theta^4\Big(4+\|Z_1\|^2+\|\tilde{Z}_1\|^2+\|Z_2\|^2+\|\tilde{Z}_2\|^2\Big)^2 \\
&\quad\times\Big(1+2\mathrm{R}(\tilde\gamma_p^\sigma, \nu_{n}) + 2\int_{\Theta}\|\theta\|^p\tilde\gamma_p^\sigma(\mrd \theta)+\frac{2g(\tilde\gamma^\sigma_p)}{n}\Big(4+\|\ols Z_1\|^2+ \|Z_1\|^2+\|\ols Z_2\|^2+ \|Z_2\|^2\Big)\Big)^4\\
&\quad \times\Big[(1+\|\ols{Z}_1\|^2)^2+\mathbb{E}_{\ols{Z}_3}[(1+\|\ols{Z}_3\|^2)^2]\Big]\\
&\leq 2^3\Big(\frac{2\beta^2}{\sigma^2}\Big)^2\,C_\theta^4\Big(4+\|Z_1\|^2+\|\tilde{Z}_1\|^2+\|Z_2\|^2+\|\tilde{Z}_2\|^2\Big)^2 \\
&\quad\times\Big(1+ 2\int_{\Theta}\|\theta\|^p\tilde\gamma_p^\sigma(\mrd \theta)+2g(\tilde\gamma^\sigma_p)\Big(4+\|\ols Z_1\|^2+ \|Z_1\|^2+\|\ols Z_2\|^2+ \|Z_2\|^2+\frac{1}{n}\sum_{j=1}^n(1+\|Z_j\|^2)\Big)\Big)^4\\
&\quad \times\Big[(1+\|\ols{Z}_1\|^2)^2+\mathbb{E}_{\ols{Z}_3}[(1+\|\ols{Z}_3\|^2)^2]\Big]\\
&\leq 2^{-5}\Big(\frac{2\beta^2}{\sigma^2}\Big)^2\,C_\theta^4 \Big(4+\|Z_1\|^2+\|\tilde{Z}_1\|^2+\|Z_2\|^2+\|\tilde{Z}_2\|^2\Big)^2\\
&\quad \times  \Big(1+ 2\int_{\Theta}\|\theta\|^p\tilde\gamma_p^\sigma(\mrd \theta)+8g(\tilde\gamma^\sigma_p)\Big)^4\\
&\quad \times\Big(4+\|\ols Z_1\|^2+ \|Z_1\|^2+\|\ols Z_2\|^2+ \|Z_2\|^2+\frac{1}{n}\sum_{j=1}^n(1+\|Z_j\|^2)\Big)^4\Big[(1+\|\ols{Z}_1\|^2)^2+\mathbb{E}_{\ols{Z}_3}[(1+\|\ols{Z}_3\|^2)^2]\Big]\\
\end{split}\]
where we use the inequality $(a+b(c+d))\leq  (a+cb)(c+d)/c$ for $a,b,c,d\ge 0$ when passing to the penultimate line.

Applying Lemma \ref{lem:polybound} gives
\[\begin{split}
&\mathbb{E}_{\mathbf{Z}_n, \ols{\mathbf{Z}}_2}\bigg[\Big(4+\|\ols Z_1\|^2+ \|Z_1\|^2+\|\ols Z_2\|^2+ \|Z_2\|^2\Big)^2\\
&\qquad \times \Big(4+\|\ols Z_1\|^2+ \|Z_1\|^2+\|\ols Z_2\|^2+ \|Z_2\|^2+\frac{1}{n}\sum_{j=1}^n(1+\|Z_j\|^2)\Big)^4\\
&\qquad \times\Big((1+\|\ols{Z}_1\|^2)^2+\mathbb{E}_{\ols{Z}_3}[(1+\|\ols{Z}_3\|^2)^2]\Big)\bigg]\\
&\leq (2^6\times 5^4)\mathbb{E}_{Z_1}\big[(1+\|Z_1\|^2)^{8}\big].
\end{split}\]
Therefore,
\[\begin{split}
&\mathbb{E}_{\mathbf{Z}_n, \mathbf{\ols{Z}}_2}\Big[\tilde{h}_2(\mathbf{Z}_n, \mathbf{\ols{Z}}_2)^2\Big]\leq (2\times 5^4)\Big(\frac{2\beta^2}{\sigma^2}\Big)^2C_\theta^4\Big(1+2\int_{\Theta}\|\theta\|^p\tilde\gamma_p^\sigma(\mrd \theta)+ 8g(\tilde\gamma^\sigma_p)\Big)^4\mathbb{E}_{Z_1}[(1+\|Z_1\|^2)^{8}].
\end{split}\]
The result follows from Theorem~\ref{thm: another rep LGE}.
\end{proof}

\section{Proofs from Section~\ref{Sec: Application}}\label{App: proofs of Sec: one hidden example}

\begin{proof}[Proof of Lemma~\ref{lem: equ with main ass NN}]
 We recall that our loss is of the form $\ell(m, z) = \ell_o(\mathbb{E}_{\theta\sim m}[\phi(\theta,x)],y)$, for $\ell_o$ a convex function. It follows that $\ell$ is convex and nonnegative with respect to $m$.
 
 We write $\partial_{\hat y}\ell_o$ for the derivative of $\ell_o$ with respect to its first argument. By the chain rule, we can differentiate with respect to $m$, to obtain, with $\Phi(m,x) := \mathbb{E}_{\theta\sim m}[\phi(\theta,x)]$,
    \begin{equation}\label{Eq: derivative NN loss}
        \begin{split}
        \frac{\delta \Phi}{\delta m}(m,x,\theta) &= \phi(\theta,x)-\Phi(m,x),\\
            \frac{\delta \ell}{\delta m}(m,z,\theta)&=\partial_{\hat{y}}\ell_o(\Phi(m,x),y)\Big(\phi(\theta,x)-\Phi(m,x)\Big).
         \end{split}
    \end{equation}
    Note that, 
   \[
    \frac{\delta^2 \ell}{\delta m^2}(m,z,\theta, \theta') = \frac{\delta}{\delta m}\left(  \frac{\delta \ell}{\delta m}(\cdot,\cdot, \theta)\right)(m,z, \theta')\,,
    \] 
  and due to the normalization convention, see \citep[Remark 2.5]{cardaliaguet2019master},  
    \begin{equation}
        \begin{split}
             \frac{\delta^2 \ell}{\delta m^2}(m,z,\theta, \theta')&=\partial_{\hat{y}\hat{y}}\ell_o(\Phi(m,x),y)\Big(\phi(\theta,x)-\Phi(m,x)\Big)\Big(\phi(\theta',x)-\Phi(m,x)\Big).
        \end{split}
    \end{equation}
It follows that $\ell$ is $\mathcal{C}^2$ with respect to $m$ (the required bounds on the derivatives will be obtained below).

Without loss of generality, for notational convenience, we assume that $L_\ell, L_{\ell,1}, L_{\ell,2}$ and $L_\phi$ in Assumption \ref{ass:NN_KL} are all at least $1$. Throughout, $c$ is a constant which can vary from line to line, but does not depend on $n$.   
 
 We first consider obtaining bounds on $\ell$. We know that 
\[\begin{split}
    \ell(\mfm(\nu_n),z)&\leq L_\ell(1+ \|y\|^2 + \|\hat y\|^2)= L_\ell(1+ \|y\|^2 + \|\Phi(m,x)\|^2))\,.
\end{split}
\]
As
\[\|\Phi(m,x)\|^2=\|\mathbb{E}_{\theta\sim m}[\phi(\theta,x)]\|^2\leq L_\phi^2 (1+\|x\|)^2\mathbb{E}_{\theta\sim m}[1+\|\theta\|^2]^2,\]
we know 
\[\begin{split}
    \ell(\mfm(\nu_n),z)
    &\leq L_\ell\big(1+ \|y\|^2 + L_\phi^2 (1+\|x\|)^2\mathbb{E}_{\theta\sim m}[1+\|\theta\|^2]^2\big)\\
    &\leq 2L_\ell L_\phi^2 \mathbb{E}_{\theta\sim m}[1+\|\theta\|^2]^2 \big(1+\|y\|^2 + \|x\|^2).
\end{split}
\]
We therefore take $g(m):= 2L_\ell L_\phi^2 \mathbb{E}_{\theta\sim m}[1+\|\theta\|^2]^2$. As we know $p\ge 2$ and $\tilde\gamma_p^\sigma$ is a density in $\mathcal{P}_p$, it follows that $g(\tilde\gamma_p^\sigma)<\infty$.

We can also compute, using the above bounds,
\[
\begin{split}\max_{Z\in \{Z_1, \ols{Z}_1\}}\mathbb{E}\big[\big(\ell(\mfm(\nu_n),Z)\big)^2\big]&\leq 4L_\ell^2 L_\phi^4 \mathbb{E}_{\mathbf{Z}_n}\Big[\mathbb{E}_{\theta\sim \mfm(\nu_n)}[1+\|\theta\|^2]^4 (1+\|Z_1\|^2)^2\Big].
\end{split}\]
As $\mfm(\nu_n)\in B(\nu_n)$, we know 
\[\begin{split}
    \mathbb{E}_{\theta\sim m}[1+\|\theta\|^2]^2 
    &\leq 2+2\mathrm{R}(\tilde\gamma^\sigma_p, \nu_n)^2 +2\mathbb{E}_{\theta\sim \tilde\gamma^\sigma_p}[\|\theta\|^2]^2\\
    &\leq 2+2\frac{1}{n}\sum_{j=1}^n \ell(\tilde\gamma^\sigma_p, Z_j)^2 +2\mathbb{E}_{\theta\sim \tilde\gamma^\sigma_p}[\|\theta\|^2]^2\\
    &\leq 2+2\frac{1}{n}\sum_{j=1}^n  4L_\ell^2 L_\phi^4 \mathbb{E}_{\theta\sim \tilde\gamma^\sigma_p}[1+\|\theta\|^2]^4 \big(1+\|Z_j\|^2)^2 +2\mathbb{E}_{\theta\sim \tilde\gamma^\sigma_p}[\|\theta\|^2]^2.
\end{split}\]
As $\tilde\gamma^\sigma_p \in \mathcal{P}_p$ and $p\geq 2$, we have
$\mathbb{E}_{\theta\sim \tilde\gamma^\sigma_p}[1+\|\theta\|^2]^4\leq c$,
and hence, applying Lemma \ref{lem:polybound},
\[
\begin{split}\max_{Z\in \{Z_1, \ols{Z}_1\}}\mathbb{E}\big[\big(\ell(\mfm(\nu_n),Z)\big)^2\big]&\leq c\mathbb{E}_{\mathbf{Z}_n}\Big[\Big(1+ \frac{1}{n}\sum_{j=1}^n \big(1+\|Z_j\|^2)^2\Big)^2 (1+\|Z_1\|^2)^2\Big]\\
&\leq c \mathbb{E}_{Z_1}\Big[(1+ \|Z_1\|^2)^{6}\Big].
\end{split}\]

We now consider the derivative of $\ell$. Writing $m_2:=\mathbb{E}_{\theta\sim m}[\|\theta\|^2]$, we have 
    \begin{align*}
        \Big|\frac{\delta \ell}{\delta m}\big(m,z,\theta\big)\Big|& = \Big|\partial_{\hat{y}}\ell_o(\Phi(m,x),y)\Big|\,|\phi(\theta,x)-\Phi(m,x)|\\
        &\leq L_{\ell,1}(1+|\Phi(m,x)|+\|y\|)\big|\phi(\theta,x)-\Phi(m,x)\big|\\
        &\leq L_{\ell,1}L_{\phi}\Big(1+L_{\phi}(1+\|x\|)(1+m_2)+\|y\|\Big)(1+\|x\|)\Big(2+\|\theta\|^2+m_2\Big)\\
        &\leq L_{\ell,1}L_{\phi}(1+L_{\phi})(1+m_2)\Big(2+\|\theta\|^2+m_2\Big)\Big(1+\|x\|+\|y\|\Big)(1+\|x\|)\\
         &\leq 2 L_{\ell,1}L_{\phi}(1+L_{\phi})\Big(2+\|\theta\|^2+m_2\Big)^2\Big(1+\|x\|^2+\|y\|^2\Big)\\
         &\leq 6L_{\ell,1}L_{\phi}(1+L_{\phi})\Big(4+\|\theta\|^4+\mathbb{E}_{\theta\sim m}[\|\theta\|^4]\Big)\Big(1+\|x\|^2+\|y\|^2\Big).
    \end{align*}
  We then take \[g_1(m,\theta) = 6L_{\ell,1}L_{\phi}(1+L_{\phi})\Big(4+\|\theta\|^4+\mathbb{E}_{\theta\sim m}[\|\theta\|^4]\Big),\] and we have, with $c$ a constant varying from line to line,
  \begin{equation}
  \begin{split}
      \mathbb{E}_{\theta\sim m'}\big[g_1(m,\theta)^2\big]^{1/2}&=c\mathbb{E}_{\theta\sim m'}\Big[\Big(4+\|\theta\|^4+\mathbb{E}_{\theta\sim m}[\|\theta\|^4]\Big)^2\Big]^{1/2}\\
      &\leq c\Big(8+2\mathbb{E}_{\theta\sim m'}[\|\theta\|^8]+2\mathbb{E}_{\theta\sim m}[\|\theta\|^8]\Big)^{1/2}\\
      &\leq c\Big(1+\mathbb{E}_{\theta\sim m'}[\|\theta\|^8]+\mathbb{E}_{\theta\sim m}[\|\theta\|^8]\Big).
      \end{split}
  \end{equation}
Together with Jensen's inequality, this yields \eqref{eq: bounded g ass} in the case  $p \geq 8$.

  Finally, we consider the second derivative of $\ell$. We have,  
   \begin{equation}
   \begin{split}
        &\mathbb{E}_{\theta,\theta'\sim m}\mathbb{E}_{z\sim \nu}\Big[\Big(\frac{\delta^2 \mathrm{\ell}}{\delta m^2}(m, z, \theta, \theta') \Big)^2\Big]\\
&=\int_\mathcal{Z}\Big[\partial_{\hat{y}\hat{y}}\ell_o(\Phi(m,x),y)\Big]^2\mathbb{V}_{\theta\sim m}\Big(\phi(\theta,x)\Big)\mathbb{V}_{\theta'\sim m}\Big(\phi(\theta',x)\Big)\nu(\mrd z)\\
        &= \int_\mathcal{Z}\Big[\partial_{\hat{y}\hat{y}}\ell_o(\Phi(m,x),y)\Big]^2\Big[\mathbb{V}_{\theta\sim m}\Big(\phi(\theta,x)\Big)\Big]^2\nu(\mrd z)\\
        &\leq c\, \mathbb{E}_{Z\sim \nu}[(1+\|Z\|)^2] \mathbb{E}_{\theta\sim m}\big[(1+\|\theta\|^2)^2\big].\\
    \end{split}
    \end{equation}
giving the stated inequalities. As $\nu\in \mathcal{P}_2(\mathcal{Z})$, and assuming $m\in \mathcal{P}_p(\Theta)$ for $p\ge 8$, this implies \eqref{eq: bounded second moment ass}. It follows that Assumption \ref{ass:KLreg_assn} holds whenever $p\ge 8$.
\end{proof}

\section{Application: Expected parametric loss}\label{app: expected loss}

For comparison with the neural-network setting considered above, we now introduce a second application of our model, for the expected parametric loss. As before, we will derive an upper bound on expected generalization error. 

\paragraph{Expected loss:}  Suppose $\theta$ directly captures the relationship between $X$ and $Y$. We train using a parametric loss function, $(\theta,z)\mapsto\ell_p(\theta,z)$, inspired by \citep{xu2017information}.
As described in Section~\ref{sec: problem formulation}, we describe learning as a map $\nu_n\mapsto\mfm(\nu_n)\in \mathcal{P}(\Theta)$. We consider the single-observation expected parametric loss  
\begin{equation}
    \label{eq: expected parametric loss func}
    (m,z) \mapsto \mathbb{E}_{\theta\sim m}[\ell_p(\theta,z)].
\end{equation}
 For example, in a least-squares regression problem, take $\theta = (a,b)$ and $\ell_p(\theta,z) = \big(y-(a+bx)\big)^2$. Note that in this case, our generic loss function is defined as
 \begin{equation}
     \ell(m,z):=\mathbb{E}_{\theta\sim m}[\ell_p(\theta,z)].
 \end{equation}

\paragraph{WGE of expected loss:} We apply our approach based on functional derivatives to the expected parametric loss \eqref{eq: expected parametric loss func}. It is worth noting that the WGE, derived from the expected parametric loss function in \eqref{eq: expected parametric loss func}, is equivalent to the expected generalization error proposed by \citep{xu2017information,raginsky2016information}. 
Let restate Theorem~\ref{thm: another rep WGE} in terms of the parametric loss function.
\begin{theorem}[Theorem~\ref{thm: another rep WGE} in terms of $\ell_p(\theta,z)$]\label{thm: WGE for param loss}
Assume that $|\ell_p(\theta,z)|\leq C(1+|\theta|^2)$, and training map, $\nu_n\mapsto \mfm(\nu_n)$ is differentiable, in the sense of Definition \ref{Def: func derivative m2m}. Then,
\begin{align*}
       & \mathrm{gen}(\mfm(\nu_n),\nu)  =\frac{1}{n}\mathbb{E}_{\mathbf{Z_n},\ols{Z}_1}\bigg[\int_{0}^1 \int_{\Theta}\ell_p(\theta,\ols{Z}_1) \Big(\frac{\delta \mfm}{\delta \nu}(\nu_{(1)}(\tilde{\lambda}),Z_1)-\frac{\delta \mfm}{\delta \nu}(\nu_{(1)}(\tilde{\lambda}),\ols{Z}_1)\Big)\!(\mrd \theta)\, \mrd \tilde{\lambda}\bigg]\,,
       \end{align*} 
\end{theorem}

To demonstrate an easy application of Theorem~\ref{thm: WGE for param loss}, we discuss a simple example of a learning problem, where the representation of the expected generalization error in Theorem~\ref{thm: WGE for param loss} can be computed exactly, and regularization is not needed. This example has been previously considered in \citep{bu2020tightening,aminian2021exact}; we obtain the same results, based on our more abstract approach.

\begin{example}[Gaussian mean estimation]\label{Example:mean gaussian}
    We consider the problem of estimating the mean $\mu$ of Gaussian random variable $Z\sim \mathcal{N}(\mu,\tilde{\sigma}^2)$ with $n$ i.i.d. samples $\{Z_i\}_{i=1}^n$. We consider the squared parametric loss function $\ell_p(\theta,Z)=(\theta-Z)^2$ and the empirical risk minimization problem is solved by $\theta=\frac{1}{n}\sum_{i=1}^n Z_i$.  Consequently, with the empirical risk minimizer, we have $\theta\sim \mathcal{N}(\mu,\frac{\tilde{\sigma}^2}{n})$. Therefore, we suppose the learning map is given by the distribution $\mfm(\nu_n)=\mathcal{N}(\mathbb{E}_{Z\sim \nu_n}[Z],\frac{\tilde{\sigma}^2}{n})$. Then, we have $\mathrm{gen}(\mfm(\nu_n),\nupop) =\frac{2\tilde{\sigma}^2}{n}$.
\end{example}
\begin{proof}[Proof of Example \ref{Example:mean gaussian}]
    We compute $\frac{\delta \mfm}{\delta \nu}(\nu_n,z;\theta)$ using the Gaussian density, to give
    \begin{align*}
        \frac{\delta \mfm}{\delta \nu}(\nu_n,z;\theta)=\mfm(\nu_n)\frac{n}{\tilde{\sigma}^2}(\theta-\mathbb{E}_{Z\sim\nu_n}[Z])(z-\mathbb{E}_{Z\sim \nu_n}[Z]).
    \end{align*}
    Recall that $\nu_{(1)}(\tilde{\lambda})=\nu_{n,(1)} + \tilde{\lambda}(\nu_{n}-\nu_{n,(1)})$. Note that 
     \begin{equation}
        \begin{split}
        \mathbb{E}_{Z\sim\nu_{(1)}(\tilde{\lambda})}[Z]&=\frac{\tilde \lambda}{n}\ols{Z}_1+\frac{(1-\tilde \lambda)}{n}Z_1+\frac{1}{n}\sum_{j=2}^n Z_j,\\
        \mathbb{E}_{\theta\sim\mfm(\nu_{(1)}(\tilde{\lambda}))}[\theta]&=\mathbb{E}_{Z\sim\nu_{(1)}(\tilde{\lambda})}[Z],\\
        \mathbb{E}_{\theta\sim\mfm(\nu_{(1)}(\tilde{\lambda}))}[\theta^2]&=\mathbb{E}_{Z\sim\nu_{(1)}(\tilde{\lambda})}[Z]^2+\frac{\tilde{\sigma}^2}{n},\\
        \mathbb{E}_{\theta\sim\mfm(\nu_{(1)}(\tilde{\lambda}))}[\theta^3]&=\mathbb{E}_{Z\sim\nu_{(1)}(\tilde{\lambda})}[Z]^3+3\mathbb{E}_{Z\sim\nu_{(1)}(\tilde{\lambda})}[Z]\frac{\tilde{\sigma}^2}{n}
        .
    \end{split}
       \end{equation}
    From Theorem~\ref{thm: another rep WGE} we have
       \begin{align*}
        &\mathrm{gen}(\mfm(\nu_n),\nu)\\
        &=\frac{1}{n}\mathbb{E}_{\mathbf{Z_n}\ols{Z}_1}\Big[\int_{0}^1\int_{\Theta} (\theta-\ols{Z}_1)^2 \Big(\frac{\delta \mfm}{\delta \nu}(\nu_{(1)}(\tilde{\lambda}),Z_1;\theta)-\frac{\delta \mfm}{\delta \nu}(\nu_{(1)}(\tilde{\lambda}),\ols{Z}_1;\theta)\Big)\!(\mrd \theta) \mrd \tilde{\lambda} \Big]\\
        &= \frac{1}{\tilde{\sigma}^2}\mathbb{E}_{\mathbf{Z_n}\ols{Z}_1}\Big[\int_{0}^1\int_{\Theta} (\theta-\ols{Z}_1)^2 (\theta-\mathbb{E}_{Z\sim \nu_{(1)}(\tilde{\lambda})}[Z])(Z_1-\ols{Z}_1)\mfm(\nu_{(1)}(\tilde{\lambda}))\!(\mrd \theta) \mrd \tilde{\lambda} \Big]\\
        &=\frac{1}{\tilde{\sigma}^2}\Bigg(\mathbb{E}_{\mathbf{Z_n}\ols{Z}_1}\Big[(Z_1-\ols{Z}_1)\int_{0}^1\mathbb{E}_{\theta\sim \mfm(\nu_{(1)}(\tilde{\lambda}))}\Big[\theta(\theta-\ols{Z}_1)^2\Big]\mrd \tilde{\lambda}  \Big]
        \\&\quad-\mathbb{E}_{\mathbf{Z_n}\ols{Z}_1}\Big[(Z_1-\ols{Z}_1)\int_{0}^1\mathbb{E}_{Z\sim \nu_{(1)}(\tilde{\lambda})}[Z]\mathbb{E}_{\theta\sim \mfm(\nu_{(1)}(\tilde{\lambda}))}\Big[(\theta-\ols{Z}_1)^2\Big]\mrd \tilde{\lambda}  \Big]\Bigg)\\
        &= \frac{1}{\tilde{\sigma}^2}\Bigg(\mathbb{E}_{\mathbf{Z_n}\ols{Z}_1}\Big[(Z_1-\ols{Z}_1)\int_{0}^1\mathbb{E}_{\theta\sim \mfm(\nu_{(1)}(\tilde{\lambda}))}\Big[\theta^3-2\theta^2\ols{Z}_1+\ols{Z}_1^2\theta)\Big]\mrd \tilde{\lambda}  \Big]
        \\&\quad-\mathbb{E}_{\mathbf{Z_n}\ols{Z}_1}\Big[(Z_1-\ols{Z}_1)\int_{0}^1\mathbb{E}_{Z\sim \nu_{(1)}(\tilde{\lambda})}[Z]\mathbb{E}_{\theta\sim \mfm(\nu_{(1)}(\tilde{\lambda}))}\Big[\theta^2-2\theta\ols{Z}_1+\ols{Z}_1^2\Big]\mrd \tilde{\lambda}  \Big]\Bigg)
        \\
        &= \frac{1}{\tilde{\sigma}^2}\Bigg(\mathbb{E}_{\mathbf{Z_n}\ols{Z}_1}\Big[(Z_1-\ols{Z}_1)\int_{0}^1\Big(\mathbb{E}_{Z\sim\nu_{(1)}(\tilde{\lambda})}[Z]^3+3\mathbb{E}_{Z\sim\nu_{(1)}(\tilde{\lambda})}[Z]\frac{\tilde{\sigma}^2}{n}\\&\qquad\qquad-2(\mathbb{E}_{Z\sim\nu_{(1)}(\tilde{\lambda})}[Z]^2+\frac{\tilde{\sigma}^2}{n})\ols{Z}_1+\ols{Z}_1^2\mathbb{E}_{Z\sim\nu_{(1)}(\tilde{\lambda})}[Z])\Big)\mrd \tilde{\lambda}  \Big]
        \\&\quad-\mathbb{E}_{\mathbf{Z_n}\ols{Z}_1}\Big[(Z_1-\ols{Z}_1)\int_{0}^1\Big(\mathbb{E}_{Z\sim\nu_{(1)}(\tilde{\lambda})}[Z]^3+\mathbb{E}_{Z\sim\nu_{(1)}(\tilde{\lambda})}[Z]\frac{\tilde{\sigma}^2}{n}\\&\qquad\qquad-2\mathbb{E}_{Z\sim\nu_{(1)}(\tilde{\lambda})}[Z]^2\ols{Z}_1+\ols{Z}_1^2\mathbb{E}_{Z\sim\nu_{(1)}(\tilde{\lambda})}[Z]\Big)\mrd \tilde{\lambda}  \Big]\Bigg)
        \\
        &=\frac{2}{n}\mathbb{E}_{\mathbf{Z_n}\ols{Z}_1}\Big[(Z_1-\ols{Z}_1)\int_{0}^1 (\mathbb{E}_{Z\sim\nu_{(1)}(\tilde{\lambda})}[Z]-\ols{Z}_1)\mrd \tilde{\lambda}  \Big]
        \\
        &=\frac{2}{n}\mathbb{E}_{\mathbf{Z_n}\ols{Z}_1}\Big[(Z_1-\ols{Z}_1)\int_{0}^1 \Big(\frac{\tilde \lambda}{n}\ols{Z}_1+\frac{(1-\tilde \lambda)}{n}Z_1+\frac{1}{n}\sum_{j=2}^n Z_j-\ols{Z}_1\Big)\mrd \tilde{\lambda}  \Big] \\
        &=\frac{2}{n}\mathbb{E}_{\mathbf{Z_n}\ols{Z}_1}\Big[\Big(Z_1-\ols{Z}_1\Big) \Big(\frac{1}{2n}(Z_1-\ols{Z}_1)+\frac{1}{n}\sum_{j=2}^n Z_j\Big)  \Big]
        \\&=
        \frac{2\tilde{\sigma}^2}{n}.
   \end{align*}
\end{proof}

\paragraph{Parametric Gibbs measure:} For a slightly less trivial example, we will now apply our approach based on KL-regularized risk minimization to a parametric loss setting. In \citep{raginsky2016information,aminian2021exact} and \citet{kuzborskij2019distribution}, the KL-regularized expected empirical risk minimization problem \eqref{Eq: regularized risk}
 is studied with an expected parametric loss, leading to the following parametric Gibbs measure (a.k.a. Gibbs algorithm) as solution:
\begin{align}\label{Eq: Gibbs distribution}
    \mdens^{p,\beta,\sigma}(\nu_n;\theta) =\frac{1}{F_{\beta,\sigma}}\exp\Bigg\{-\frac{2\beta^2}{\sigma^2}\Big[\frac{1}{n}\sum_{i=1}^n \ell_p(\theta,Z_i)+ \frac{1}{2\beta^2} U(\theta)\Big] \Bigg\}.
\end{align}
We make the following assumptions on the parametric loss to investigate generalization performance.

\begin{assumption}\label{ass:expected_loss_KL}
For a constant $p\geq 2$, the regularizing potential $U$ satisfies $\lim_{\|\theta\|\to \infty}\frac{U(\theta)}{\|\theta\|^p} = \infty,$ the loss $\ell_p$ is nonnegative, and  there exists a constant $M_p$ such that, for all $m\in\mathcal{P}_2(\Theta)$,
\[ \begin{split}  
|\ell_p(\theta,z)|&\leq M_p(1+\|\theta\|^2)(1+\|z\|^2).
\end{split}\]
\end{assumption}

\begin{lemma}\label{lem: equ with ass 5}
Under Assumption~\ref{ass:expected_loss_KL}, for all $z\in\mathcal{Z}$,  $\theta\in \Theta$, $m\in \mathcal{P}_p(\Theta)$, and $\nu\in \mathcal{P}_2(\mathcal{Z})$, we have
\begin{align}\label{eq: loss G}
|\ell(m,z)|&\leq M_p(1+\mathbb{E}_{\theta\sim m}[\|\theta\|^2])(1+\|z\|^2)\\\label{eq: diff loss G}
\Big| \frac{\delta \ell}{\delta m}(m,z,\theta)\Big|
       &\leq M_p(2+\mathbb{E}_{\theta\sim m}[\|\theta\|^2]+\|\theta\|^2)(1+\|z\|^2)\\\label{eq: K param G}
\max_{Z\in \{Z_1, \ols{Z}_1\}}\mathbb{E}\big[\big(\ell(\mfm(\nu_n),Z)\big)^2\big]&\leq 4 M_p^2 \Big(2+\mathbb{E}_{\theta\sim \tilde\gamma_p^\sigma}[\|\theta\|^2]\Big)^2 \mathbb{E}_{\mathbf{Z}_n,\ols{Z}_1}\Big[(1+\|Z_1\|^2)^4 \Big],\\\label{eq: second diff loss G}
            \mathbb{E}_{\theta,\theta'\sim m}\mathbb{E}_{z\sim \nu}\Big[\Big(\frac{\delta^2 \mathrm{\ell}}{\delta m^2}(m, z, \theta, \theta') \Big)^2\Big]&=0.
\end{align}
In particular, if $\mathbb{E}_{Z\sim \nupop}[\|Z\|^{8}]<\infty$ then $\max_{Z\in \{Z_1, \ols{Z}_1\}}\mathbb{E}\big[\big(\ell(\mfm(\nu_n),Z)\big)^2\big]<\infty.$
\end{lemma}
\begin{proof}
    We recall that our loss is of the form $\ell(m, z) = \mathbb{E}_{\theta\sim m}\big[\ell_p(\theta,z)\big]$. We can differentiate with respect to $m$, to obtain
    \begin{equation}\label{Eq: derivative param loss}
        \begin{split}
            \frac{\delta \ell}{\delta m}(m,z,\theta)&=\ell_p(\theta,z)-\mathbb{E}_{\theta\sim m}\big[\ell_p(\theta,z)\big],\\
            \frac{\delta^2 \ell}{\delta m^2}(m,z,\theta, \theta')&=0.
         \end{split}
    \end{equation}
    To obtain \eqref{eq: loss G}, we calculate
    \[\begin{split}
        |\ell(m,z)| &= |\mathbb{E}_{\theta\sim m}[\ell_p(\theta,z)]|\leq M_p(1+\mathbb{E}_{\theta\sim m}[\|\theta\|^2])(1+\|z\|^2).\end{split}\]
Therefore, we have
\[g(m)= M_p(1+\mathbb{E}_{\theta\sim m}[\|\theta\|^2]).\]
We can compute the functional derivative of expected loss, \eqref{eq: diff loss G}, for all $m\in \mathcal{P}_2(\Theta)$,
        \begin{equation}
        \begin{split}
       \Big| \frac{\delta \ell}{\delta m}(m,z,\theta)\Big|&=\Big|\ell_p(\theta,z)-\mathbb{E}_{\theta'\sim m}\big[\ell_p(\theta',z)\big]\Big|\\
       &\leq M_p(2+\mathbb{E}_{\theta'\sim m}[\|\theta'\|^2]+\|\theta\|^2)(1+\|z\|^2).
       \end{split}
       \end{equation}
       We then take
       \begin{align*}
           g_1(m,\theta)=M_p(2+\mathbb{E}_{\theta'\sim m}[\|\theta'\|^2]+\|\theta\|^2),
       \end{align*}
       and we have
         \begin{equation}
        \begin{split}
          \mathbb{E}_{\theta\sim m'}[ g_1(m,\theta)^2]^{1/2}&=M_p \mathbb{E}_{\theta\sim m'}[(2+\mathbb{E}_{\theta'\sim m}[\|\theta'\|^2]+\|\theta\|^2)^2]^{1/2}\\
          &\leq 2\sqrt{2}M_p \Big(1+\mathbb{E}_{\theta\sim m}[\|\theta\|^4]+\mathbb{E}_{\theta\sim m'}[\|\theta\|^4]\Big).
       \end{split}
       \end{equation}
       For \eqref{eq: K param G}, we have,
       \begin{equation}\label{Eq: eq k param}
       \begin{split}
        \mathbb{E}\big[\big(\ell(\mfm(\nu_n),Z)\big)^2\big]&= \mathbb{E}_{\mathbf{Z}_n,Z}\Big[\Big(\mathbb{E}_{\theta\sim \mfm(\nu_n)}[\ell_p(\theta,Z)]\Big)^2\Big]\\
        &\leq M_p^2 \mathbb{E}_{\mathbf{Z}_n,Z}\Big[\Big(\mathbb{E}_{\theta\sim \mfm(\nu_n)}[(1+\|\theta\|^2)](1+\|Z\|^2)\Big)^2\Big].
        \end{split}
    \end{equation}
    From Lemma \ref{lem:optimizerinBnu}, we know 
\begin{equation}\begin{split}\label{eq: K param 2}
    \mathbb{E}_{\theta \sim \mfm(\nu_{n})}\big[\|\theta\|^2\big]&\leq \mathrm{R}(\tilde\gamma_p^\sigma, \nu_{n}) + \int_{\Theta}\|\theta\|^2\tilde\gamma_p^\sigma(\mrd \theta)\\
    &\leq (1+\mathbb{E}_{\theta\sim \tilde\gamma_p^\sigma}[\|\theta\|^2])\frac{1}{n}\sum_{j=1}^n (1+\|Z_j\|^2)+\mathbb{E}_{\theta\sim \tilde\gamma_p^\sigma}[\|\theta\|^2]\\
    &\leq \Big(1+\mathbb{E}_{\theta\sim \tilde\gamma_p^\sigma}[\|\theta\|^2]\Big(1+\frac{1}{n}\sum_{j=1}^n (1+\|Z_j\|^2)\Big)\Big) .\end{split}\end{equation}
    Substituting \eqref{eq: K param 2} in \eqref{Eq: eq k param}, we have,
    \begin{equation}
       \begin{split}
    \mathbb{E}\big[\big(\ell(\mfm(\nu_n),Z)\big)^2\big]&\leq M_p^2 \mathbb{E}_{\mathbf{Z}_n,Z}\Big[\Big(2+\mathbb{E}_{\theta\sim \tilde\gamma_p^\sigma}[\|\theta\|^2]\Big(1+\frac{1}{n}\sum_{j=1}^n (1+\|Z_j\|^2)\Big)\Big)(1+\|Z\|^2)\Big)^2\Big]\\
        &\leq M_p^2 \Big(2+\mathbb{E}_{\theta\sim \tilde\gamma_p^\sigma}[\|\theta\|^2]\Big)^2\mathbb{E}_{\mathbf{Z}_n,Z}\Big[\Big(\Big(1+\frac{1}{n}\sum_{j=1}^n (1+\|Z_j\|^2)\Big)(1+\|Z\|^2)\Big)^2 \Big]\\
        &\leq 4 M_p^2 \Big(2+\mathbb{E}_{\theta\sim \tilde\gamma_p^\sigma}[\|\theta\|^2]\Big)^2 \mathbb{E}_{Z\sim\nupop}\Big[(1+\|Z\|^2)^4 \Big].
        \end{split}
    \end{equation}
\end{proof}

In the following, we derive upper bounds on WGE and LGE of the parametric Gibbs measure by combining Lemma~\ref{lem: equ with ass 5} with Theorems~\ref{thm: WGE Gibbs KL} and \ref{thm: LGE upper KL}, respectively.
 \begin{proposition}[WGE and the parametric Gibbs measure]\label{prop: WGE class Gibbs}
      Assume that Assumption~\ref{ass:expected_loss_KL} and $\mathbb{E}_{Z\sim\nupop}\big[\|Z\|^{8}\big]<\infty$ hold. Then, the following upper bound holds on the WGE of the parametric Gibbs measure

      \[  |\mathrm{gen}(\mfm(\nu_n),\nupop)|\leq \frac{c}{n}\frac{2\beta^2}{\sigma^2}  \mathbb{E}_{Z_1}\Big[\big(1+\|Z_1\|^2\big)^4\Big]\]
where $c<\infty$ is independent of $n$.
 \end{proposition}

By leveraging the property of the parametric Gibbs measure being differentially private for bounded parametric loss functions, as stated in \citep[Theorem~1]{xu2017information}, a sub-optimal bound, i.e., $|\mathrm{gen}(\mfm^{\beta,\sigma}(\nu_n),\nupop) |\leq \sqrt{\frac{\beta^2}{\sigma^2 n}}$, can be obtained. However, a tighter upper bound, i.e., $|\mathrm{gen}(\mfm^{\beta,\sigma}(\nu_n),\nupop) |\leq \frac{\beta^2}{2\sigma^2 n}$, can be derived by further examinination with a  bounded loss assumption using Hoeffding's lemma, as demonstrated in \citep{raginsky2016information}. This tighter bound exhibits a similar rate of decay, namely $\mathcal{O}\big(1/n\big)$. Another upper bound, i.e., $\mathrm{gen}(\mfm^{\beta,\sigma}(\nu_n),\nupop) \leq \frac{4 \gamma^2 \beta^2}{\sigma^2 n}$, is established in \citep[Theorem 1]{kuzborskij2019distribution}, albeit under a different assumption that $\ell_p(\theta,z)$ is $\gamma$-sub-Gaussian under the parametric Gibbs measure, i.e., $\mfm^{\beta,\sigma}(\nu_n)$. Similarly, \citep[Theorem 2]{aminian2021exact} presents an upper bound, i.e., $\mathrm{gen}(\mfm^{\beta,\sigma}(\nu_n),\nupop) \leq \frac{2\gamma^2 \beta^2}{\sigma^2 n}$, assuming that the parametric loss function is $\gamma$-sub-Gaussian under the distribution $\nupop$ for all $\theta\in\Theta$.

In contrast to the aforementioned studies, we adopt a weaker assumption in this work, namely that $\nupop$ possesses a finite 8th moment. This condition holds for sub-Gaussian distributions, as well as numerous others such as the Student's $t$ distribution with more than $8$ degrees of freedom or lognormal distributions. Making the stronger assumption that $\nupop$ possesses a finite $16$th moment (which is still weaker than assuming it is sub-Gaussian), we can apply Theorem \ref{thm: LGE upper KL} to give a bound on the strong generalization error.

 \begin{proposition}[LGE and the parametric Gibbs measure]\label{prop: LGE class Gibbs}
      Assume that Assumption~\ref{ass:expected_loss_KL} and $\mathbb{E}_{Z\sim\nupop}\big[\|Z\|^{16}\big]<\infty$ hold. Then, the following upper bound holds on the $L_2$ generalization error of the parametric Gibbs measure
\begin{equation}
\begin{split}
  | \mathrm{gen}_{L_2}(\mfm^{\beta,\sigma}(\nu_n),\nupop)|\leq \frac{1}{n} \Big(4K_1 + 2K_1^{1/2} A^{1/2}+\frac{A}{n} \Big),
    \end{split}
\end{equation}
where, for some $c>0$ independent of $n$,
\[\begin{split}
A&= c\,\mathbb{E}_{Z_1}[(1+\|Z_1\|^2)^{8}]<\infty,\\
K_1&= c\,\mathbb{E}_{Z_1}\Big[(1+\|Z\|^2)^4 \Big]< \infty.
\end{split}\] 
 \end{proposition}
\end{document}